\documentclass[runningheads]{llncs}

\usepackage{eccv}

\usepackage{eccvabbrv}

\usepackage{graphicx}\graphicspath{{images/}}
\usepackage{booktabs}
\usepackage{enumitem}

\usepackage[accsupp]{axessibility}  

\usepackage{hyperref}
\usepackage{orcidlink}
\usepackage{graphicx}\graphicspath{{images/}} 
\usepackage[table,x11names]{xcolor}
\usepackage{array}
\usepackage{tabularx}
\usepackage{import}
\usepackage{pifont}
\usepackage{fontawesome5}
\usepackage{relsize}

\newcommand{\beginsupplement}{ %
  \setcounter{figure}{0}       %
  \setcounter{table}{0}        %
  \setcounter{equation}{0}     %
  \setcounter{section}{0}      %
  \renewcommand{\thefigure}{{S}\arabic{figure}}%
  \renewcommand{\thetable}{{S}\arabic{table}}%
  \renewcommand{\theequation}{{S}\arabic{equation}}%
  \renewcommand{\thesection}{{S}\arabic{section}}%
  \crefname{figure}{Fig.\,S}{Figs.\,S}%
  \crefname{table}{Table\,S}{Tables\,S}%
  \crefname{equation}{Eq.\,S}{Eqs.\,S}%
}

\usepackage{tikz}
\usetikzlibrary{calc, positioning, shadows, shadings, arrows.meta, shapes.symbols, shapes.geometric}

\definecolor{dinoFrozen}{RGB}{200, 200, 200} 
\definecolor{dinoTuned}{RGB}{255, 140, 0}    
\definecolor{decoder}{RGB}{70, 130, 180}     
\definecolor{skip}{RGB}{220, 20, 60}         

\begin{document}

\title{Unwarping the Lens: A Physics-Grounded Approach to Video Glasses Removal} 

\titlerunning{Unwarping the Lens}

\author{Radim Spetlik\inst{1}$^*$\orcidlink{0000-0002-1423-7549} \and
David Futschik\inst{2}\orcidlink{0000-0003-3254-0290} \and Radek Danecek\inst{2}\orcidlink{0000-0002-1651-030X} \and\\ Feitong Tan\inst{2} \and Ziqian Bai\inst{2} \and Rohit Pandey\inst{2} \and Yinda Zhang\inst{2}
}

\authorrunning{R.~Spetlik et al.}

\institute{Czech Technical University in Prague, Faculty of Electrical Engineering \and
Google, $^*$ work done while at Google}

\maketitle

\begin{figure*}[t]
  \centering
  \renewcommand{\arraystretch}{0}
  \setlength{\tabcolsep}{0pt}
  \begin{tabularx}{\textwidth}{@{} X @{\hskip 1pt} X @{\hskip 1pt} X @{\hskip 1pt} X @{\hskip 1pt} X @{}}
    \includegraphics[width=\linewidth]{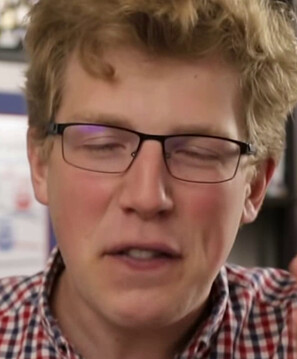} & \includegraphics[width=\linewidth]{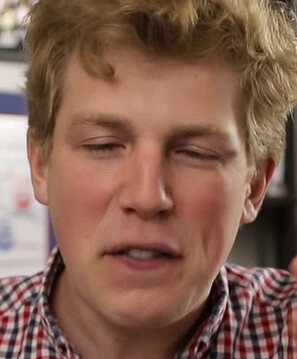} & \includegraphics[width=\linewidth]{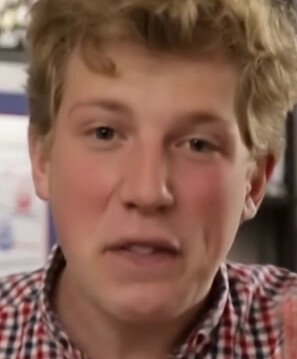} & \includegraphics[width=\linewidth]{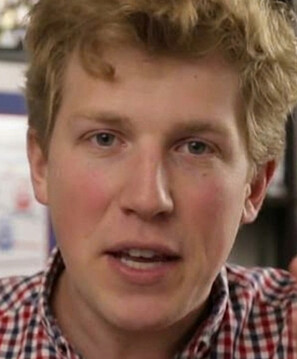} & \includegraphics[width=\linewidth]{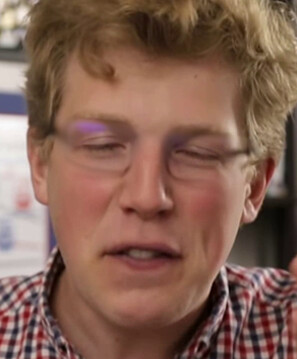} \\[0.5em]%
    \parbox[b]{\linewidth}{\centering (a) Input\hspace{2em}Frame} & \parbox[b]{\linewidth}{\centering (b) Our Approach} & \parbox[b]{\linewidth}{\centering (c) Runway Gen-4.5~\cite{runway_ai_inc_runway_2025}} & \parbox[b]{\linewidth}{\centering (d) Nano Banana~\cite{google_gemini_2025}} & \parbox[b]{\linewidth}{\centering (e) Pro-\hspace{2em}Painter~\cite{zhou_propainter_2023}} \\%
    \end{tabularx}
    \caption{Joint Feature-Spatial network removes glasses in videos producing results that are both temporally consistent and structurally faithful. Given an input video frame~(a), our approach~(b) effectively erases the glasses while preserving facial identity. In comparison, the commercial video editing model Runway Gen-4.5~\cite{runway_ai_inc_runway_2025} alters the subject's expression and introduces blurriness~(c). Similarly, the image-based editor Nano Banana~\cite{google_gemini_2025} fails to preserve the expression when applied to video frames~(d). The video inpainting method ProPainter~\cite{zhou_propainter_2023} faces different limitations, resulting in incomplete glasses removal~(e). In contrast, our method~(b) ensures coherent removal and maintains high-fidelity structural integrity throughout the sequence. Source: CelebV-Text~\cite{yu_celebv-text_2023}.}
  \label{fig:front_page_teaser}
\end{figure*}

\begin{abstract}
High-fidelity removal of eyeglasses from video is a major challenge in facial attribute editing, as the underlying facial geometry is often obscured by complex refractive distortions and view-dependent specular reflections.
While large-scale generative priors have shown promise in eyeglasses removal via static image inpainting, they often lack the structural constraints necessary to maintain identity, expression, and pose, leading to visible ``identity drift'' in both static images and dynamic sequences.
In this paper, we propose a novel transfer framework that addresses the stochastic nature of generative priors.
Our pipeline first extracts high-fidelity synthetic face images from a commercial-grade generative model (Nano Banana, Gemini 3 Pro Image), regularizes them via a three-stage structural filtering process to preserve identity, expression, and pose, and finally applies physically-based simulation of lens optics during training to provide diverse, paired data.
This process transfers Nano Banana's photo-realistic, multi-view knowledge into a specialized restoration architecture, \textit{JFSnet} (Joint Feature-Spatial network).
JFSnet integrates DINOv2-based semantic features with a convolutional decoder for spatial reconstruction, leveraging translation equivariance constraints to improve temporal consistency and high-frequency detail preservation.
Evaluations on the curated Flickr-Faces-HQ (FFHQ) subset (12,163 images) show that our approach achieves high fidelity and structural accuracy, while maintaining inference speed of 27.68 FPS.
In perceptual studies on CelebV-Text video sequences, our results are consistently preferred over diffusion and GAN-based baselines for ocular consistency, temporal stability, and overall restoration quality.
\end{abstract}

\section{Introduction}
\label{sec:intro}

The growth of high-resolution video content has increased the demand for sophisticated facial attribute editing.
While significant progress has been made in global style transfer and large-scale object manipulation~\cite{openai_sora_2025,google_veo2_2025,bar-tal_lumiere_2024}, precise local editing remains a difficult task.
Eyeglasses, specifically, are not just blocking layers; they are complex optical elements that introduce depth-dependent refractive distortions and view-dependent specular reflections.
Standard generative inpainting models~\cite{rombach_high-resolution_2022,yu_inpaint_2023} tend to struggle to separate these rare optical effects from the underlying facial geometry, which can lead to ``identity drift'' and structural inconsistencies in both static images and videos.
Existing approaches to glasses removal~\cite{lee_byeglassesgan_2020,shalev-arkushin_v-lasik_2025} primarily rely on stochastic generative priors to hallucinate the occluded regions.
However, in the presence of strong refractive lenses, the human visual system relies on stable geometric cues, such as the continuity of the head contour and the undistorted shape of the eyes, to maintain identity.
When a model relies on black-box priors to reconstruct the eye area, it often produces results that look realistic in isolation but can fail to preserve the subject's unique structural features, such as the exact expression, pose, and identity.
Furthermore, applying these methods frame-by-frame in a naive manner leads to temporal flickering, as the generative process is unconstrained across the temporal domain.
In this work, we propose a novel transfer framework that addresses the stochastic nature of these generative priors.
Our approach leverages large-scale models like Nano Banana (Gemini 3 Pro Image)~\cite{google_gemini_2025}, which provide photo-realistic results with multi-view consistency, but often struggle to preserve fine-grained identity, expression, and pose.
To address these limitations, we propose improvements to both data curation and model architecture. On the side of data generation, we introduce a structured curation and augmentation pipeline: (i)~we perform multi-view generative synthesis of eyeglass-free (``clean'') and glasses-wearing identities in sets of 13 poses and expressions using Nano Banana, (ii)~we apply a structural filtering process --- first identifying pose shifts via background $L_1$ error, then filtering gaze and identity drift via localized structural consistency analysis, and finally utilizing a learned reconstruction threshold --- to ensure a~high-fidelity curated source, and (iii) we finally apply physically-based substitution of lens optics during training to provide diverse and realistic pairs.
This process allows us to transfer the knowledge of large-scale generative models into a restoration architecture, \textit{JFSnet}, to achieve inference speeds suitable for real-time video applications.
In contrast to latent diffusion approaches that are constrained by the manifold of a fixed pre-trained autoencoder, our method operates directly in pixel space. This allows JFSnet to preserve facial details that might be lost during latent reconstruction.
By leveraging a self-supervised DINOv2-based encoder, we capture stable semantic priors that are invariant to local distortions, while a~precision-focused convolutional decoder reconstructs fine facial features.
To address the challenge of temporal instability without the high computational cost of multi-frame propagation or flow estimation, we enforce \emph{translation equivariance} with a training constraint.
By encouraging the network to commute with spatial shifts, we improve the coherence of restored features across frames, helping to suppress temporal flickering.
\textbf{In summary, our contributions are as follows. We}
\begin{enumerate}[nosep]
\renewcommand{\labelenumi}{(\roman{enumi})}
    \item formulate video glasses removal as the offline transfer of a generative prior into a deterministic, feed-forward model, and introduce a pipeline that combines curated generative face sets with physically-based augmentation of lens optics during training;
    \item present a framework utilizing \textit{JFSnet}, a ViT-CNN architecture enabling identity preservation and temporal consistency;
    \item evaluate on the curated FFHQ subset, outperforming state-of-the-art generative and video inpainting baselines with inference speed of 27.68 FPS;
    \item demonstrate through perceptual studies that our approach is preferred across gaze preservation, temporal stability, removal quality, identity preservation, and sharpness.
\end{enumerate}
Project site is available at \href{https://radimspetlik.github.io/unwarpingthelens/}{radimspetlik.github.io/unwarpingthelens/}.

\section{Related Work}

\subsection{Eyeglasses Removal}
Early work in glasses removal primarily focused on static portrait editing using Generative Adversarial Networks (GANs). 
ByeglassesGAN~\cite{lee_byeglassesgan_2020} and subsequent work~\cite{lyu_portrait_2022} leveraged GAN-based architectures to synthesize eyeglass-free ocular regions, with the latter utilizing 3D synthetic data to improve the restoration of cast shadows. 
More recently, video-specific approaches have emerged to address temporal consistency. 
V-LASIK~\cite{shalev-arkushin_v-lasik_2025} introduces identity-preserving constraints and color normalization to align restored regions with the rest of the face. 
Similarly, IP-FaceDiff~\cite{anand_ip-facediff_2025} utilizes diffusion priors to handle the removal of eyewear in dynamic sequences. 
While these methods show promise, they often exhibit identity drift and local blurring because they rely on stochastic generative priors that lack explicit geometric constraints. 
Our work differs by utilizing a physics-based simulation of lens optics during training, allowing the model to learn to invert the specific refractive distortions introduced by eyewear.

\subsection{Generative Image and Video Editing}
Advances in diffusion models have enabled broad image editing capabilities~\cite{brooks_instructpix2pix_2023,tsaban_ledits_2023,hertz_prompt--prompt_2023}. 
InstructPix2Pix~\cite{brooks_instructpix2pix_2023} and LEDITS~\cite{tsaban_ledits_2023} allow for instruction-based and semantic editing, while video-to-video translation models~\cite{kara_rave_2024,geyer_tokenflow_2023,qi_fatezero_2023} propagate edits across frames to maintain coherence. 
However, many of these methods operate in the latent space of pre-trained autoencoders, which introduce a bottleneck that inevitably sacrifices fine-grained spatial details like skin texture and eye geometry. 
Furthermore, while general-purpose models excel at global style transfer, they often struggle with precise local attribute removal, frequently failing to preserve the underlying facial structure.

Video inpainting and object removal methods~\cite{zhou_propainter_2023,zhang_flow-guided_2022,zeng_learning_2020} address occlusions by propagating information from adjacent frames. 
However, in the context of glasses removal, the ``background'' is the subject's own eyes --- critical features for identity that are often refracted or partially occluded rather than entirely missing. 
Generic inpainting approaches lack the specialized facial priors needed to accurately reconstruct these details, often leading to hallucinatory results that do not meet the bar of perceived geometric fidelity.

\subsection{Identity Preservation and 3D-Aware Editing}
Preserving identity is a core challenge in facial editing. 
Recent works utilize pre-trained Vision Transformers (ViTs) as semantic identity priors~\cite{oquab_dinov2_2023,spetlik_structureiser_2025}, while 3D-aware methods leveraging Morphable Models (3DMMs)~\cite{blanz_morphable_1999} or Neural Radiance Fields (NeRFs)~\cite{mildenhall_nerf_2021} maintain consistency under varying viewpoints. 
Recent methods integrate physical optical transport --- refraction, shadows, and specularities --- into 3D morphable models and neural radiance fields to accurately model eyewear~\cite{li_megane_2023, zhan_nerfrac_2023, li_eyenerf_2022}. 
Similarly, 3D Gaussian Splatting enables high-fidelity dynamic avatars~\cite{serifi_hypergaussians_2026, qian_gaussianavatars_2024, liang_fastavatar_2025, zhang_hravatar_2025}, though these representations rarely support removal or refraction. 
While some explicitly regress 3D facial topology to handle occlusion~\cite{zhao_generative_2022}, these representations often demand multi-view calibration or expensive optimization. 
Our approach bridges the gap between 2D translation and 3D geometric reasoning by incorporating physics-based refraction models, ensuring identity preservation without the need for per-scene 3D reconstruction or expensive overhead.

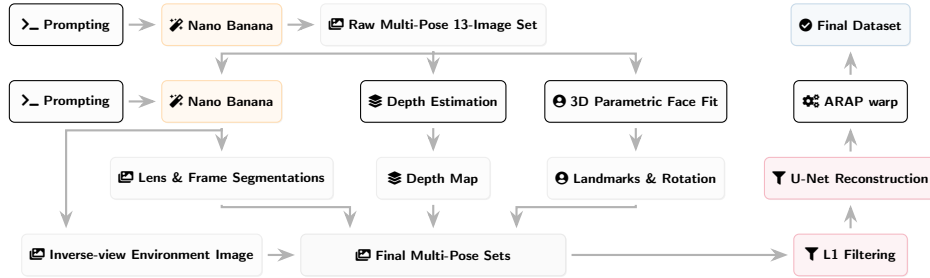
\begin{figure}[t] 
  \centering
  \begin{tikzpicture}[
    node distance=0.45cm and 0.5cm,
    wstep/.style={draw, rounded corners=2pt, minimum width=1.5cm, minimum height=0.55cm, align=center, font=\sffamily\tiny, fill=white, line width=0.4pt, drop shadow={shadow xshift=0.4pt, shadow yshift=-0.4pt, opacity=0.1}},
    arrow/.style={-Stealth, thick, color=black!30, shorten >=2pt, shorten <=2pt}
  ]
    \node (prompt) [wstep, fill=gray!2] {\faTerminal\ \textbf{Prompting}};
    \node (banana) [wstep, right=of prompt, fill=dinoTuned!5, draw=dinoTuned!30] {\faMagic\ \textbf{Nano Banana}};
    \node (raw) [wstep, right=of banana, fill=dinoFrozen!5, draw=dinoFrozen!30] {\faImages\ \textbf{Raw Multi-Pose 13-Image Set}};
    
    \node (depthModel) [wstep, below=of raw, fill=gray!2] {\faLayerGroup\ \textbf{Depth Estimation}};
    \node (depth) [wstep, below=of depthModel, fill=dinoFrozen!5, draw=dinoFrozen!30] {\faLayerGroup\ \textbf{Depth Map}};
    \node (segNB) [wstep, below=of banana, fill=dinoTuned!5, draw=dinoTuned!30] {\faMagic\ \textbf{Nano Banana}};
    \node (prompt2) [wstep, left=of segNB, fill=gray!2] {\faTerminal\ \textbf{Prompting}};
    \node (seg) [wstep, below=of segNB, fill=dinoFrozen!5, draw=dinoFrozen!30] {\faImages\ \textbf{Lens \& Frame Segmentations}};
    \node (flameModel) [wstep, right=of depthModel, fill=gray!2] {\faUserCircle\ \textbf{3D Parametric Face Fit}};
    \node (flame) [wstep, below=of flameModel, fill=dinoFrozen!5, draw=dinoFrozen!30] {\faUserCircle\ \textbf{Landmarks \& Rotation}};

    \node (sets) [wstep, below=of depth, fill=dinoFrozen!10, draw=dinoFrozen!40, minimum width=3.5cm] {\faImages\ \textbf{Final Multi-Pose Sets}};
    \node (sceneView) [wstep, left=of sets, fill=dinoFrozen!5, draw=dinoFrozen!30] {\faImages\ \textbf{Inverse-view Environment Image}};

    \node (l1) [wstep, right=3cm of sets, fill=skip!5, draw=skip!30] {\faFilter\ \textbf{L1 Filtering}};
    \node (unet) [wstep, above=of l1, fill=skip!5, draw=skip!30] {\faFilter\ \textbf{U-Net Reconstruction}};
    \node (arap) [wstep, above=of unet, fill=gray!2] {\faCogs\ \textbf{ARAP warp}};
    \node (final) [wstep, above=of arap, fill=decoder!5, draw=decoder!30] {\faCheckCircle\ \textbf{Final Dataset}};

    \draw [arrow] (prompt) -- (banana);
    \draw [arrow] (prompt2) -- (segNB);
    \draw [arrow] (banana) -- (raw);
    
    \draw [arrow] (raw.south) -- +(0,-0.1) -| (segNB.north);
    \draw [arrow] (segNB.south) -- +(0,-0.1) -| (seg.north);
    \draw [arrow] (segNB.south) -- ++(0,-0.1) -- ++(-2,0) -| ($(sceneView.north) + (-1.0,0)$);
    \draw [arrow] (raw.south) -- (depthModel.north);
    \draw [arrow] (depthModel.south) -- (depth.north);
    \draw [arrow] (raw.south) -- +(0,-0.1) -| (flameModel.north);
    \draw [arrow] (flameModel.south) -- (flame.north);
    \draw [arrow] (sceneView.east) -- (sets.west);
    
    \draw [arrow] (seg.south) -- +(0,-0.1) -| ($(sets.north) + (-1.1,0)$);
    \draw [arrow] (depth.south) -- (sets.north);
    \draw [arrow] (flame.south) -- +(0,-0.1) -| ($(sets.north) + (1.1,0)$);
    
    \draw [arrow] (sets.east) -- (l1.west);
    
    \draw [arrow] (l1) -- (unet);
    \draw [arrow] (unet) -- (arap);
    \draw [arrow] (arap) -- (final);

  \end{tikzpicture}
  \caption{\textbf{Dataset creation pipeline.} Raw multi-pose face sets are generated from the Nano Banana prior and enriched with segmentation masks, monocular depth, and parametric face model fits. These sets undergo multi-stage filtering (joint background/ocular \textit{L1 Filtering}, followed by U-Net reconstruction) and an As-Rigid-As-Possible (ARAP) warp to align clean and glasses-wearing pairs, giving the final dataset.}
  \label{fig:data_generation_workflow}
\end{figure}

\section{Method}
\label{sec:method}

We propose a novel transfer framework for consistent glasses removal that addresses the tendency of generative priors toward identity divergence and the introduction of structural inconsistencies (see Fig.~\ref{fig:front_page_teaser}).
Our approach utilizes a~restoration network, \emph{JFSnet}, which learns to invert complex optical effects while preserving identity, expression, and pose.
\subsection{Generative Data Creation and Filtering}
\label{sec:data_curation}
To support our training pipeline, we constructed a dataset of eyeglass-free (``clean'') and glasses-wearing face pairs using a generative workflow powered by Nano Banana (Gemini 3 Pro Image)~\cite{google_gemini_2025} shown in Fig.~\ref{fig:data_generation_workflow}.
For each synthetic identity, we prompted the model to generate a set of 13 high-fidelity clean images capturing a~diverse range of head poses and facial expressions.
We further generated a standalone image of eyeglasses and a corresponding set of composed images where the subject wears the glasses in each pose, along with precise segmentation masks for the left lens, right lens, and the frames for every sample in the sequence.
See prompts in Supplementary Sec.~\ref{sec:data_generation_prompts}.
This provided us with a comprehensive set of multi-pose and multi-expression pairs with ground-truth frame textures and pixel-accurate lens regions.
A challenge in generative data synthesis is maintaining spatial alignment between the eyeglass-free and glasses-wearing pairs due to a number of effects, mainly the propensity of ViTs to overcompress spatial information and stochasticity in latent generative models.
To address this, we used the composed Nano Banana images and their associated masks to define the precise geometric boundaries of the frames and lenses in each pose.
Furthermore, we employ an \textit{As-Rigid-As-Possible} (ARAP)~\cite{igarashi_as-rigid-as-possible_2005} deformation mesh optimized with landmark constraints and a multi-scale photometric loss to align the clean face to the exact pose and facial configuration found in the glasses-wearing image (see Supplementary Sec.~\ref{sec:arap}).

To ensure high data quality, we performed a three-stage hard-sampling-based filtering process.
In the first two stages, we perform background and ocular $L_1$ filtering (\textit{L1 Filtering} in Fig.~\ref{fig:data_generation_workflow}) to filter out identities with significant shifts in head pose, gaze, or eye identity. This process effectively eliminates samples where the generative model introduced unwanted expression drift; such expression drift is exemplified by Nano Banana in Fig.~\ref{fig:front_page_teaser}~(d).
This stage remains crucial because we down-weight the physics-based lens substitution for extreme head poses --- specifically rotations approaching a profile view (beyond roughly three-quarters of the way from frontal to full profile) --- where monocular depth-based simulation becomes less reliable and we rely on the generative model's native outputs.
In the third stage, we trained a lightweight U-Net on the remaining pairs to perform a preliminary restoration task, which serves as a quality proxy; identities with high reconstruction error were discarded as they typically exhibited residual misalignments or generative artifacts.
This elaborate filtering resulted in a final training dataset of 1,860 identities exhibiting high structural and photometric consistency across the 13-image sets (see the Supplementary Sec.~\ref{sec:dataset_overview}).

\begin{figure*}[t]
  \centering
  \renewcommand{\arraystretch}{0}
  \setlength{\tabcolsep}{0pt}
  \begin{tabularx}{\textwidth}{@{} X X X X @{}}
    \includegraphics[width=\linewidth]{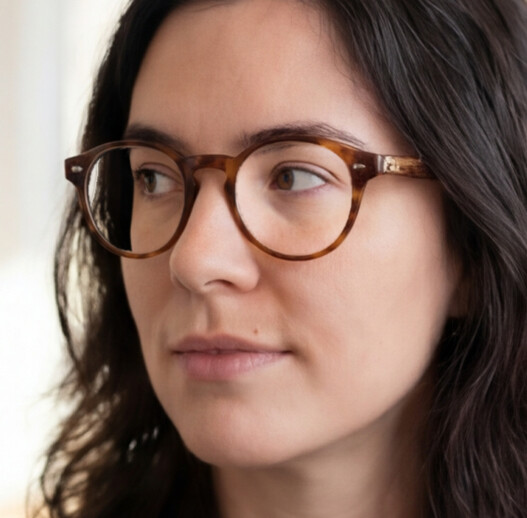} & \includegraphics[width=\linewidth]{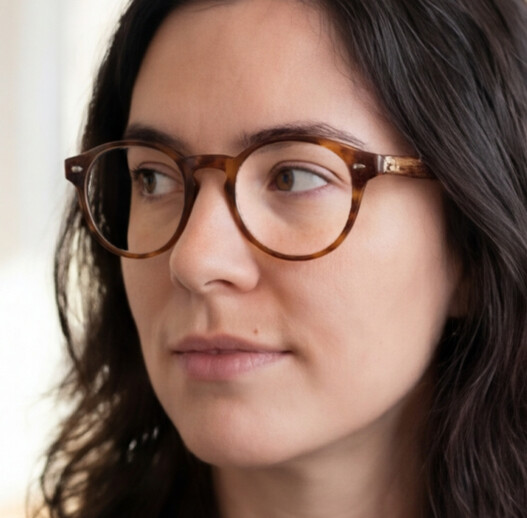} & \includegraphics[width=\linewidth]{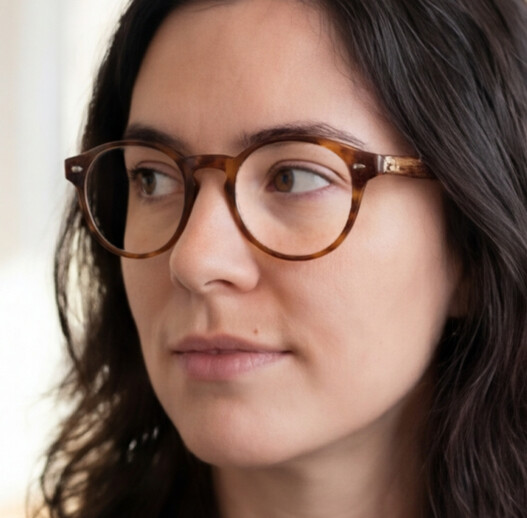} & \includegraphics[width=\linewidth]{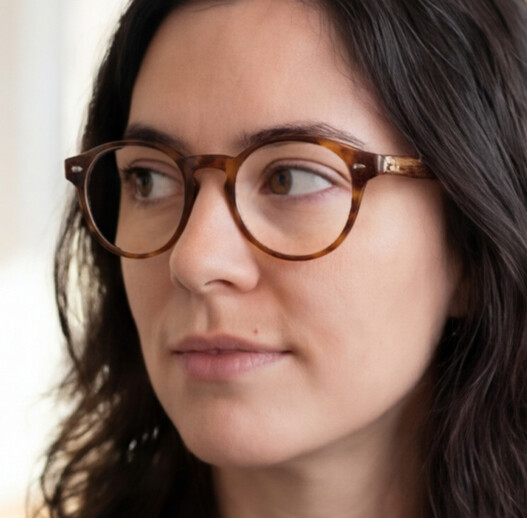}
    \\[0.4em]
    \parbox[b]{\linewidth}{\centering \footnotesize $-6$} & \parbox[b]{\linewidth}{\centering \footnotesize $-4$} & \parbox[b]{\linewidth}{\centering \footnotesize $-2$} & \parbox[b]{\linewidth}{\centering \footnotesize $+1$} \\
  \end{tabularx}
  \caption{\textbf{Refraction simulation.} Physics-based lens distortion for $-6, -4, -2$, and $+1$ diopters on a sample from our dataset. Our refraction model captures realistic magnification and minification while preserving facial identity and structure.}
  \label{fig:refraction_visualization}
\end{figure*}

\subsection{Physically-Grounded Training Augmentation}\label{sec:synthesis}
Although this dataset is rich in identities, the optical effects represented in generated data are far more constrained, likely following the generative model's simplicity bias. To address the lack of large-scale paired data, we introduce a physics-based augmentation pipeline that simulates the optical properties of lenses on the curated generative face sets (see Fig.~\ref{fig:refraction_visualization}).
This pipeline operates during training to provide a source of consistent and diverse training pairs.
\vspace{-0.5em}\paragraph{Refraction Simulation (Fig.~\ref{fig:refraction_visualization}).}
To ensure realistic placement and distortion, we first identify the lens boundaries using precise segmentation masks provided by Nano Banana.
The relative 3D pose of the head is estimated by fitting a 3D face parametric model \cite{egger_3d_2020} to the input frame to extract the global rotation $\mathbf{R}$ and translation $\mathbf{t}$ relative to the camera.
We define the optical center of each lens by back-projecting the detected pupil landmarks using the depth map from a single-frame depth estimation model~\cite{yang_depth_2024}.
To resolve the scale ambiguity inherent in monocular depth estimation, we assume a standard adult head width to anchor the metric scale.
The lenses are then positioned by projecting these metric 3D pupil coordinates 15mm along the head pose direction (the normal vector derived from $\mathbf{R}$).
We assume a standard perspective camera model with a typical portrait focal length to initialize the 3D projection.
The corrective lens is represented as a physical 3D object defined by two spherical surfaces with radii $R_1$ and $R_2$, positioned relative to the eye centers.
These radii are derived from the target diopter strength $D$ using Vogel's Rule~\cite{brooks_system_2007}: $D_1 = D/2 + 6.0$ and $D_2 = D - D_1$, where $D = (n-1)(1/R_1 - 1/R_2)$.
For each pixel within the lens boundary, a ray is cast and intersected with these surfaces.
At each interface, the ray direction is updated via Snell's law ($n \approx 1.5$).
The resulting intersection with the facial surface defines the source coordinates for the dense warp field $\mathcal{W}$, simulating the depth-dependent magnification of prescription eyewear.
Our rendering pipeline operates on high-resolution grids to preserve detail.
\vspace{-0.5em}\paragraph{Reflection Simulation.}
Specular reflections $\mathcal{R}$ are simulated by leveraging Nano Banana to generate an inverse-view environment map, conditioned on the subject's portrait, that approximates the scene behind the camera, sampled along the reflected view ray $\mathbf{r}=\mathbf{i}-2(\mathbf{n}^\top\mathbf{i})\mathbf{n}$ ($\mathbf{i}$: incident view direction, $\mathbf{n}$: lens-surface normal).
To achieve realistic highlights, we simulate reflections in High Dynamic Range (HDR), allowing for strong specularities of glass surfaces.
We further augment these reflections by simulating a variable reflection tint and opacity, accounting for the diverse range of lens coatings and lighting conditions.
\vspace{-0.5em}\paragraph{Training Input Formation.}
To provide a physically grounded yet photo-realistic training source, we define the observed image $I \in \mathbb{R}^{H \times W \times 3}$ as a composition of generative templates and optical simulations.
Let $G_{clean}$ and $G_{glasses}$ be the multi-pose, multi-expression sets from our curated dataset.
During training, we synthesize the final input $I$ by selecting a pose-expression pair and substituting the lens regions of $G_{glasses}$ with a physically simulated rendering of the corresponding clean face $G_{clean}$:
\begin{equation}
    I = (1 - M_{lens}) \cdot G_{glasses} + M_{lens} \cdot (\mathcal{W}(G_{clean}) + \mathcal{R}) + \epsilon
\end{equation}
where $M_{lens}$ is a binary mask for the lens regions (provided by the left and right lens masks from the generative model), $\mathcal{W}(\cdot)$ represents the refractive warping operator, and $\mathcal{R}$ is the additive HDR specular reflection layer.
This formulation ensures that the network learns to invert the refractive transformation $\mathcal{W}$ and handle additive reflections $\mathcal{R}$ while benefiting from the realistic frame textures and global illumination present in the generative model $G_{glasses}$.
The network's goal is to map this input $I$ back to the original clean identity $G_{clean}$.
\subsection{JFSnet Architecture}
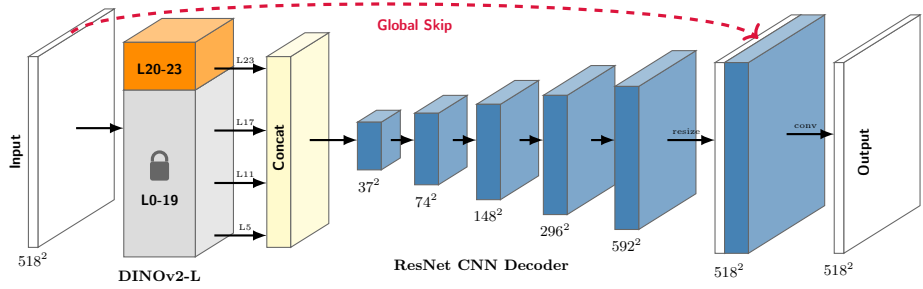
\begin{figure}[t]
    \centering
    \begin{tikzpicture}[
    x={(1cm,0cm)}, 
    y={(0cm,1cm)}, 
    z={(0.4cm,0.25cm)}, 
    scale=0.63,
    every node/.append style={transform shape},
    font=\sffamily
]

\definecolor{cBackboneFrozen}{RGB}{220,220,220}
\definecolor{cBackboneTuned}{RGB}{255,140,0}
\definecolor{cDecoder}{RGB}{70,130,180}
\definecolor{cSkip}{RGB}{220,20,60}
\definecolor{cConcat}{HTML}{FFFACD} 

\newcommand{\drawcuboid}[9]{
    \filldraw[fill=#8!70, draw=black!60] (#2+#5,#3,#4) -- ++(0,#6,0) -- ++(0,0,#7) -- ++(0,-#6,0) -- cycle;
    \filldraw[fill=#8!50, draw=black!60] (#2,#3+#6,#4) -- ++(#5,0,0) -- ++(0,0,#7) -- ++(-#5,0,0) -- cycle;
    \filldraw[fill=#8, draw=black!60] (#2,#3,#4) -- ++(#5,0,0) -- ++(0,#6,0) -- ++(-#5,0,0) -- cycle;
    
    \coordinate (#1-right) at (#2+#5, #3+#6/2, #4+#7/2);
    \coordinate (#1-left)  at (#2,    #3+#6/2, #4+#7/2);
    \coordinate (#1-top)   at (#2+#5/2, #3+#6, #4+#7/2);
    \coordinate (#1-front) at (#2+#5, #3+#6/2, #4);
    
    \node[below, yshift=-2pt] at (#2+#5/2, #3, #4) {#9};
}

\drawcuboid{input}{0}{-2}{0}{0.2}{4}{4}{white}{}
\node[rotate=90, anchor=south] at (0, 0, 0) {\bfseries Input};
\node[below] at (0.1, -2, 0) {$518^2$};
\draw[-latex, thick] (input-right) -- ++(1.0,0,0);

\drawcuboid{dinoF}{2}{-2.2}{0}{1.5}{3.5}{2}{cBackboneFrozen}{}
\node[font=\sffamily\bfseries\small, text=black] at (2.75, -1.0, 0) {L0-19};
\drawcuboid{dinoT}{2}{1.3}{0}{1.5}{1.0}{2}{cBackboneTuned}{}
\node[font=\sffamily\bfseries\small, text=black] at (2.75, 1.75, 0) {L20-23};
\node[align=center, below=0.5cm] at (2.75, -2.5, 0) {};
\node[font=\bfseries, text=black] at (2.75, -2.6, 0) {\bfseries DINOv2-L};


\draw[black!60, very thick] (2.6, -0.2, 0) -- ++(0, 0.05) arc (180:0:0.15) -- ++(0, -0.05);

\fill[black!60, rounded corners=1pt] (2.55, -0.6, 0) rectangle (2.95, -0.2, 0);


\draw[-latex, thick] (3.5, -2.0, 1) -- (4.6, -2.0, 1) node[pos=0.59, above, font=\tiny] {L5};
\draw[-latex, thick] (3.5, -0.9, 1) -- (4.6, -0.9, 1) node[pos=0.59, above, font=\tiny] {L11};
\draw[-latex, thick] (3.5, 0.2, 1)  -- (4.6, 0.2, 1)  node[pos=0.59, above, font=\tiny] {L17};
\draw[-latex, thick] (3.5, 1.5, 1) -- (4.6, 1.5, 1) node[pos=0.59, above, font=\tiny] {L23};

\drawcuboid{concat}{5}{-2}{0}{0.5}{4.0}{2}{cConcat}{}
\node[rotate=90] at (5.25, 0.1, 0) {\bfseries Concat};

\draw[-latex, thick] (5.5, 0, 1) -- (6.5, 0, 1);

\drawcuboid{dec1}{6.7}{-0.5}{0.5}{0.5}{1}{1}{cDecoder}{$37^2$}
\draw[-latex, thick] (dec1-right) -- ++(0.5,0,0);

\drawcuboid{dec2}{8}{-0.75}{0.25}{0.5}{1.5}{1.5}{cDecoder}{$74^2$}
\draw[-latex, thick] (dec2-right) -- ++(0.5,0,0);

\drawcuboid{dec3}{9.4}{-1}{0}{0.5}{2}{2}{cDecoder}{$148^2$}
\draw[-latex, thick] (dec3-right) -- ++(0.5,0,0);

\drawcuboid{dec4}{10.9}{-1.25}{-0.25}{0.5}{2.5}{2.5}{cDecoder}{$296^2$}
\draw[-latex, thick] (dec4-right) -- ++(0.5,0,0);

\drawcuboid{dec5}{12.5}{-1.5}{-0.5}{0.5}{3}{3}{cDecoder}{$592^2$}
\node[above=0.2cm] at (14.7, -2.93, 0) {$518^2$};

\node[above=0.2cm, font=\bfseries] at (9.5, -2.8, 0) {ResNet CNN Decoder};

\draw[-latex, thick] (dec5-right) -- ++(1.0,0,0) node[pos=0.4, above, font=\tiny] {resize};

\drawcuboid{skipInput}{14.6}{-2}{-0.5}{0.2}{4}{4}{white}{}
\drawcuboid{featureOutput}{14.8}{-2}{-0.5}{0.5}{4}{4}{cDecoder}{}

\draw[-latex, thick] (featureOutput-right) -- ++(1.0,0,0) node[pos=0.4, above, font=\tiny] {conv};

\drawcuboid{output}{17.1}{-2}{-0.5}{0.2}{4}{4}{white}{$518^2$}

\node[rotate=90, anchor=south] at (17.8, 0.0, 0) {\bfseries Output};

\draw[cSkip, dashed, very thick, ->] (input-top) to[out=45, in=135, looseness=0.22] node[pos=0.5, below=5pt] {\bfseries Global Skip} (skipInput-top);


\end{tikzpicture}\vspace{-2em}
    \caption{Overview of the proposed Joint Feature-Spatial network (JFSnet). Our architecture integrates a pre-trained DINOv2 encoder with a convolutional decoder to restore facial details behind eyeglasses. By leveraging multi-level feature integration and selective fine-tuning, the network bridges high-level semantic identity preservation with high-resolution spatial precision, ensuring consistent restoration across diverse facial poses and optical conditions.}
    \label{fig:method_overview}
\end{figure}

We utilize a restoration network ($f_{\theta}$), referred to as \textit{JFSnet}, designed to map the input image to the clean RGB face.
As detailed in Fig.~\ref{fig:method_overview}, \textit{JFSnet} is a combination of standard architectural blocks: a pre-trained DINOv2 (ViT-L/14) encoder~\cite{oquab_dinov2_2023} and a ResNet-based convolutional decoder.
To preserve representational power while adapting to the restoration task, we selectively fine-tune the deeper layers of the ViT backbone and fuse multi-level features into the decoder.
Crucially, by incorporating a direct skip connection from the input image, we bypass the information bottleneck typical of latent models.
This design allows JFSnet to reconstruct high-frequency details that might fall outside standard generative manifolds, ensuring that the restored facial features remain faithful to the original input. As discussed in Sec.~\ref{sec:arch_ablation}, this configuration was found to significantly outperform alternative architectures.
\subsection{Training Objectives}
The network $f_{\theta}$ is trained using a weighted combination of reconstruction and perceptual losses.
Let $f_{\theta}(I)$ be the predicted clean face and $G_{clean}$ be the ground truth clean face.
The total loss $\mathcal{L}$ is defined as:
\begin{equation}
    \begin{split}
        \mathcal{L} &= \lambda_{pix} \mathcal{L}_{pix} + \lambda_{perc} \mathcal{L}_{perc} + \lambda_{eye-pix} \mathcal{L}_{eye-pix} \\
        &\quad + \lambda_{eye-perc} \mathcal{L}_{eye-perc} + \lambda_{adv} \mathcal{L}_{adv} + \lambda_{temp} \mathcal{L}_{temp}
    \end{split}
\end{equation}
\emph{Global Reconstruction and Perceptual.}
$\mathcal{L}_{pix}$ is the global $L_1$ distance between the predicted output and the ground truth.
$\mathcal{L}_{perc}$ ensures high-level texture consistency across the full image, defined by the distance between Gram matrices of VGG-19 features.
\emph{Eye-specific Objectives.}
To restore features behind lenses, we utilize localized objectives centered on the center of gravity (CoG) of each lens mask within $64 \times 64$ pixel blocks.
$\mathcal{L}_{eye-pix}$ is the localized $L_1$ reconstruction loss, while $\mathcal{L}_{eye-perc}$ is the corresponding Gram-matrix based perceptual loss.

\emph{Adversarial Loss.}
$\mathcal{L}_{adv}$ incorporates feedback from a PatchGAN-based critic~\cite{isola_image--image_2017} to ensure realistic texture generation, encouraging the network to produce sharp, high-frequency details. The critic is trained in the standard GAN min-max setting~\cite{goodfellow_generative_2014}, though we omit this formulation for brevity.
\emph{Temporal Consistency.}
To ensure temporally stable results on video data, we enforce translation equivariance during training.
For each training sample, we simulate motion by applying a random spatial shift $\Delta s$ to the input image $I$.
The temporal consistency loss $\mathcal{L}_{temp}$ is defined as:
\begin{equation}
    \mathcal{L}_{temp} = \| f_{\theta}(\mathcal{T}_{\Delta s}(I)) - \mathcal{T}_{\Delta s}(f_{\theta}(I)) \|_1
\end{equation}
where $\mathcal{T}_{\Delta s}(\cdot)$ represents a translation by $\Delta s$.
By forcing $f_{\theta}$ to commute with small spatial translations, we ensure that high-frequency artifacts and identity-defining features move coherently with the subject's motion.
This constraint encourages the model to learn a mapping that is robust to spatial misalignments and resists flickering across frames.

\section{Experiments}   
\label{sec:experiments}

In order to assess the efficacy of our proposed approach, we evaluate the task of glasses removal from images and consistent removal from videos.
Our experiments are designed to validate the effectiveness of the physics-based synthesis pipeline and the \textit{JFSnet} architecture in handling optical distortions and maintaining temporal stability.
\subsection{Experimental Setup}

\begin{figure*}[t]
  \centering
  \renewcommand{\arraystretch}{0}
  \setlength{\tabcolsep}{0pt}
  \begin{tabularx}{\textwidth}{@{} X @{\hskip 5pt} @{} X @{} X @{} X @{} X @{} X @{} X @{} X @{{}}}
    \parbox[b]{\linewidth}{\centering \scriptsize Input Frame} & \parbox[b]{\linewidth}{\centering \scriptsize Our Approach} & \parbox[b]{\linewidth}{\centering \scriptsize Nano Banana~\cite{google_gemini_2025}} & \parbox[b]{\linewidth}{\centering \scriptsize TOE\\ \cite{lyu_portrait_2022}} & \parbox[b]{\linewidth}{\centering \scriptsize ProPainter\\ \cite{zhou_propainter_2023}} & \parbox[b]{\linewidth}{\centering \scriptsize FGT\\ \cite{zhang_flow-guided_2022}} & \parbox[b]{\linewidth}{\centering \scriptsize STTN\\ \cite{zeng_learning_2020}} & \parbox[b]{\linewidth}{\centering \scriptsize LEDITS\\ \cite{tsaban_ledits_2023}} \\%
    \includegraphics[width=\linewidth]{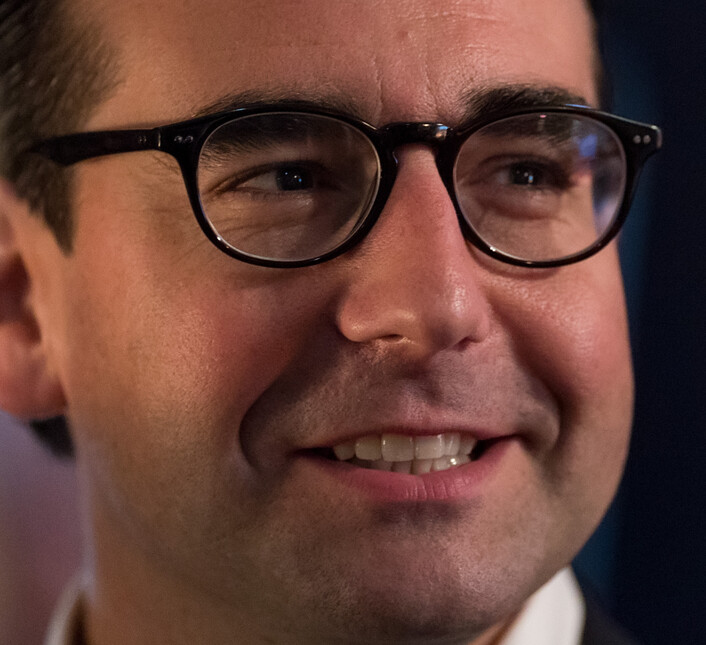} & \includegraphics[width=\linewidth]{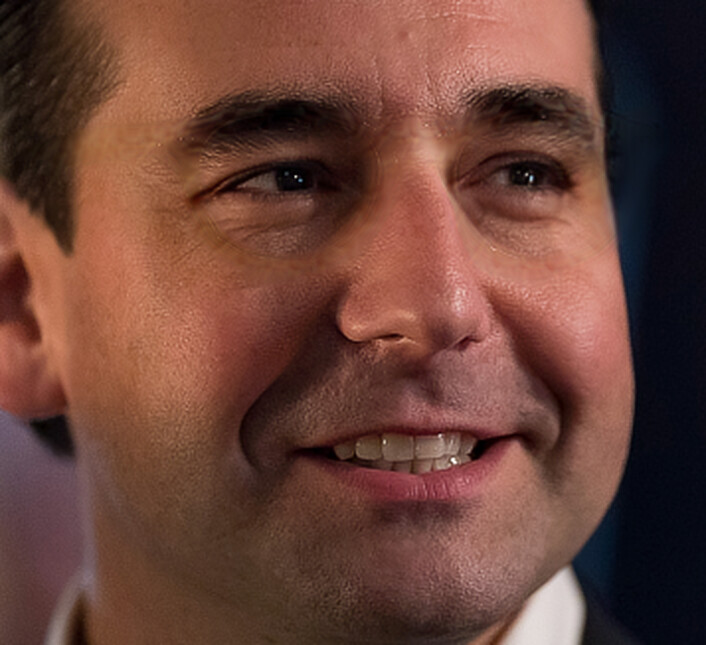} & \includegraphics[width=\linewidth]{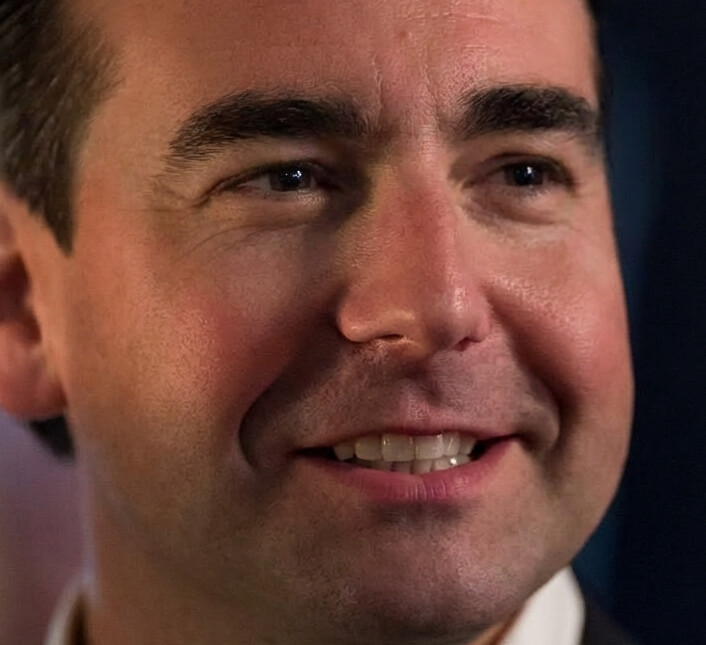} & \includegraphics[width=\linewidth]{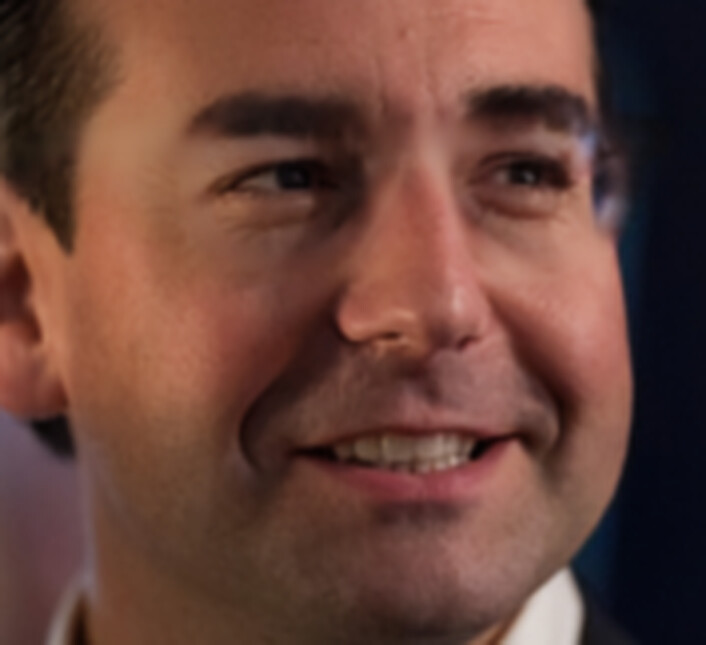} & \includegraphics[width=\linewidth]{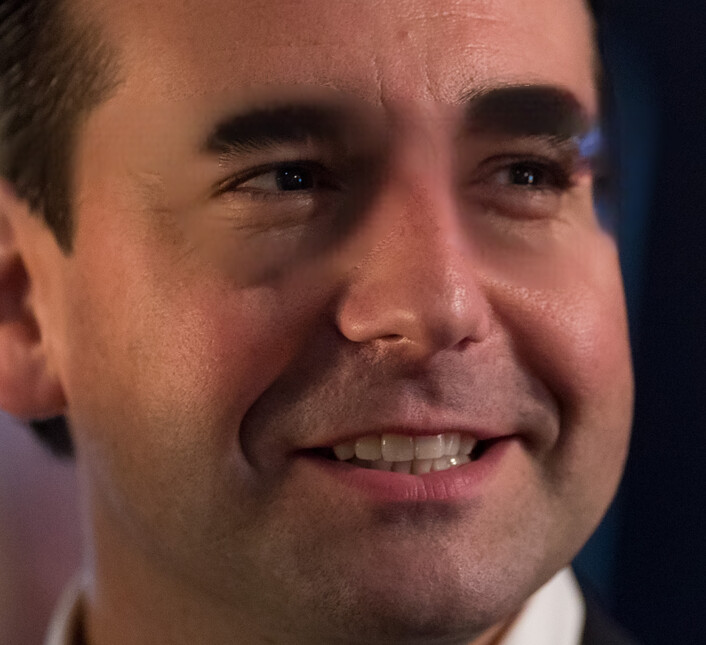} & \includegraphics[width=\linewidth]{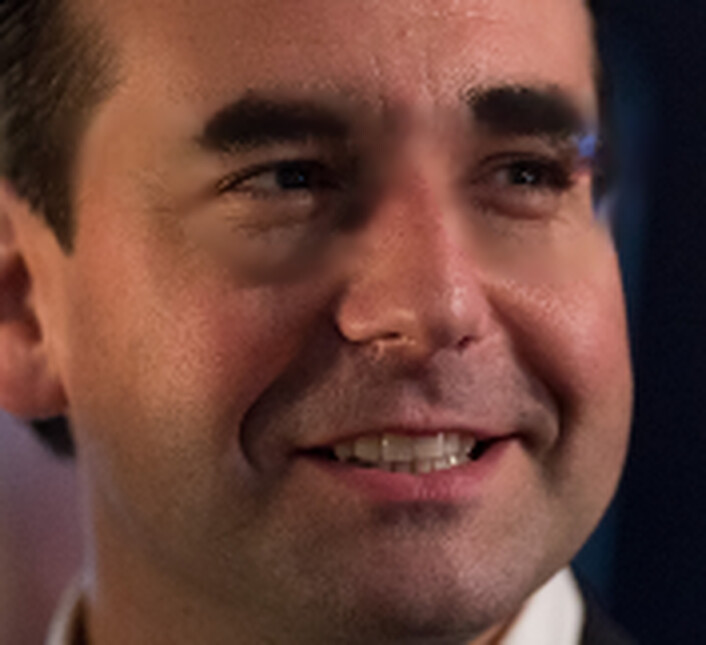} & \includegraphics[width=\linewidth]{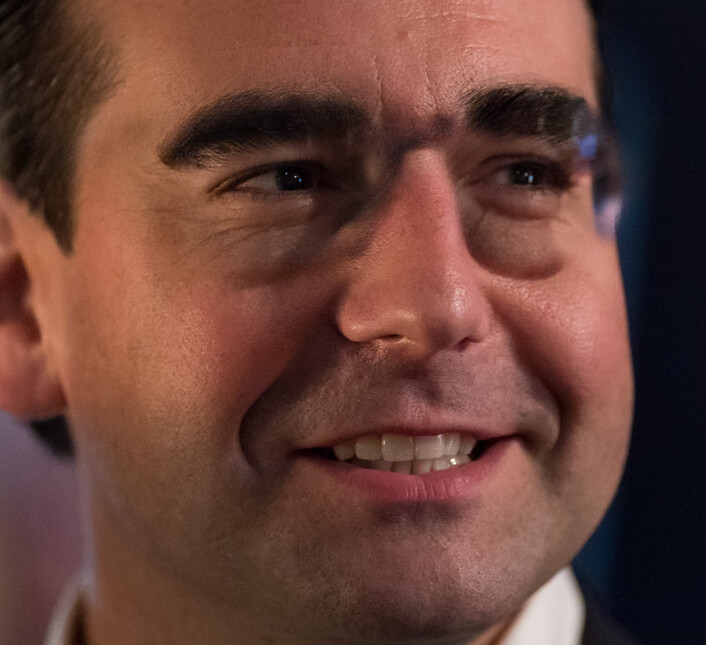} & \includegraphics[width=\linewidth]{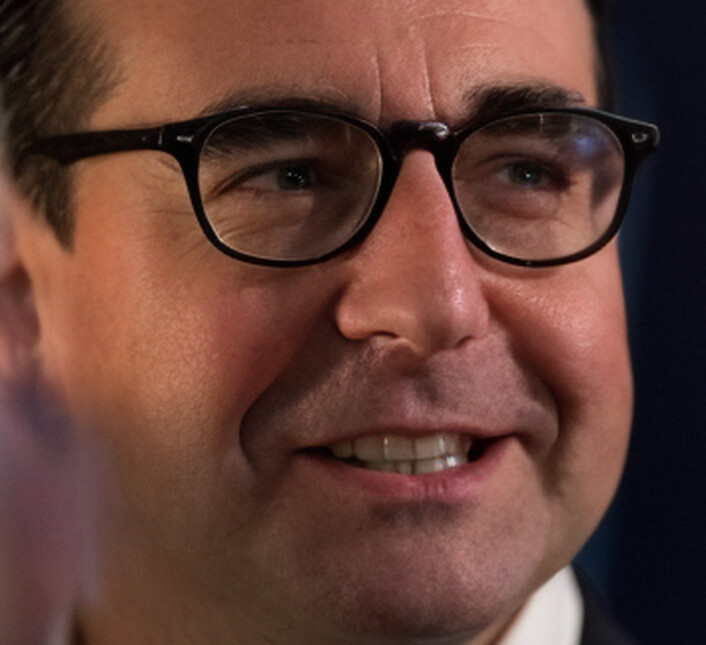} \\
    \includegraphics[width=\linewidth]{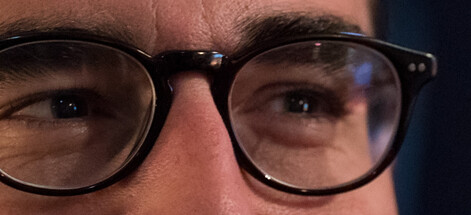} & \includegraphics[width=\linewidth]{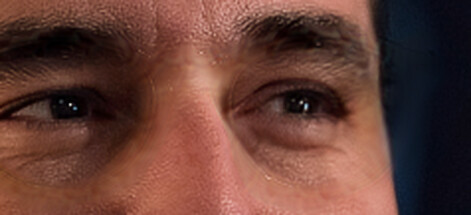} & \includegraphics[width=\linewidth]{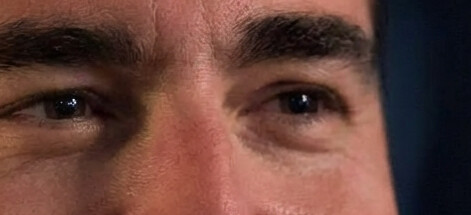} & \includegraphics[width=\linewidth]{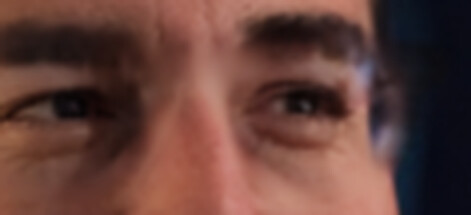} & \includegraphics[width=\linewidth]{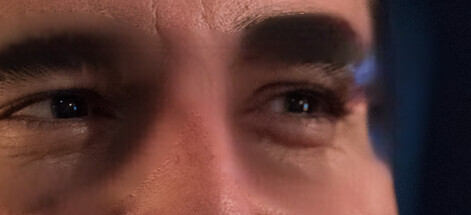} & \includegraphics[width=\linewidth]{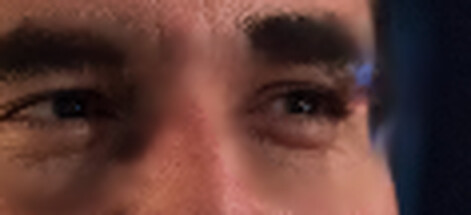} & \includegraphics[width=\linewidth]{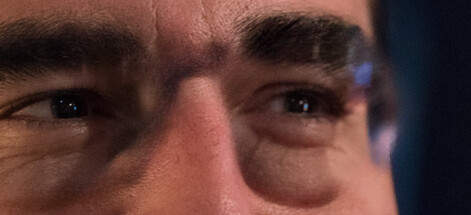} & \includegraphics[width=\linewidth]{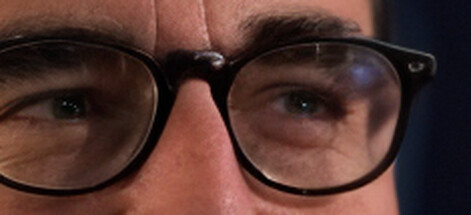} \\[0.2em]
    \includegraphics[width=\linewidth]{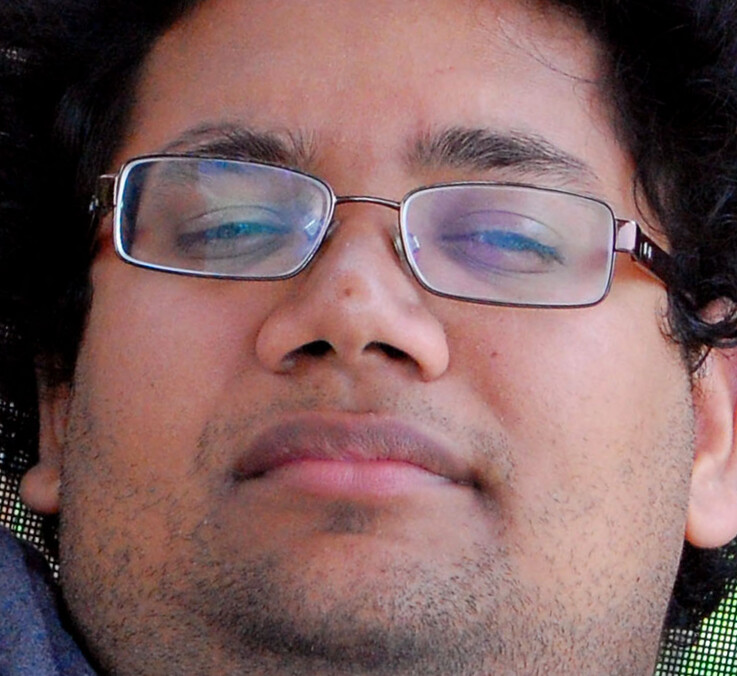} & \includegraphics[width=\linewidth]{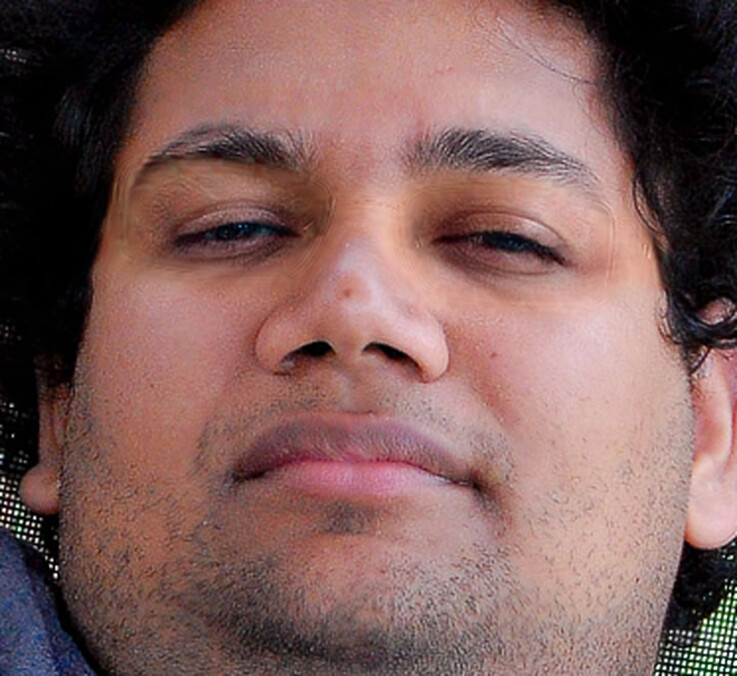} & \includegraphics[width=\linewidth]{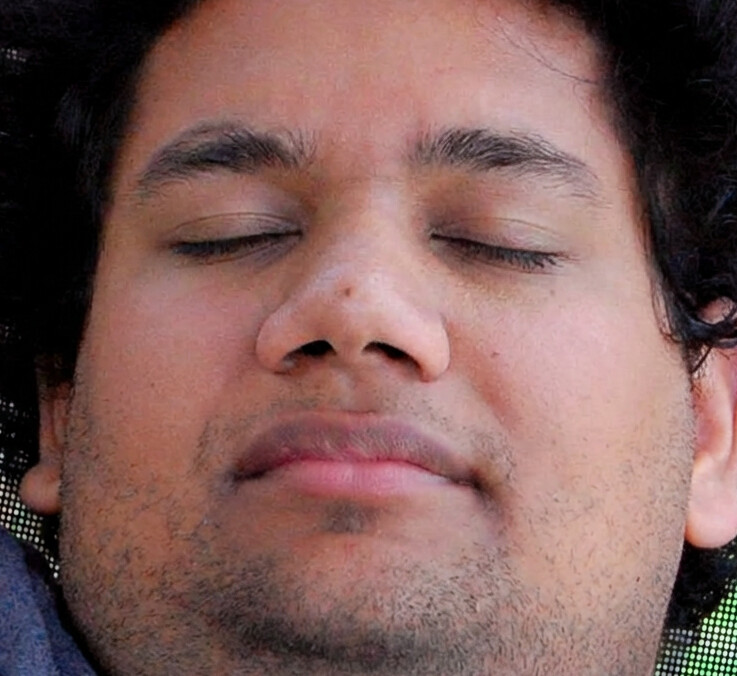} & \includegraphics[width=\linewidth]{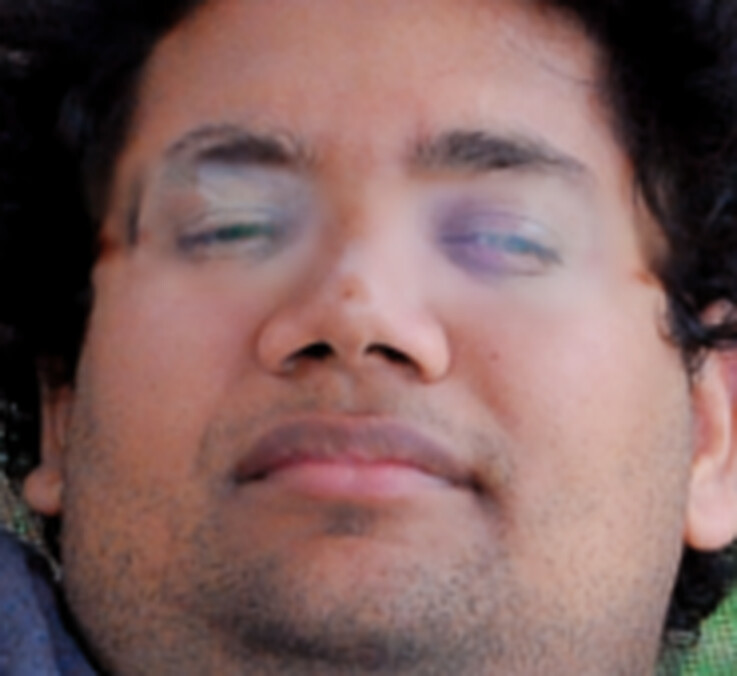} & \includegraphics[width=\linewidth]{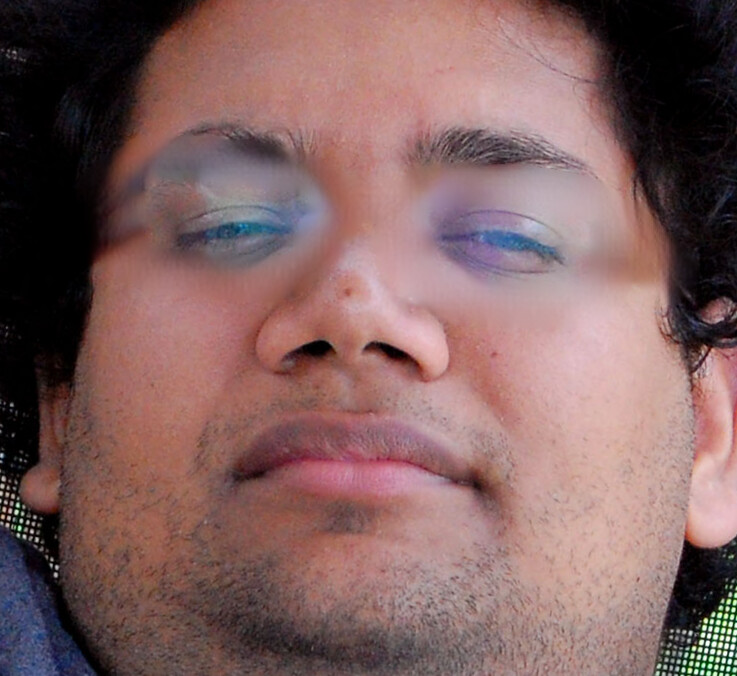} & \includegraphics[width=\linewidth]{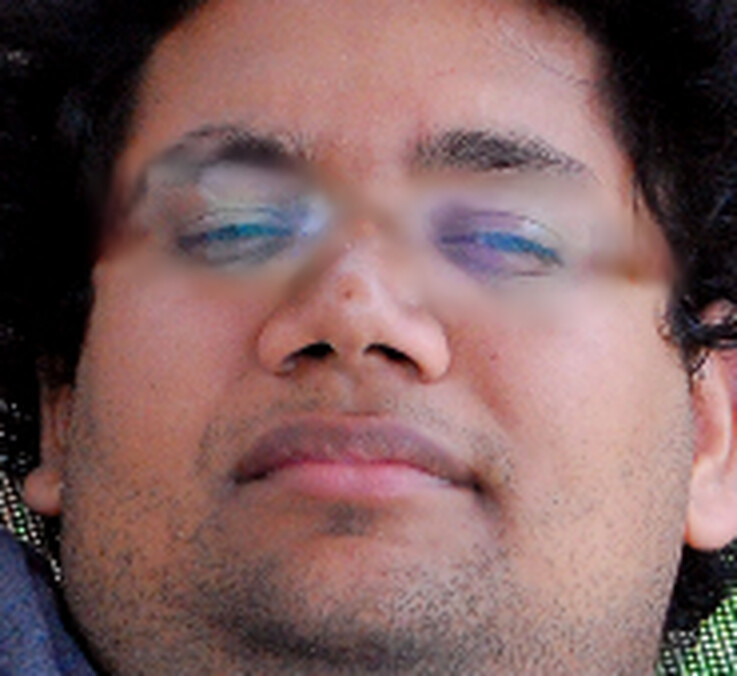} & \includegraphics[width=\linewidth]{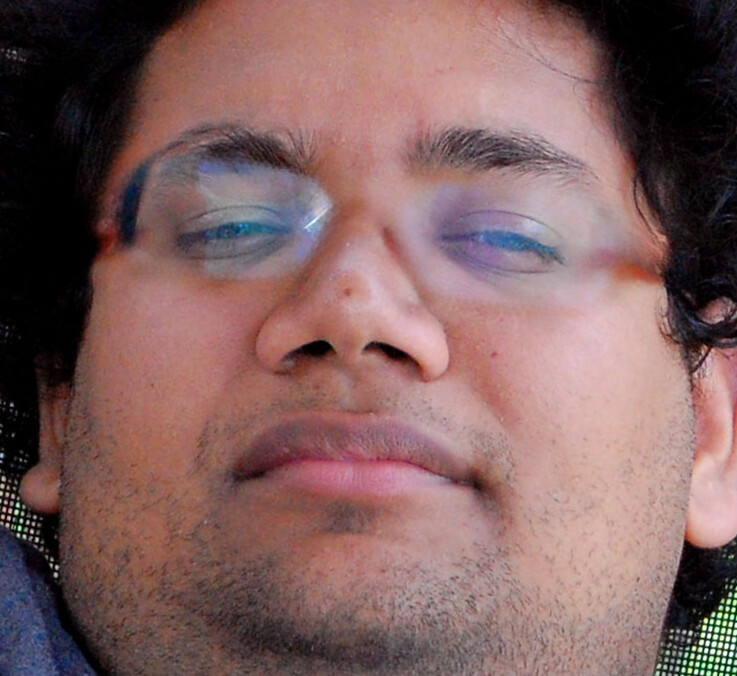} & \includegraphics[width=\linewidth]{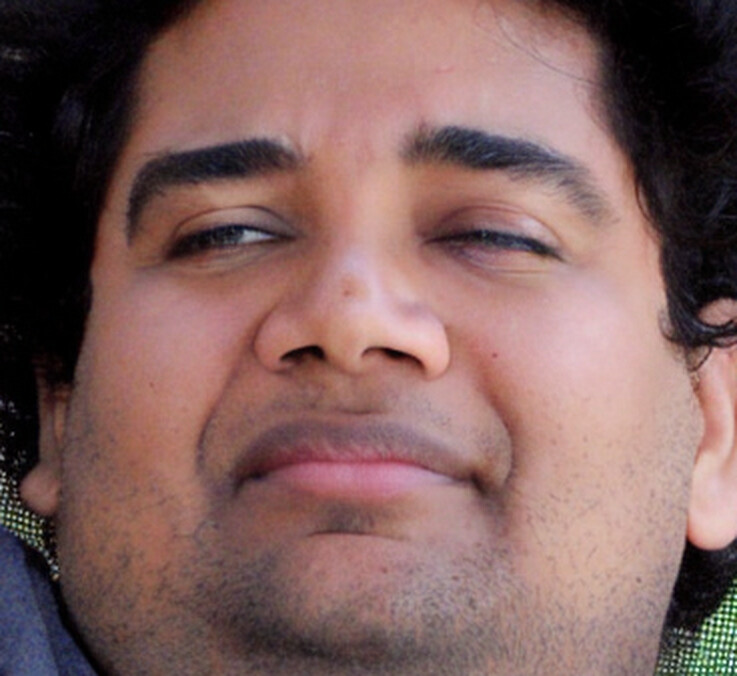} \\
    \includegraphics[width=\linewidth]{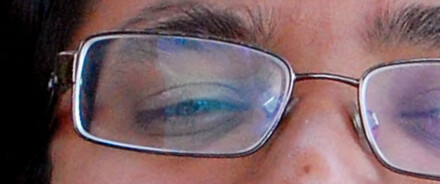} & \includegraphics[width=\linewidth]{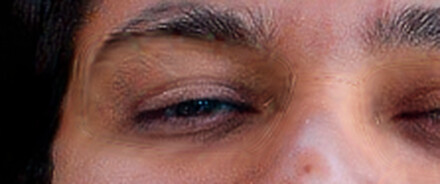} & \includegraphics[width=\linewidth]{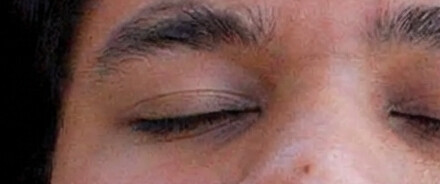} & \includegraphics[width=\linewidth]{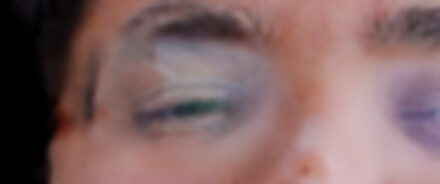} & \includegraphics[width=\linewidth]{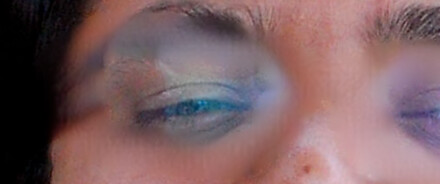} & \includegraphics[width=\linewidth]{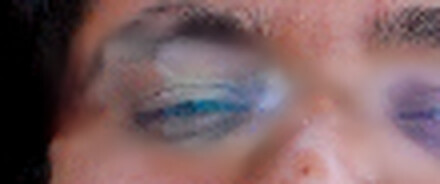} & \includegraphics[width=\linewidth]{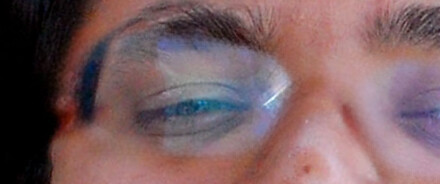} & \includegraphics[width=\linewidth]{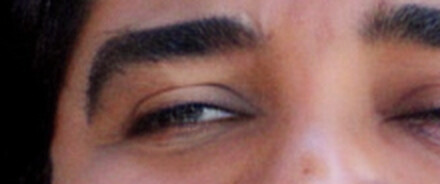} \\[0.2em]
    \includegraphics[width=\linewidth]{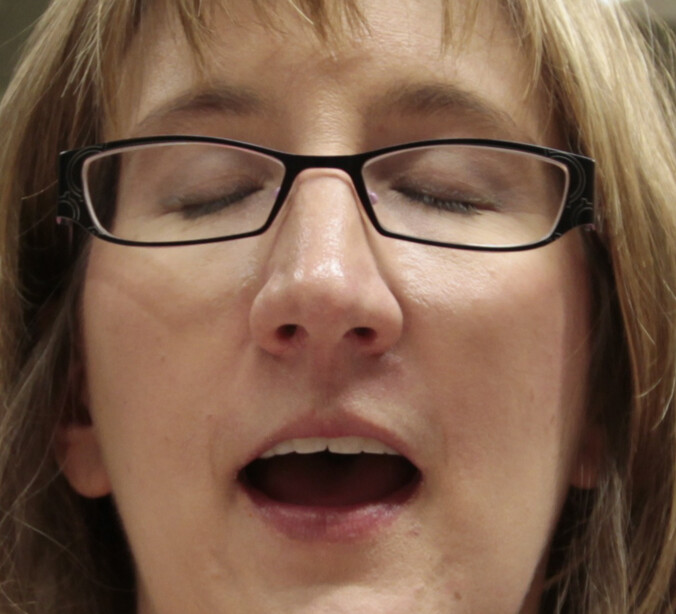} & \includegraphics[width=\linewidth]{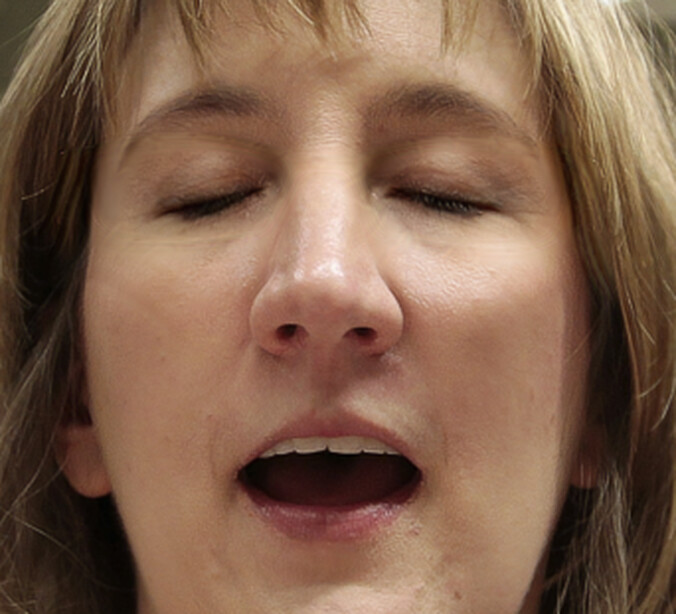} & \includegraphics[width=\linewidth]{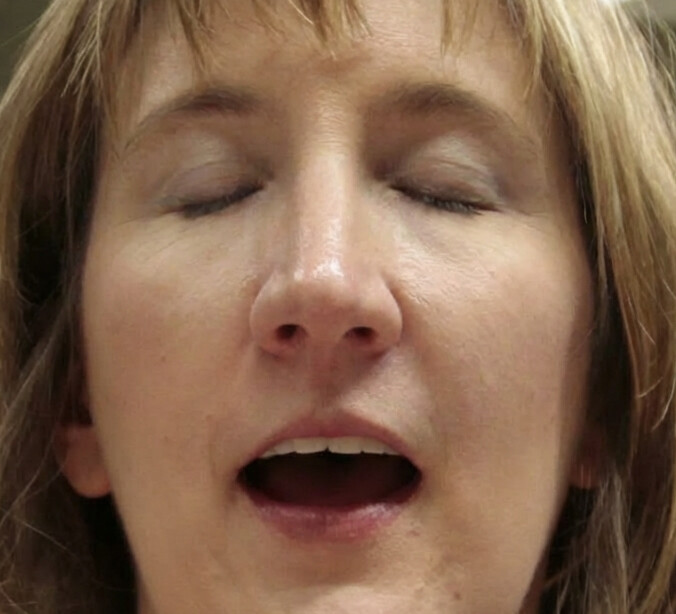} & \includegraphics[width=\linewidth]{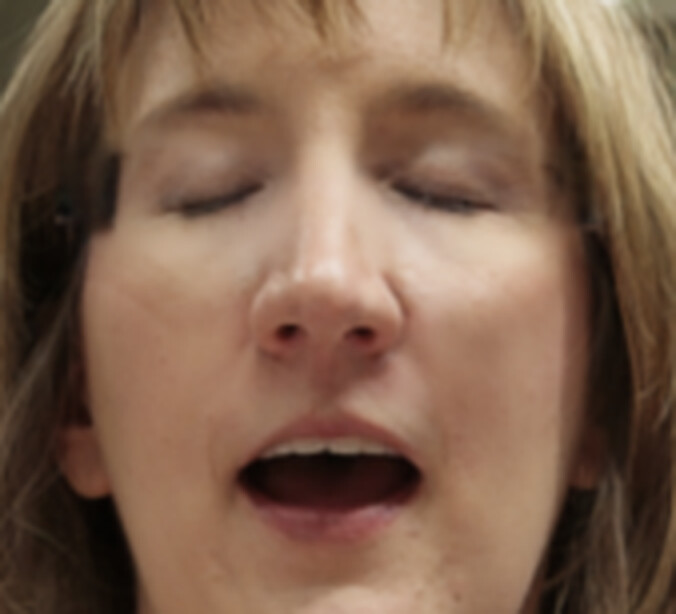} & \includegraphics[width=\linewidth]{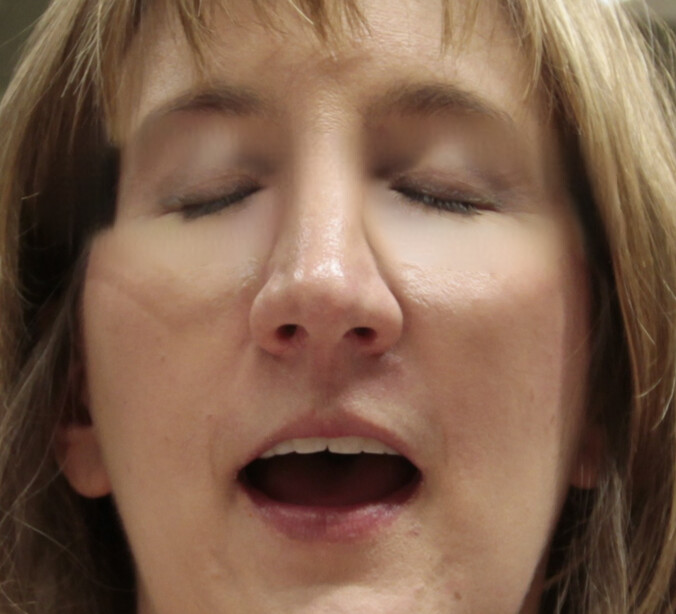} & \includegraphics[width=\linewidth]{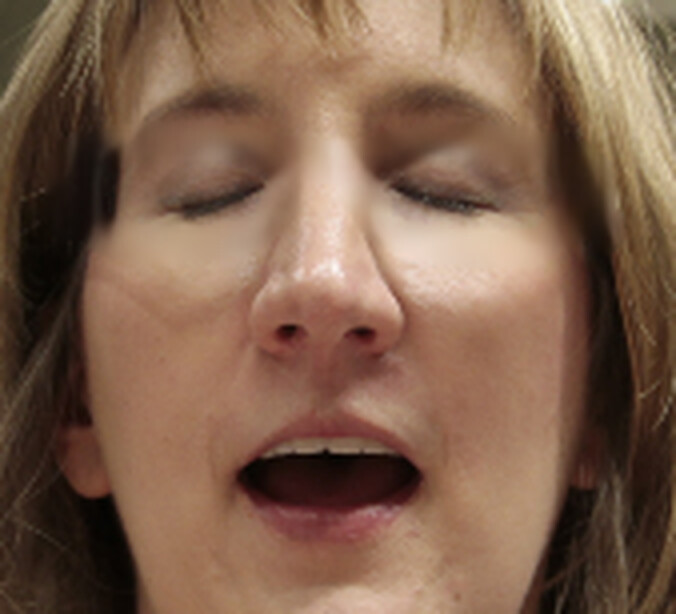} & \includegraphics[width=\linewidth]{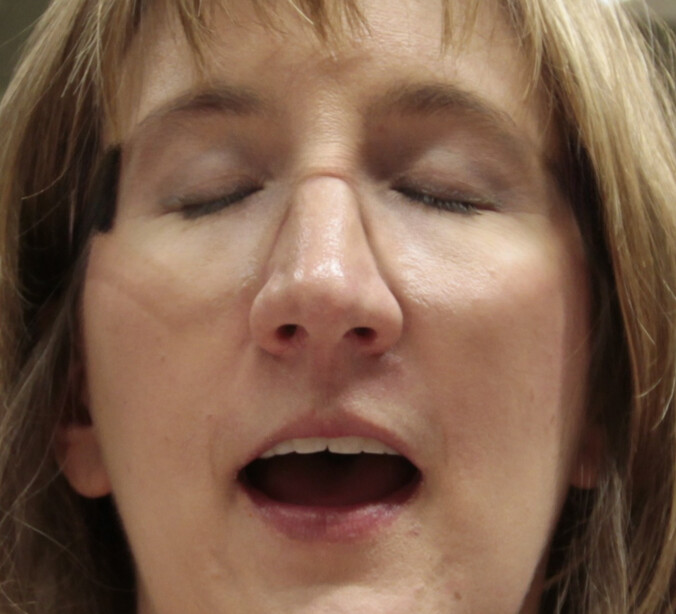} & \includegraphics[width=\linewidth]{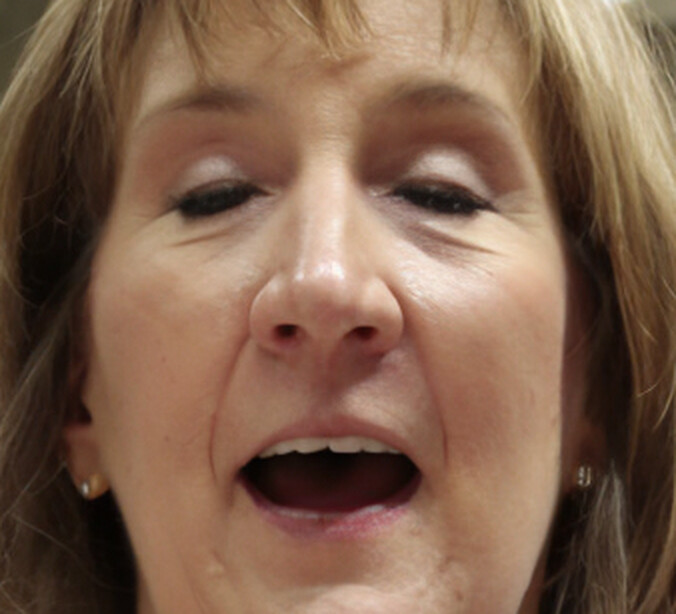} \\
    \includegraphics[width=\linewidth]{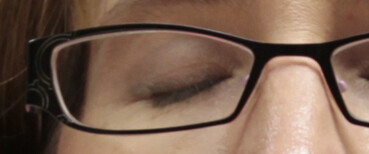} & \includegraphics[width=\linewidth]{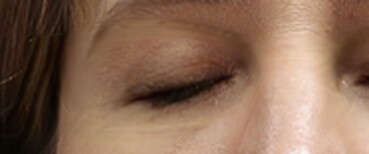} & \includegraphics[width=\linewidth]{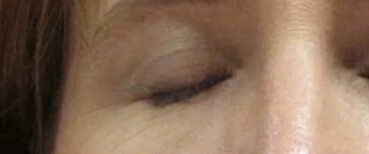} & \includegraphics[width=\linewidth]{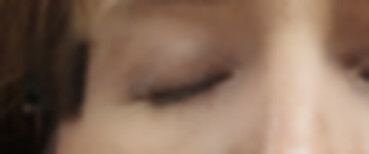} & \includegraphics[width=\linewidth]{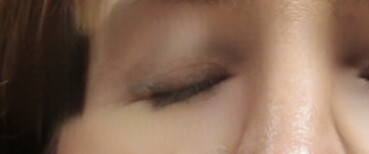} & \includegraphics[width=\linewidth]{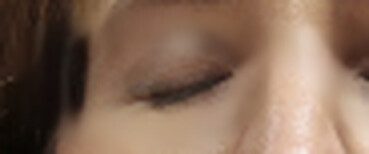} & \includegraphics[width=\linewidth]{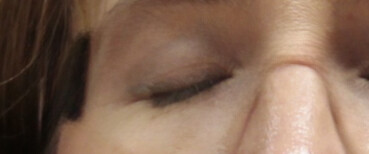} & \includegraphics[width=\linewidth]{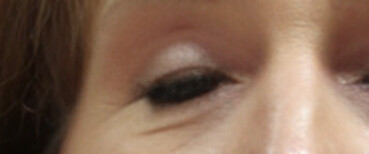} \\[0.2em]
    \includegraphics[width=\linewidth]{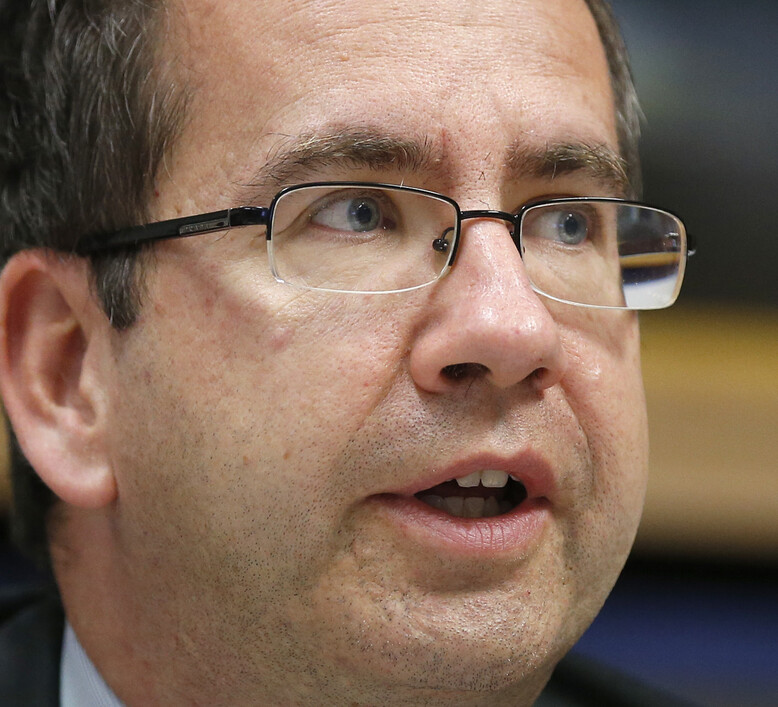} & \includegraphics[width=\linewidth]{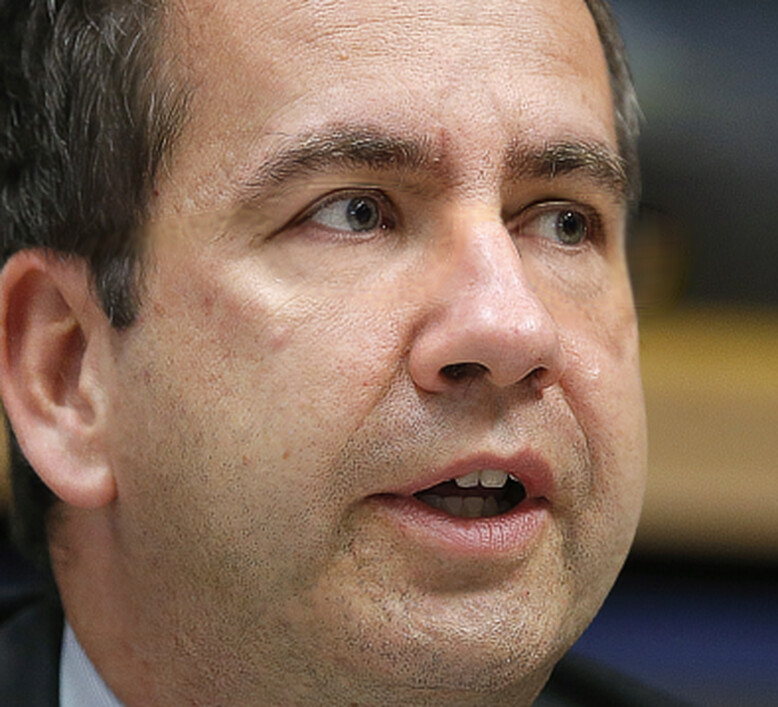} & \includegraphics[width=\linewidth]{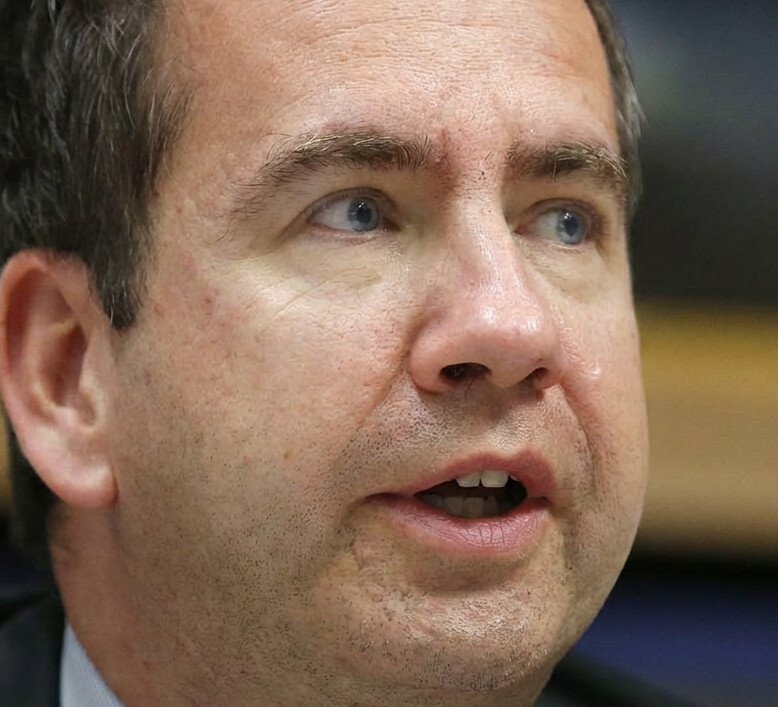} & \includegraphics[width=\linewidth]{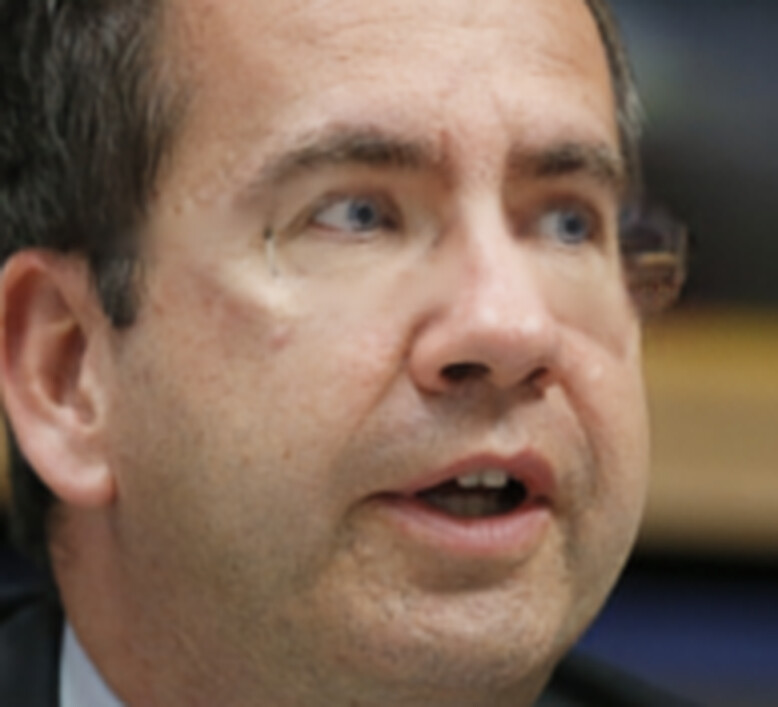} & \includegraphics[width=\linewidth]{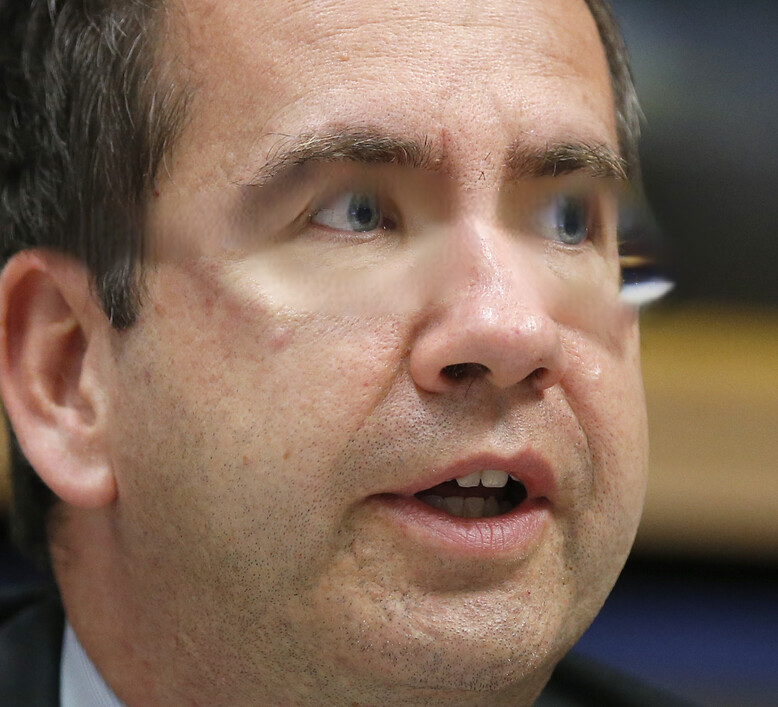} & \includegraphics[width=\linewidth]{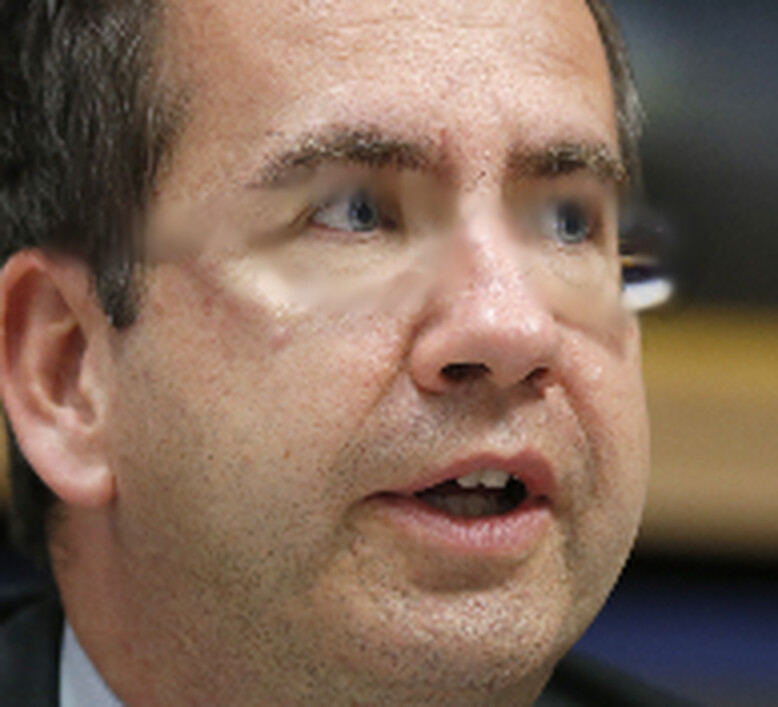} & \includegraphics[width=\linewidth]{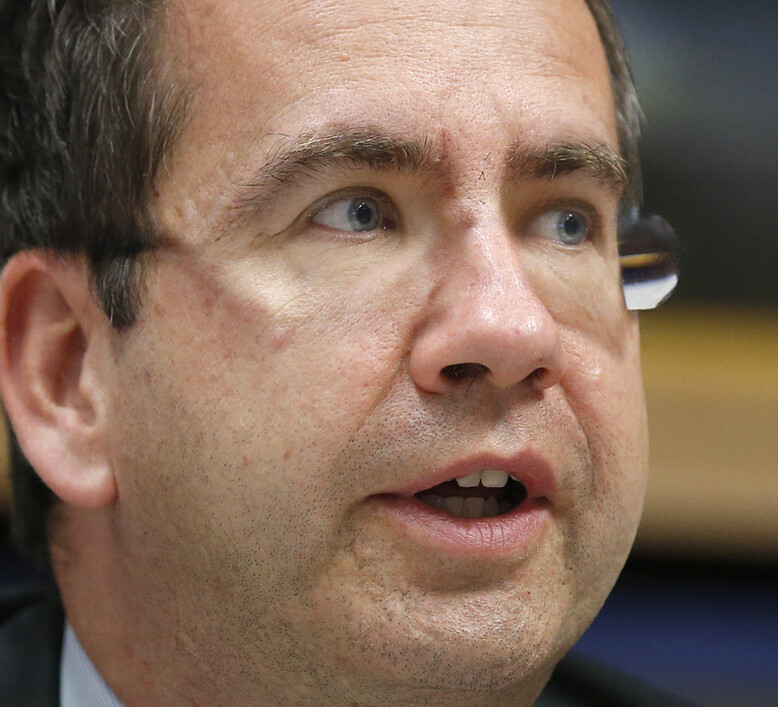} & \includegraphics[width=\linewidth]{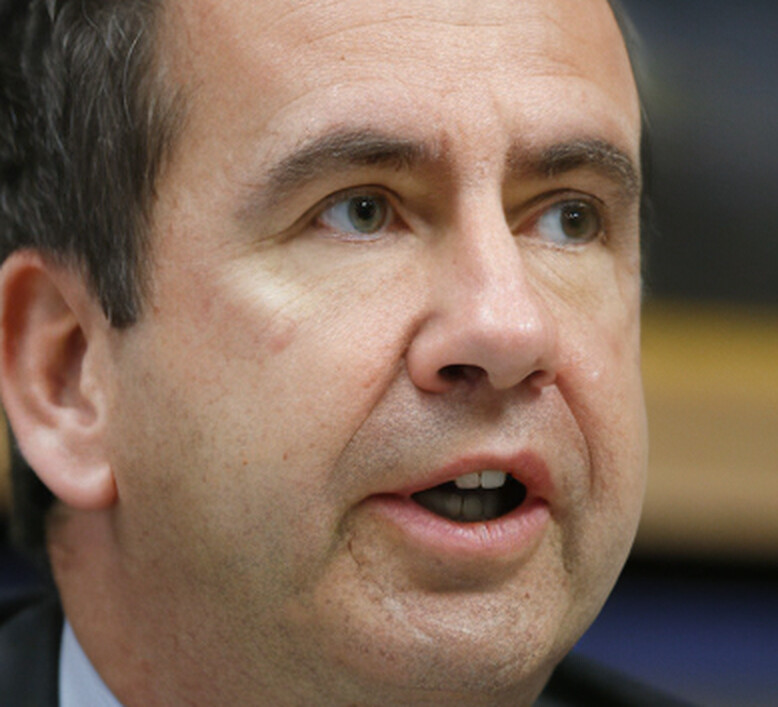} \\
    \includegraphics[width=\linewidth]{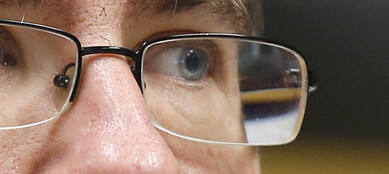} & \includegraphics[width=\linewidth]{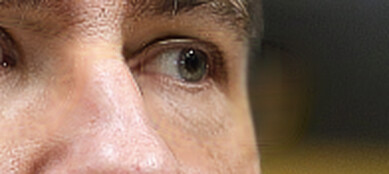} & \includegraphics[width=\linewidth]{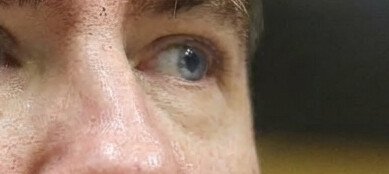} & \includegraphics[width=\linewidth]{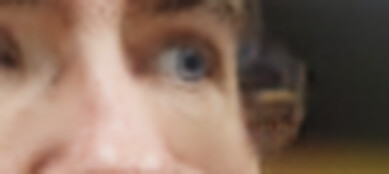} & \includegraphics[width=\linewidth]{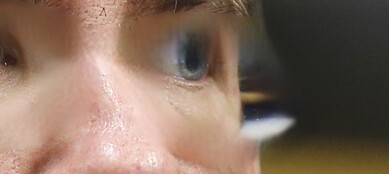} & \includegraphics[width=\linewidth]{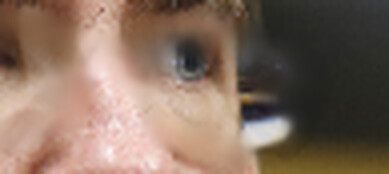} & \includegraphics[width=\linewidth]{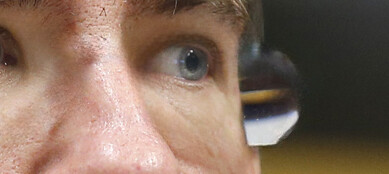} & \includegraphics[width=\linewidth]{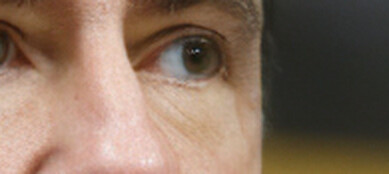} \\
  \end{tabularx}
  \caption{\textbf{Qualitative comparison on high-refraction samples from FFHQ~\cite{karras_style-based_2021}.} Close-up eye regions are shown for each subject. While Nano Banana~\cite{google_gemini_2025} removes reflections, it alters structural details (e.g., gaze shift in row~1, eye closure in row~2). Diffusion-based LEDITS~\cite{tsaban_ledits_2023} exhibits identity drift or fails to remove glasses (row~1). Video inpainting methods~\cite{zeng_learning_2020, zhang_flow-guided_2022, zhou_propainter_2023} often fail to recover underlying facial structures. Our method effectively removes reflections and corrects refraction while preserving fine facial details and maintaining high structural integrity. See Sec.~\ref{sec:ffhq_qualitative_overview} for more details.}
  \label{fig:refraction_qualitative_overview}
\end{figure*}

\emph{Implementation Details.} We implement our framework using JAX/Flax, training the models on high-resolution images resized to $518 \times 518$ pixels. We use the AdamW optimizer with a base learning rate of $3 \times 10^{-4}$ for the \emph{JFSnet} and \emph{Adversarial Critic}. The pre-trained DINOv2 encoder is fine-tuned with a lower learning rate of $1 \times 10^{-5}$ for its last four layers, while the rest of the backbone is kept frozen. Training is performed on TPU/GPU accelerators for approximately 400,000 batch iterations.
For the physics-based augmentation, we sample diopter powers ranging from $-6.0$ to $+1.0$ with an assumed refractive index of $n \approx 1.5$. Since negative diopters (minification) introduce visible background artifacts into the eye area while positive diopters (magnification) are harder to distinguish, we focus on minification in our qualitative analysis (see Fig.~\ref{fig:refraction_qualitative_overview}) and simulation. Specular reflections are simulated via sphere mapping with high-dynamic-range (HDR) emulation and randomized anti-reflective (AR) coating effects. To mitigate aliasing, we compute a Level of Detail (LOD) map based on the Jacobian of the ray divergence, enabling interpolation between sharp and pre-blurred background images.
\emph{Data Augmentation Pipeline.} We use our curated dataset of 1,860 synthetic identities (24,180 image pairs) with on-the-fly augmentation applied during training. By randomizing specular reflection intensity and tint to simulate diverse lens coatings, the model encounters a wide variety of optical distortions while remaining anchored to the eyeglass-free ground truth.
\subsection{Dataset Generation}
We constructed our training dataset using the generative workflow and filtering process detailed in Sec.~\ref{sec:data_curation}.
This process ensures high structural and photometric consistency across the multi-view generative face sets, providing a robust foundation for the subsequent physics-based augmentation. See Supplementary Sec.~\ref{sec:dataset_overview} for a qualitative sample of a single face set.
\subsection{Datasets}\label{sec:datasets}
We utilize the following datasets exclusively for qualitative and quantitative evaluation; neither dataset was used during the training phase of our models.
\textbf{Flickr-Faces-HQ (FFHQ)}~\cite{karras_style-based_2019} consists of 70,000 high-resolution portrait images with significant variation in demographic factors and backgrounds. 
For our evaluation, we manually curated images containing eyewear from this dataset, classifying them into two distinct categories based on ocular visibility: \textbf{(i)} individuals wearing corrective glasses with visible eye regions, and \textbf{(ii)} individuals wearing sunglasses or eyewear with significant darkening and specular reflections. 
This process yielded a refined evaluation set containing 12,163 images with clear glasses and 2,176 images in the sunglasses category. The remaining images in the dataset ($\approx$ 55,000) are utilized as the ``without eyewear'' reference category.
\textbf{CelebV-Text}~\cite{yu_celebv-text_2023} is a large-scale video dataset capturing dynamic facial expressions. For our qualitative evaluation and perceptual study (Sec.~\ref{sec:perceptual_study}), we manually curated a subset of 60 video clips featuring subjects with clearly visible eyeglasses.
\subsection{Baselines}
We evaluate our method against representative approaches across three categories.
For general video editing, we compare against TokenFlow~\cite{geyer_tokenflow_2023}, RAVE~\cite{kara_rave_2024}, and IP-FaceDiff~\cite{anand_ip-facediff_2025}.
In the domain of video inpainting, we include ProPainter~\cite{zhou_propainter_2023}, Flow-Guided Transformer (FGT)~\cite{zhang_flow-guided_2022}, Inpaint Anything~\cite{yu_inpaint_2023}, and STTN~\cite{zeng_learning_2020}.
We also evaluate image-based facial editing methods including Take-Off-Eyeglasses (TOE)~\cite{lyu_portrait_2022}, LEDITS~\cite{tsaban_ledits_2023}, and InstructPix2Pix~\cite{brooks_instructpix2pix_2023}.

Additionally, Runway Gen-4.5 is a private, commercial video generation model with significant inference costs, leading us to restrict its evaluation to the human preference study.
For diffusion-based methods such as TokenFlow, RAVE, LEDITS, and IP-FaceDiff, achieving high-fidelity results is often sensitive to parameter tuning and the stability of the underlying inversion process (e.g., DDIM inversion~\cite{song_denoising_2021}).
As finding viable configurations exceeded the practical scope of our evaluation, we focused our human preference study on the examples where baseline methods achieved competitive scores on the FFHQ dataset.

\subsection{Flickr-Faces-HQ Qualitative Results}\label{sec:ffhq_qualitative_overview}
We present a qualitative comparison on the FFHQ dataset in Fig.~\ref{fig:ffhq_qualitative_overview}. The generative models, such as Nano Banana~\cite{google_gemini_2025}, often reconstruct the eye region with plausible but inaccurate details, leading to subtle identity shifts or gaze changes. 

LEDITS~\cite{tsaban_ledits_2023} struggles to fully remove frame artifacts, whereas our method preserves facial textures and gaze and compensates for the spatial minification of strong refractive lenses (see Fig.~\ref{fig:refraction_visualization}). 
See the Supplementary Sec.~\ref{sec:ffhq_results_extended} for more results and Sec.~\ref{sec:failure_cases} for failure cases.

\begin{figure*}[t]
  \centering
  \renewcommand{\arraystretch}{0}
  \setlength{\tabcolsep}{0pt}
  \begin{tabularx}{\textwidth}{@{} X @{\hskip 5pt} @{} X @{} X @{} X @{} X @{} X @{} X @{} X @{{}}}
    \parbox[b]{\linewidth}{\centering \scriptsize Input Frame} & \parbox[b]{\linewidth}{\centering \scriptsize Our Approach} & \parbox[b]{\linewidth}{\centering \scriptsize Runway Gen-4.5~\cite{runway_ai_inc_runway_2025}} & \parbox[b]{\linewidth}{\centering \scriptsize TOE\\ \cite{lyu_portrait_2022}} & \parbox[b]{\linewidth}{\centering \scriptsize ProPainter\\ \cite{zhou_propainter_2023}} & \parbox[b]{\linewidth}{\centering \scriptsize FGT\\ \cite{zhang_flow-guided_2022}} & \parbox[b]{\linewidth}{\centering \scriptsize STTN\\ \cite{zeng_learning_2020}} & \parbox[b]{\linewidth}{\centering \scriptsize V-LASIK\\ \cite{shalev-arkushin_v-lasik_2025}} \\%
    \includegraphics[width=\linewidth]{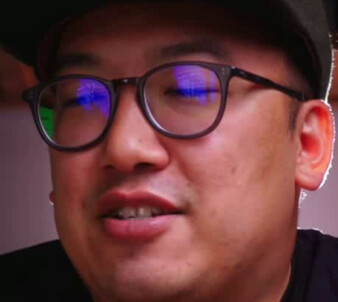} & \includegraphics[width=\linewidth]{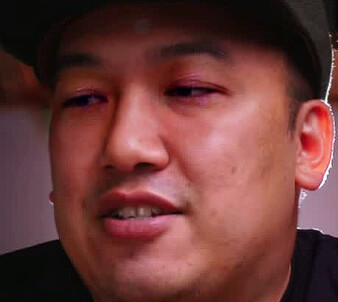} & \includegraphics[width=\linewidth]{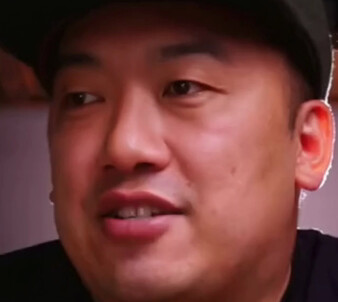} & \includegraphics[width=\linewidth]{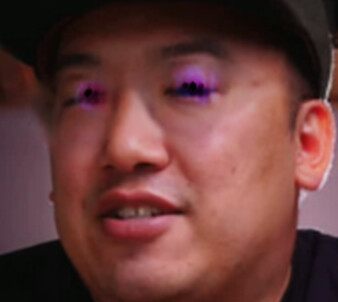} & \includegraphics[width=\linewidth]{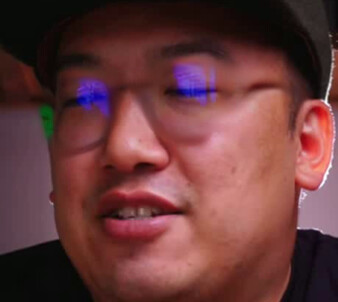} & \includegraphics[width=\linewidth]{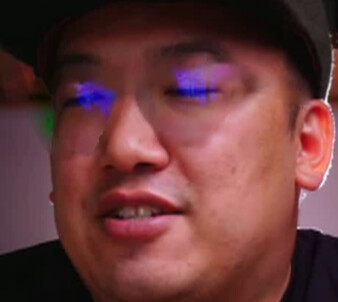} & \includegraphics[width=\linewidth]{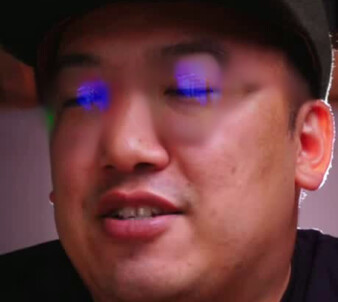} & \parbox{\linewidth}{\centering \scriptsize \vspace{-4em} N/A}\vspace{-0.2em} \\
    \includegraphics[width=\linewidth]{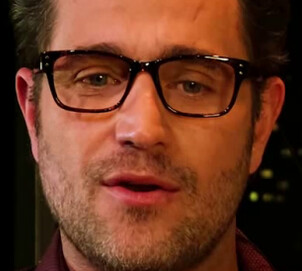} & \includegraphics[width=\linewidth]{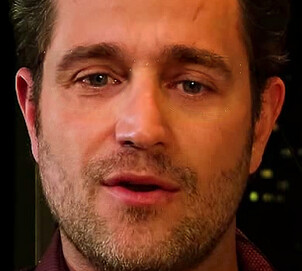} & \includegraphics[width=\linewidth]{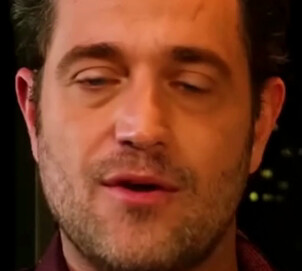} & \includegraphics[width=\linewidth]{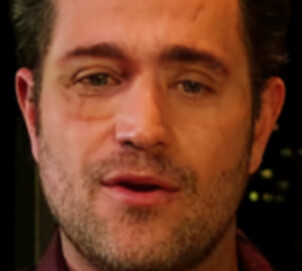} & \includegraphics[width=\linewidth]{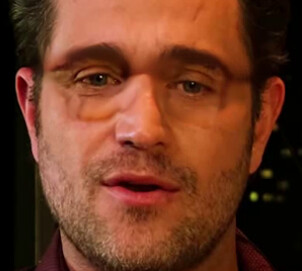} & \includegraphics[width=\linewidth]{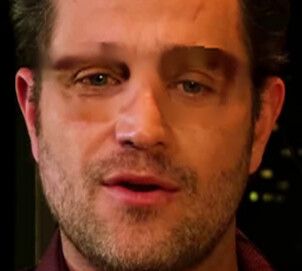} & \includegraphics[width=\linewidth]{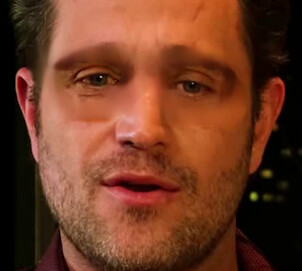} & \parbox{\linewidth}{\centering \scriptsize \vspace{-4em} N/A}\vspace{-0.2em} \\
    \includegraphics[width=\linewidth]{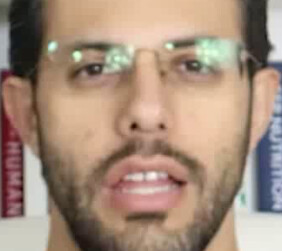} & \includegraphics[width=\linewidth]{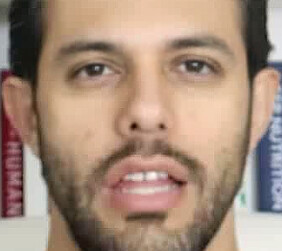} & \includegraphics[width=\linewidth]{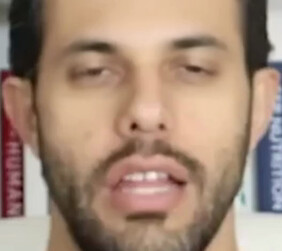} & \includegraphics[width=\linewidth]{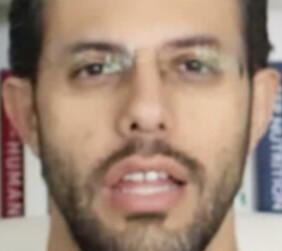} & \includegraphics[width=\linewidth]{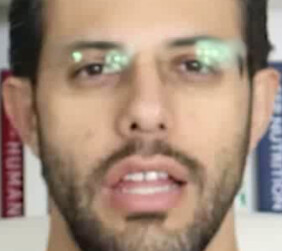} & \includegraphics[width=\linewidth]{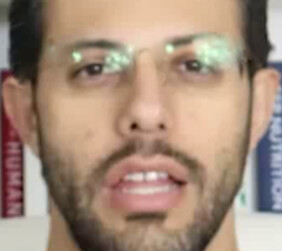} & \includegraphics[width=\linewidth]{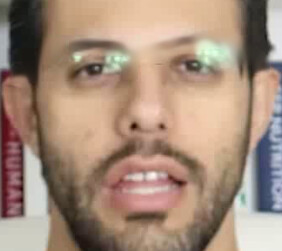} & \parbox{\linewidth}{\centering \scriptsize \vspace{-4em} N/A}\vspace{-0.2em} \\
    \includegraphics[width=\linewidth]{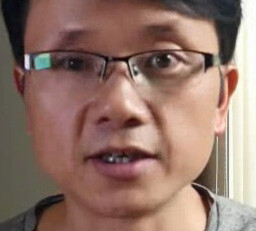} & \includegraphics[width=\linewidth]{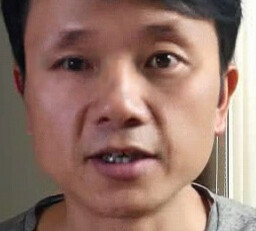} & \includegraphics[width=\linewidth]{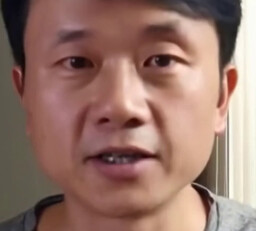} & \includegraphics[width=\linewidth]{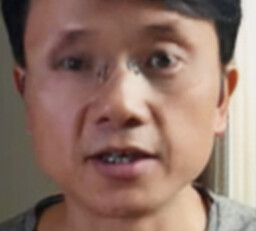} & \includegraphics[width=\linewidth]{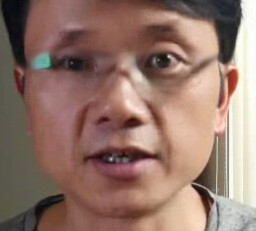} & \includegraphics[width=\linewidth]{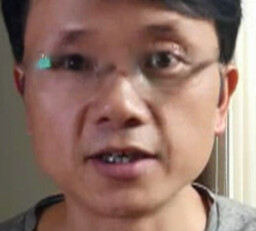} & \includegraphics[width=\linewidth]{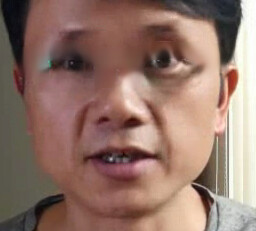} & \includegraphics[width=\linewidth]{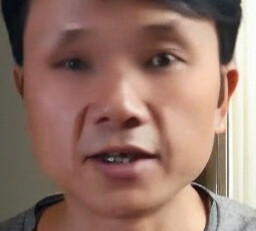} \\
    \includegraphics[width=\linewidth]{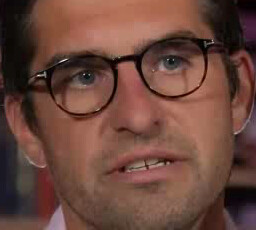} & \includegraphics[width=\linewidth]{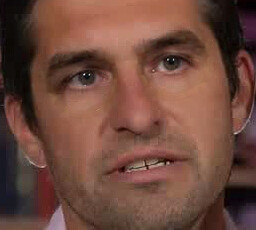} & \includegraphics[width=\linewidth]{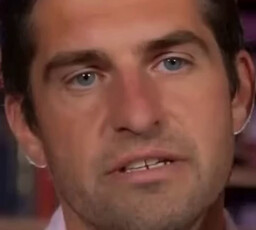} & \includegraphics[width=\linewidth]{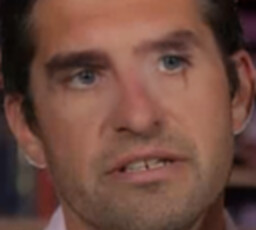} & \includegraphics[width=\linewidth]{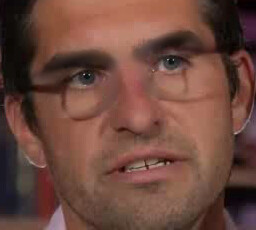} & \includegraphics[width=\linewidth]{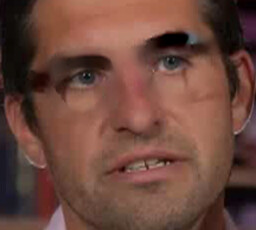} & \includegraphics[width=\linewidth]{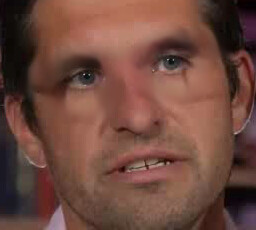} & \includegraphics[width=\linewidth]{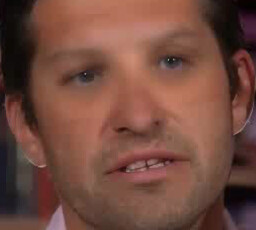} \\
    \includegraphics[width=\linewidth]{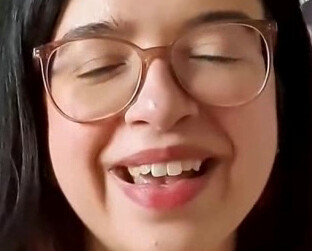} & \includegraphics[width=\linewidth]{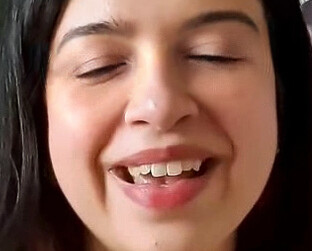} & \includegraphics[width=\linewidth]{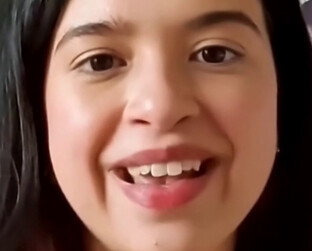} & \includegraphics[width=\linewidth]{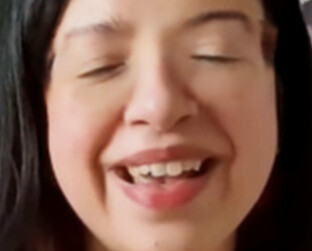} & \includegraphics[width=\linewidth]{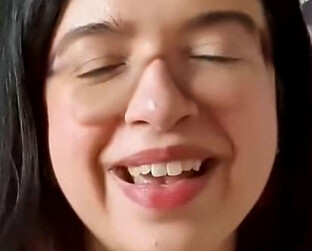} & \includegraphics[width=\linewidth]{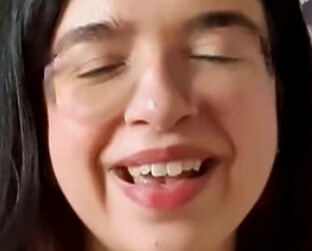} & \includegraphics[width=\linewidth]{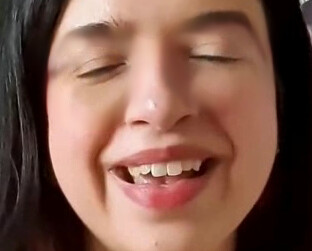} & \includegraphics[width=\linewidth]{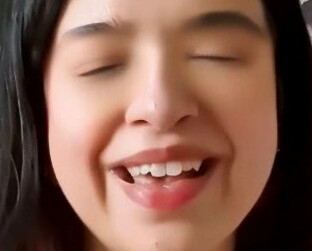} \\%
  \end{tabularx}
  \caption{\textbf{Qualitative comparison on CelebV-Text~\cite{yu_celebv-text_2023}.} Runway Gen-4.5 removes reflections but changes the gaze and eye identity. V-LASIK causes blur and changes eye shape. TOE struggles with strong reflections. ProPainter, FGT, and STTN only perform inpainting, leaving behind shadows, reflections, or parts of the glasses. Our Joint Feature-Spatial network preserves gaze and fine facial details. See Sec.~\ref{sec:celebv_qualitative_overview}.}
  \label{fig:celebv_qualitative_overview}
\end{figure*}

\subsection{CelebV-Text Qualitative Results}\label{sec:celebv_qualitative_overview}
The qualitative results for video sequences are shown in Fig.~\ref{fig:celebv_qualitative_overview}. Large-scale video-to-video model Runway Gen-4.5~\cite{runway_ai_inc_runway_2025} removes reflections, but often fails to preserve the subject's gaze and eye identity. Video inpainting methods ProPainter~\cite{zhou_propainter_2023} and FGT~\cite{zhang_flow-guided_2022} are primarily designed for object removal and lack the facial priors needed to handle the complex shadows and reflections cast by eyeglasses. These methods often leave behind ghosting artifacts. By contrast, our method leverages its physics-based training, providing temporally stable results that maintain the geometric integrity of the face. Full videos are available in the supplement.

\subsection{Perceptual Study}\label{sec:perceptual_study}
To further evaluate our method, we conducted a perceptual study consisting of two surveys focused on video and image glasses removal.
For the video survey, we utilized 60 clips from the CelebV-Text dataset, while the image survey used 60 high-resolution portraits from the FFHQ dataset.
We invited 37 participants to compare our results against 4 other representative methods, resulting in a total of 5 methods per question.
In each survey, participants were asked to rank the methods based on specific criteria derived from the qualitative characteristics of the task. For the video survey, the evaluation axes included gaze consistency, temporal stability, and overall restoration quality. The image survey focused on gaze preservation, removal completeness, identity preservation, sharpness, and overall removal quality. These criteria allow for a granular assessment of how well each method handles both the physical restoration of ocular details and the generative consistency of the face. 
The collected preferences are visualized in the heatmap of Fig.~\ref{fig:survey_heatmap}.
Our method is consistently preferred across both temporal and spatial dimensions. While Nano Banana achieves high scores for completeness due to its powerful generative prior, it is consistently outperformed in gaze and identity preservation where its unconstrained outputs diverge from the subject's original features.
Temporal metrics are in the Supplementary~Sec.~\ref{sec:temporal_eval}.

\begin{figure*}[t]
  \centering
  \definecolor{heat0}{RGB}{165,0,38}\definecolor{heat50}{RGB}{255,255,191}\definecolor{heat100}{RGB}{0,104,55}
  \renewcommand{\arraystretch}{1.1}\setlength{\tabcolsep}{2pt}
  \begin{tabularx}{\textwidth}{X c c c @{\hskip 0.5em} c @{\hskip 0.5em} c c c c c }
       & \multicolumn{3}{c}{\footnotesize \textbf{Video Survey}} & & \multicolumn{5}{c}{\footnotesize \textbf{Image Survey}} \\ [0.2em]
 & \multicolumn{3}{c}{\scriptsize Our method is preferred in} & & \multicolumn{5}{c}{\scriptsize Our method is preferred in} \\ [0.5em]
      \raggedleft \scriptsize Nano Banana~\cite{google_gemini_2025} & \cellcolor[HTML]{0F713F}\textcolor{white}{\tiny 97\%} & \cellcolor[HTML]{006837}\textcolor{white}{\tiny 100\%} & \cellcolor[HTML]{006837}\textcolor{white}{\tiny 100\%} &  & \cellcolor[HTML]{60A16A}\textcolor{white}{\tiny 81\%} & \cellcolor[HTML]{F0D6A6}\textcolor{black}{\tiny 42\%} & \cellcolor[HTML]{7FB37B}\textcolor{black}{\tiny 75\%} & \cellcolor[HTML]{E5EFB1}\textcolor{black}{\tiny 55\%} & \cellcolor[HTML]{84B67D}\textcolor{black}{\tiny 74\%} \\
      \raggedleft \scriptsize LEDITS~\cite{tsaban_ledits_2023} & \cellcolor[HTML]{006837}\textcolor{white}{\tiny 100\%} & \cellcolor[HTML]{006837}\textcolor{white}{\tiny 100\%} & \cellcolor[HTML]{147441}\textcolor{white}{\tiny 96\%} &  & \cellcolor[HTML]{338652}\textcolor{white}{\tiny 90\%} & \cellcolor[HTML]{4C955F}\textcolor{white}{\tiny 85\%} & \cellcolor[HTML]{0A6E3C}\textcolor{white}{\tiny 98\%} & \cellcolor[HTML]{338652}\textcolor{white}{\tiny 90\%} & \cellcolor[HTML]{147441}\textcolor{white}{\tiny 96\%} \\
      \raggedleft \scriptsize FGT~\cite{zhang_flow-guided_2022} & \cellcolor[HTML]{569B65}\textcolor{white}{\tiny 83\%} & \cellcolor[HTML]{147441}\textcolor{white}{\tiny 96\%} & \cellcolor[HTML]{006837}\textcolor{white}{\tiny 100\%} &  & \cellcolor[HTML]{66A46D}\textcolor{white}{\tiny 80\%} & \cellcolor[HTML]{28804C}\textcolor{white}{\tiny 92\%} & \cellcolor[HTML]{5B9E67}\textcolor{white}{\tiny 82\%} & \cellcolor[HTML]{0A6E3C}\textcolor{white}{\tiny 98\%} & \cellcolor[HTML]{0F713F}\textcolor{white}{\tiny 97\%} \\
      \raggedleft \scriptsize TOE~\cite{lyu_portrait_2022} & \cellcolor[HTML]{2D834F}\textcolor{white}{\tiny 91\%} & \cellcolor[HTML]{0F713F}\textcolor{white}{\tiny 97\%} & \cellcolor[HTML]{0A6E3C}\textcolor{white}{\tiny 98\%} &  & \cellcolor[HTML]{4C955F}\textcolor{white}{\tiny 85\%} & \cellcolor[HTML]{147441}\textcolor{white}{\tiny 96\%} & \cellcolor[HTML]{388954}\textcolor{white}{\tiny 89\%} & \cellcolor[HTML]{006837}\textcolor{white}{\tiny 100\%} & \cellcolor[HTML]{056B39}\textcolor{white}{\tiny 99\%} \\
      \raggedleft \scriptsize V-LASIK~\cite{shalev-arkushin_v-lasik_2025} & \cellcolor[HTML]{3D8C57}\textcolor{white}{\tiny 88\%} & \cellcolor[HTML]{3D8C57}\textcolor{white}{\tiny 88\%} & \cellcolor[HTML]{006837}\textcolor{white}{\tiny 100\%} &  & \cellcolor[gray]{0.9}\textcolor{gray}{\tiny N/A} & \cellcolor[gray]{0.9}\textcolor{gray}{\tiny N/A} & \cellcolor[gray]{0.9}\textcolor{gray}{\tiny N/A} & \cellcolor[gray]{0.9}\textcolor{gray}{\tiny N/A} & \cellcolor[gray]{0.9}\textcolor{gray}{\tiny N/A} \\
      \raggedleft \scriptsize STTN~\cite{zeng_learning_2020} & \cellcolor[HTML]{569B65}\textcolor{white}{\tiny 83\%} & \cellcolor[HTML]{428F5A}\textcolor{white}{\tiny 87\%} & \cellcolor[HTML]{006837}\textcolor{white}{\tiny 100\%} &  & \cellcolor[HTML]{60A16A}\textcolor{white}{\tiny 81\%} & \cellcolor[HTML]{006837}\textcolor{white}{\tiny 100\%} & \cellcolor[HTML]{75AD75}\textcolor{black}{\tiny 77\%} & \cellcolor[HTML]{28804C}\textcolor{white}{\tiny 92\%} & \cellcolor[HTML]{0A6E3C}\textcolor{white}{\tiny 98\%} \\
      \raggedleft \scriptsize ProPainter~\cite{zhou_propainter_2023} & \cellcolor[HTML]{7AB078}\textcolor{black}{\tiny 76\%} & \cellcolor[HTML]{7AB078}\textcolor{black}{\tiny 76\%} & \cellcolor[HTML]{147441}\textcolor{white}{\tiny 96\%} &  & \cellcolor[HTML]{519862}\textcolor{white}{\tiny 84\%} & \cellcolor[HTML]{237D4A}\textcolor{white}{\tiny 93\%} & \cellcolor[HTML]{519862}\textcolor{white}{\tiny 84\%} & \cellcolor[HTML]{147441}\textcolor{white}{\tiny 96\%} & \cellcolor[HTML]{0A6E3C}\textcolor{white}{\tiny 98\%} \\
      \raggedleft \scriptsize Runway Gen-4.5~\cite{runway_ai_inc_runway_2025} & \cellcolor[HTML]{84B67D}\textcolor{black}{\tiny 74\%} & \cellcolor[HTML]{D1E3A6}\textcolor{black}{\tiny 59\%} & \cellcolor[HTML]{7FB37B}\textcolor{black}{\tiny 75\%} &  & \cellcolor[gray]{0.9}\textcolor{gray}{\tiny N/A} & \cellcolor[gray]{0.9}\textcolor{gray}{\tiny N/A} & \cellcolor[gray]{0.9}\textcolor{gray}{\tiny N/A} & \cellcolor[gray]{0.9}\textcolor{gray}{\tiny N/A} & \cellcolor[gray]{0.9}\textcolor{gray}{\tiny N/A} \\
      & \multicolumn{3}{c}{\scriptsize when participants consider} & & \multicolumn{5}{c}{\scriptsize when participants consider} \\ [0.2em]
        & {\parbox{3.1em}{\centering \tiny Gaze\\\hspace{-0.1em}consistency}} & \parbox{3.1em}{\centering \tiny Temporal\\\hspace{-0.1em}consistency} & {\parbox{3.1em}{\centering \tiny Overall\\quality}} &  & \parbox{3.1em}{\centering \tiny Gaze\\\hspace{-0.1em}preservation} & \parbox{3.1em}{\centering \tiny Removal\\\hspace{-0.1em}completeness} & \parbox{3.1em}{\centering \tiny Identity\\\hspace{-0.1em}preservation} & \parbox{3.1em}{\centering \tiny Sharpness} & \parbox{3.1em}{\centering \tiny Overall\\quality} \\[0.7em]
  \end{tabularx}
  \caption{\textbf{Preference study on CelebV-Text~\cite{yu_celebv-text_2023} and FFHQ~\cite{karras_style-based_2021}.} Our approach is consistently preferred across both video and image surveys. The heatmap shows the percentage of participants who preferred our result over the baseline for each evaluation criterion. See Sec.~\ref{sec:perceptual_study} for more details.}
  \label{fig:survey_heatmap}
  \end{figure*}

\subsection{Quantitative Evaluation}\label{sec:quantitative_evaluation}
We assess the performance of the methods across several dimensions --- gaze preservation (head pose, pupil position, and ocular structure), visual realism, structural fidelity, and computational efficiency.

Gaze preservation is quantified via the \textbf{Pupil Displacement error} (in px), which measures the spatial shift of iris centers between the input and restored output. To avoid evaluation bias, we detect pupils using the \emph{MediaPipe FaceMesh} model~\cite{kartynnik_real-time_2019}, which is distinct from the models used in our data curation pipeline. We calculate the Euclidean distance in normalized coordinates and scale the average displacement by 512 for readability. Note that while physical refraction correction necessitates a minor spatial shift of the ocular features, a low displacement indicates that the model avoids the significant gaze drift common in generative baselines. Inpaint Anything~\cite{yu_inpaint_2023} achieves a competitive pupil error as it ideally avoids ocular modification, but is prone to generating artifacts when segmentation fails. Conversely, Nano Banana~\cite{google_gemini_2025} suffers from significant gaze drift despite serving as our primary data source.

To evaluate the visual realism and quality of the generated results, we employ the \emph{Fréchet Inception Distance} (\textbf{FID}), which measures the statistical distance between the feature distributions of real and synthesized images to capture distribution-level fidelity.

\begin{table}[t]
\centering
\setlength{\tabcolsep}{5.3pt}
\caption{\textbf{Quantitative evaluation on the Flickr-Faces-HQ (FFHQ) dataset~\cite{karras_style-based_2019}.} We compare our proposed approach against several state-of-the-art image-based, video editing, and video inpainting methods using the Fréchet Inception Distance (FID) $\downarrow$, Pupil Displacement $\downarrow$, landmark L2 error (L2) $\downarrow$, and inference speed (FPS) $\uparrow$. Best results in \textbf{bold}. \textsuperscript{$\dagger$ } }
\begin{tabular}{l r r @{\,$\pm$\,} l r @{\,$\pm$\,} l r}
\toprule
Method & FID $\downarrow$ & \multicolumn{2}{c}{Pupil (px) $\downarrow$} & \multicolumn{2}{c}{L2 (px) $\downarrow$} & FPS $\uparrow$ \\
\midrule
Nano Banana \cite{google_gemini_2025} & $0.389$ & $2.854$ & $2.387$ & $0.699$ & $0.462$ & $1.29$ \\
Inpaint Anything \cite{yu_inpaint_2023} & $0.391$ & $2.259$ & $2.295$ & $0.664$ & $0.902$ & $1.84$ \\
InstructPix2Pix \cite{brooks_instructpix2pix_2023} & $1.089$ & $5.541$ & $6.522$ & $8.038$ & $18.293$ & $0.05$ \\
LEDITS \cite{tsaban_ledits_2023} & $1.953$ & $3.905$ & $2.999$ & $1.527$ & $2.437$ & $0.08$ \\
Take-Off-Eyeglasses \cite{lyu_portrait_2022} & $3.256$ & $2.593$ & $1.696$ & $0.877$ & $0.773$ & $13.31$ \\
\midrule
IP-FaceDiff \cite{anand_ip-facediff_2025} & $3.832$ & $17.607$ & $8.718$ & $21.257$ & $27.171$ & $0.04$ \\
RAVE \cite{kara_rave_2024} & $11.559$ & $3.577$ & $2.373$ & $2.356$ & $2.773$ & $0.16$ \\
TokenFlow \cite{geyer_tokenflow_2023} & $2.430$ & $3.689$ & $2.812$ & $1.905$ & $1.678$ & $0.04$ \\
\midrule
Flow-Guided Transformer \cite{zhang_flow-guided_2022} & $0.569$ & $3.078$ & $2.714$ & $0.965$ & $0.952$ & $1.28$ \\
ProPainter \cite{zhou_propainter_2023} & $0.408$ & $2.684$ & $2.838$ & $0.947$ & $1.095$ & $8.77$ \\
\midrule
Ours & $\mathbf{0.379}$ & $\mathbf{2.249}$ & $\mathbf{1.957}$ & $\mathbf{0.632}$ & $\mathbf{0.271}$ & $\mathbf{27.68}$ \\
\bottomrule
\end{tabular}\label{tab:ffhq_results}
\end{table}

\let\thefootnote\relax\footnotetext{\scriptsize \textsuperscript{\textdagger}Authors of V-LASIK~\cite{shalev-arkushin_v-lasik_2025} do not provide code and did not respond to our queries for evaluation; our comparison with their method is limited to the videos available on their project website.}

Structural and geometric stability is measured by the \textbf{Landmark L2 Error} (L2), the Euclidean distance between facial landmarks in the input images (with glasses) and the corresponding landmarks in the restored outputs. This metric ensures that the restoration process does not introduce unwanted spatial distortions or identity-altering pose shifts. We empirically found that ArcFace~\cite{deng_arcface_2019} does not perform well for identity preservation analysis in our case
(see 
Supp. Sec.~\ref{sec:arcface_expected_drop}).
Finally, we report the \emph{Inference Speed} in frames per second (FPS) to evaluate the practical utility of each method for video applications. 
We use the results of our perceptual study (Sec.~\ref{sec:perceptual_study}) to measure temporal stability.

\paragraph{Results on FFHQ.}
We evaluate our method on the curated FFHQ evaluation set, primarily focusing on the clear glasses category where ocular details remain partially visible.
As shown in Table~\ref{tab:ffhq_results}, our approach achieves an FID of 0.379 and a landmark L2 error of 0.632.
For the FID evaluation, we take the real images from the curated FFHQ ``clear glasses'' subset (12,163 images), resize them to $518 \times 518$, run our removal pipeline, and compare the resulting distribution of restored images against a size-matched subset of the FFHQ ``without eyewear'' category.
This distribution-level comparison validates that our method correctly restores facial regions to a state statistically indistinguishable from natural portraits without eyewear.
The sunglasses category is in the Supplementary Sec.~\ref{sec:ffhq_full_results}.
Paired metrics and retrained baselines in the Supplementary~Sec.~\ref{sec:retrained_baselines}.
\subsection{Architectural Ablation}\label{sec:arch_ablation}
We compare the JFSnet configuration against standard baseline architectures to validate our choice of combining semantic and spatial blocks. 
Our evaluation includes a pure Vision Transformer (ViT-L/14), a standard U-Net with a ResNet-34 backbone, and the proposed JFSnet (see the Supplementary Sec.~\ref{sec:architectural_ablation_supp}).

\subsection{Ablation Study}
To validate our design choices, we investigate the impact of the core components of our pipeline, as summarized in the Supplementary Table~\ref{tab:ablation_study}.

Our results demonstrate that both reflection and refraction simulations are critical for achieving low FID scores; removing either component leads to a degradation.
While the temporal consistency loss ($\lambda_{\text{temp}}$) yields the same FID score for static images, it is included for its role in temporal stability; the combination of $\lambda_{\text{temp}}$, rigorous dataset filtering, and ARAP warping results in an FID of 0.379. The dataset filtering process---which includes $L_1$ thresholding and hard-sample mining---ensures the quality of the generative pairs, while the ARAP warping (see Supplementary Sec.~\ref{sec:arap}) accounts for residual pose shifts to maintain pixel-perfect alignment.
Unfreezing the last four layers of the DINOv2 encoder provides a measurable boost in performance.
Finally, the localized eye-specific losses ($\lambda_{\text{eye-pix}}$ and $\lambda_{\text{eye-perc}}$) enforce recovering details in the eye region.
\section{Conclusion}
\label{sec:conclusion}

In this paper, we presented a physics-grounded generative-to-deterministic transfer framework for consistent eyeglasses removal from video, bridging the gap between large-scale generative priors and physical restoration.
By training our \textit{JFSnet} architecture on a generative-to-physical synthesis pipeline, we achieved state-of-the-art visual fidelity (FID 0.379) while maintaining real-time performance at 27.68 FPS.
Our approach leverages multi-level feature fusion and enforced translation equivariance to preserve gaze and identity across dynamic sequences, significantly outperforming existing baselines in perceptual studies.

\section*{Acknowledgments} The research reported in this paper has been partly funded by BMK, BMAW, and the State of Upper Austria in the frame of the SCCH competence center INTEGRATE [Project 1.6 TFI (Transferable Intelligence)] part of the FFG COMET Competence Centers for Excellent Technologies Programme, and by the Czech Technical University in Prague grant No. SGS26/074/OHK3/1T/13.


\bibliographystyle{splncs04}
\bibliography{references}

\clearpage
\beginsupplement
\section{Supplementary}

\subsection{Dataset Overview}\label{sec:dataset_overview}
\begin{figure*}[!b]
  \centering
  \renewcommand{\arraystretch}{0}
  \setlength{\tabcolsep}{0pt}
  \begin{tabularx}{\textwidth}{@{} X X X X X @{}}
    \includegraphics[width=\linewidth]{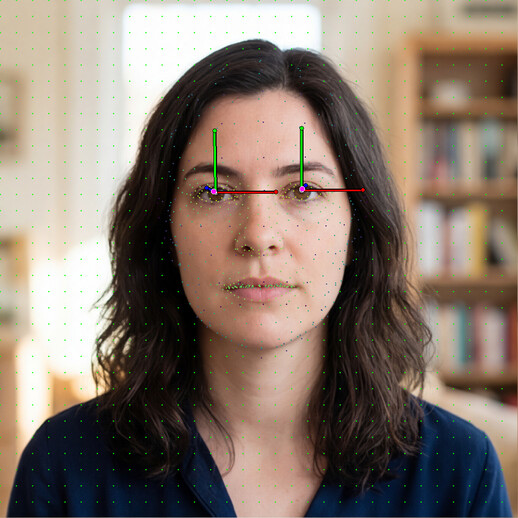} & \includegraphics[width=\linewidth]{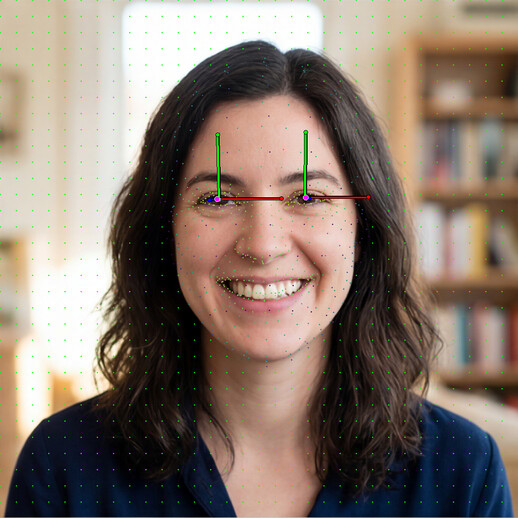} & \includegraphics[width=\linewidth]{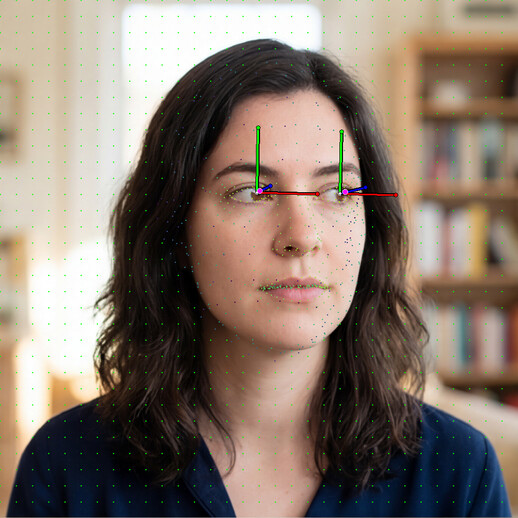} & \includegraphics[width=\linewidth]{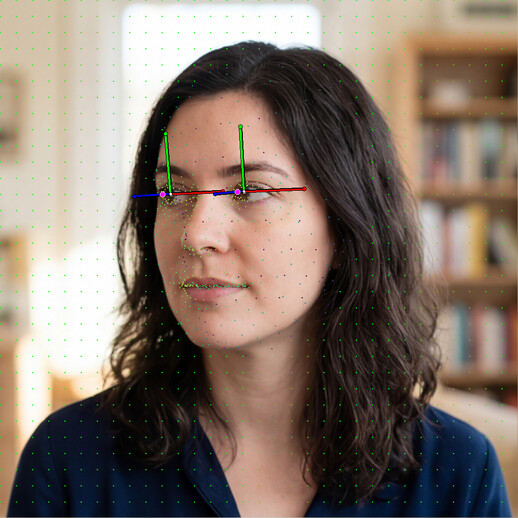} & \includegraphics[width=\linewidth]{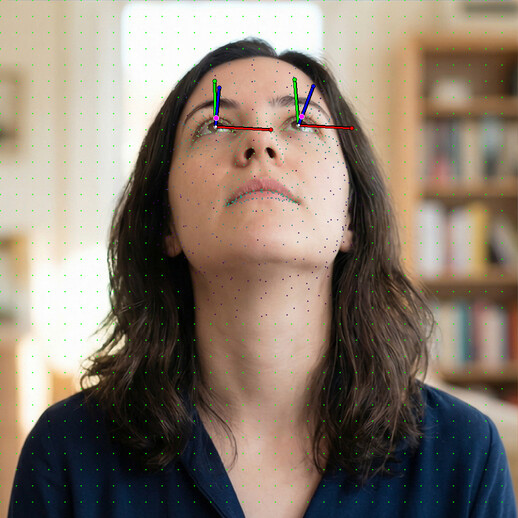} \\%
    \includegraphics[width=\linewidth]{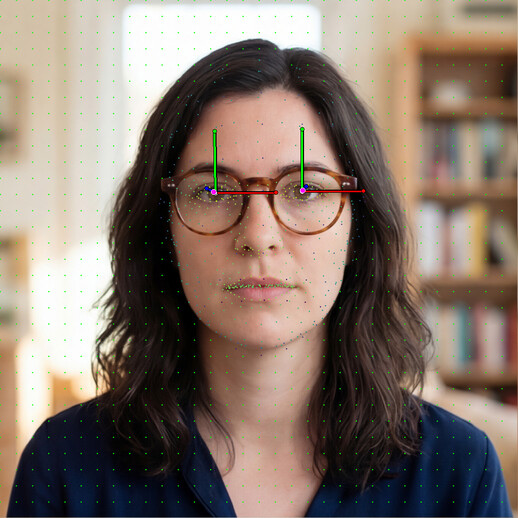} & \includegraphics[width=\linewidth]{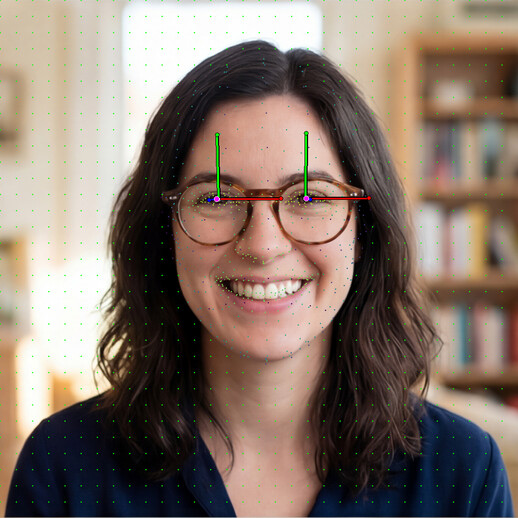} & \includegraphics[width=\linewidth]{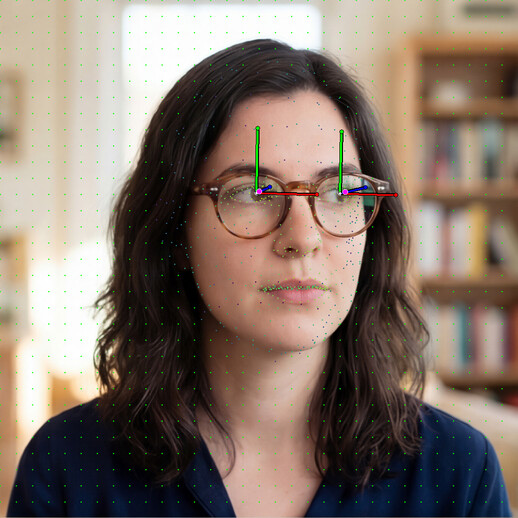} & \includegraphics[width=\linewidth]{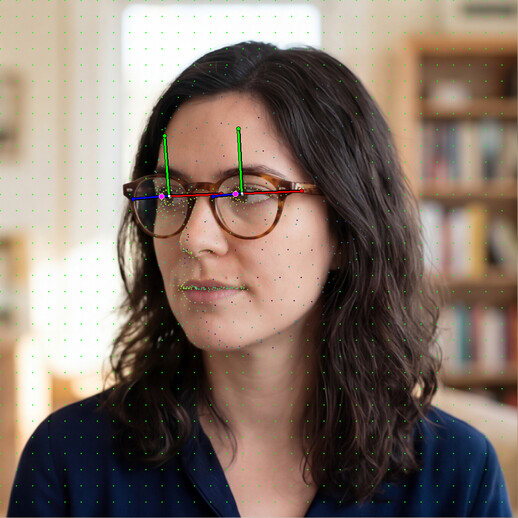} & \includegraphics[width=\linewidth]{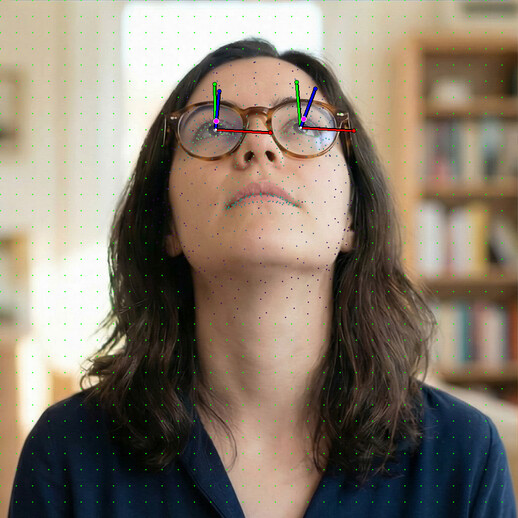} \\%
    \includegraphics[width=\linewidth]{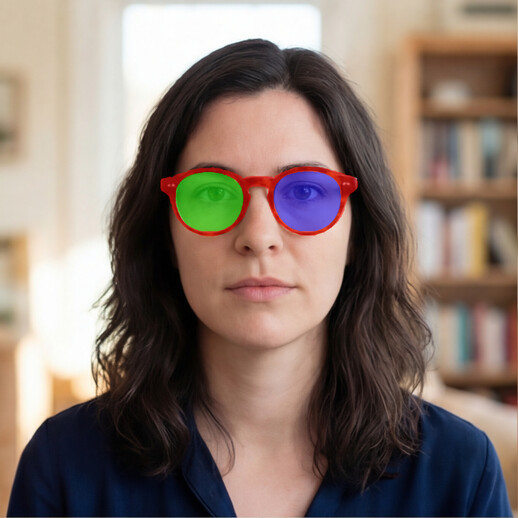} & \includegraphics[width=\linewidth]{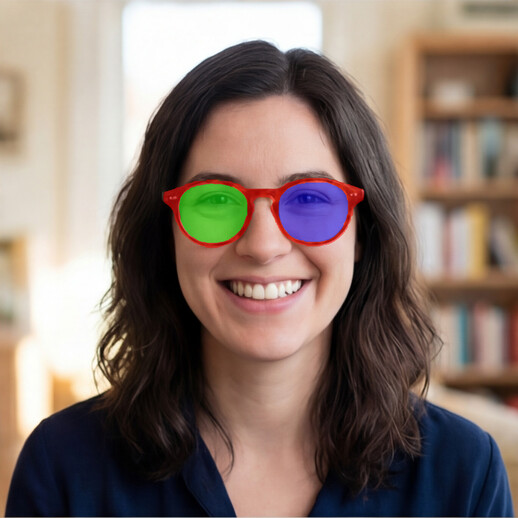} & \includegraphics[width=\linewidth]{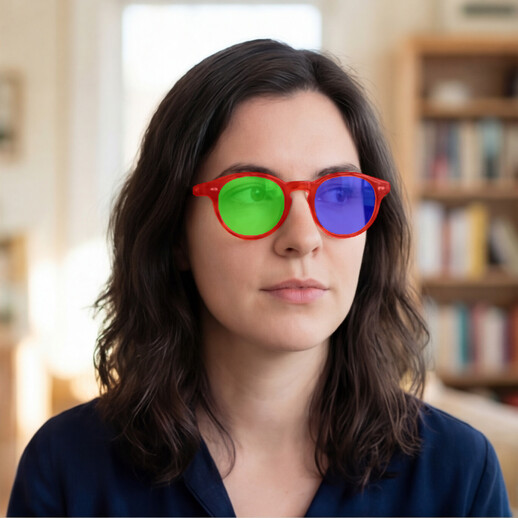} & \includegraphics[width=\linewidth]{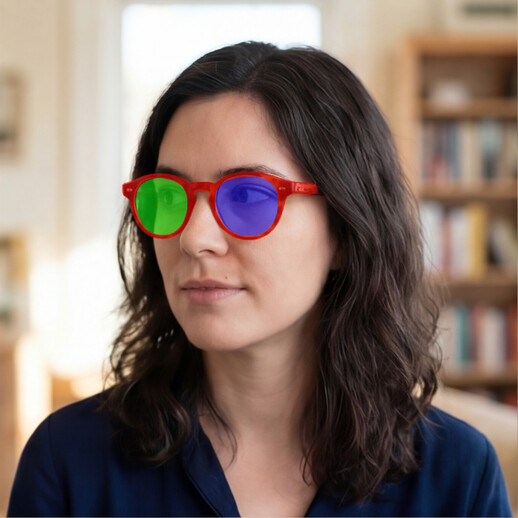} & \includegraphics[width=\linewidth]{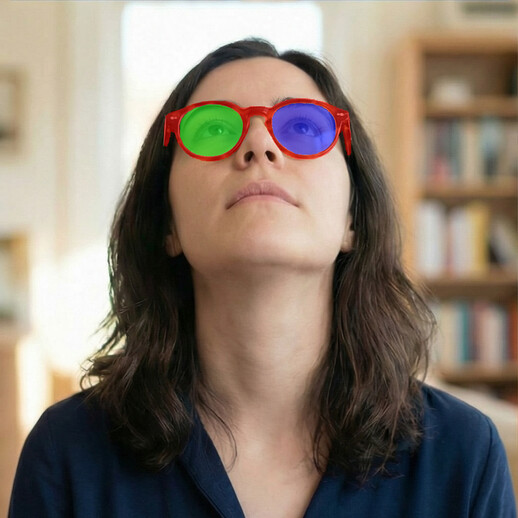} \\%
    \includegraphics[width=\linewidth]{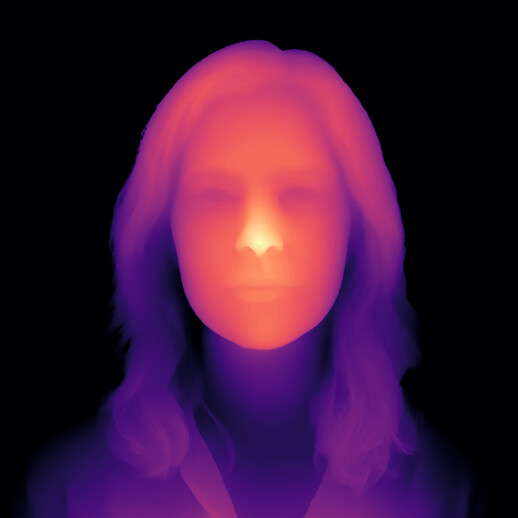} & \includegraphics[width=\linewidth]{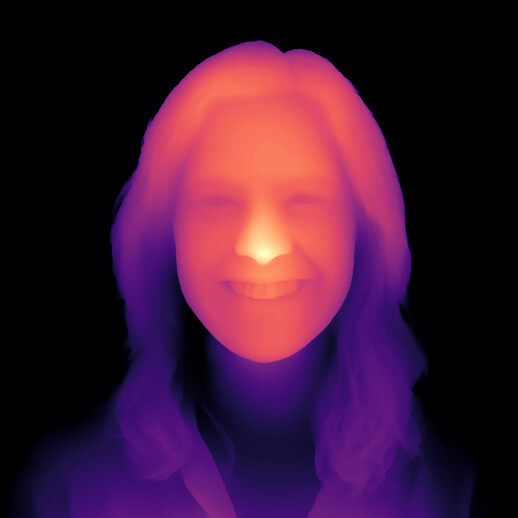} & \includegraphics[width=\linewidth]{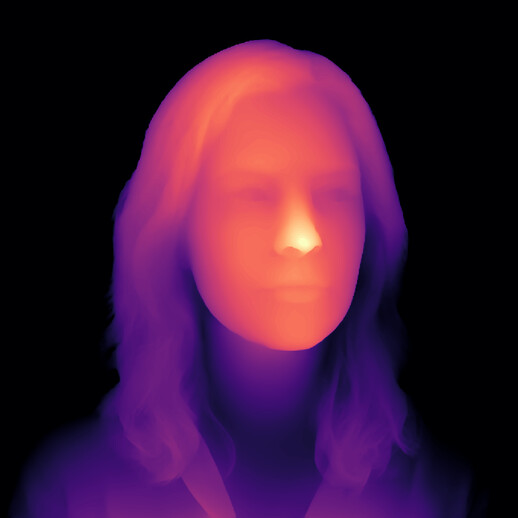} & \includegraphics[width=\linewidth]{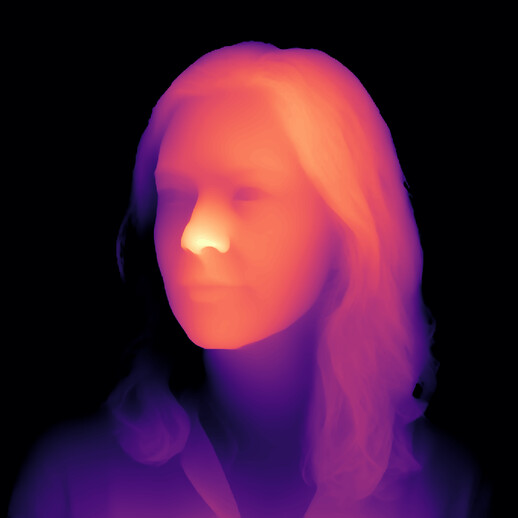} & \includegraphics[width=\linewidth]{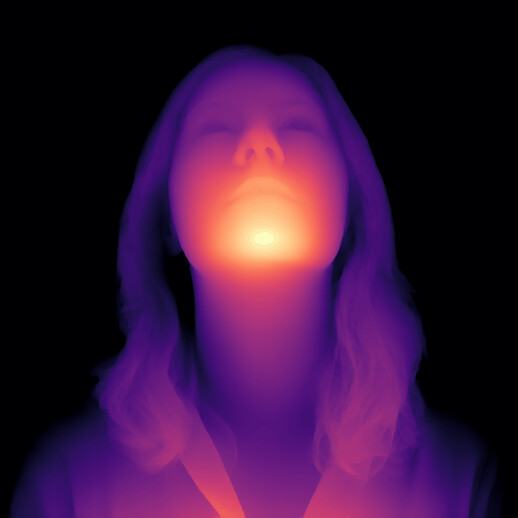} \\%
    \multicolumn{5}{c}{\includegraphics[width=0.5275\linewidth]{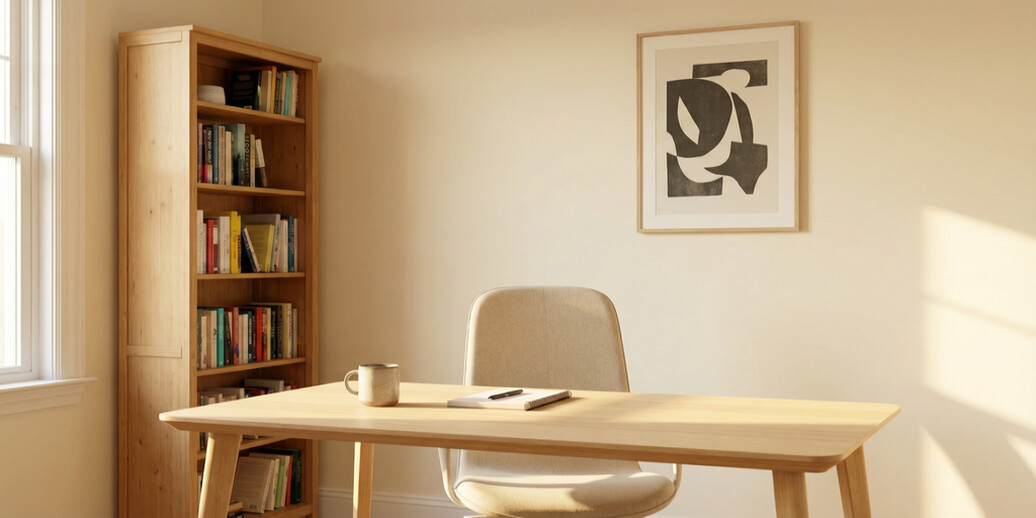}\includegraphics[width=0.4725\linewidth]{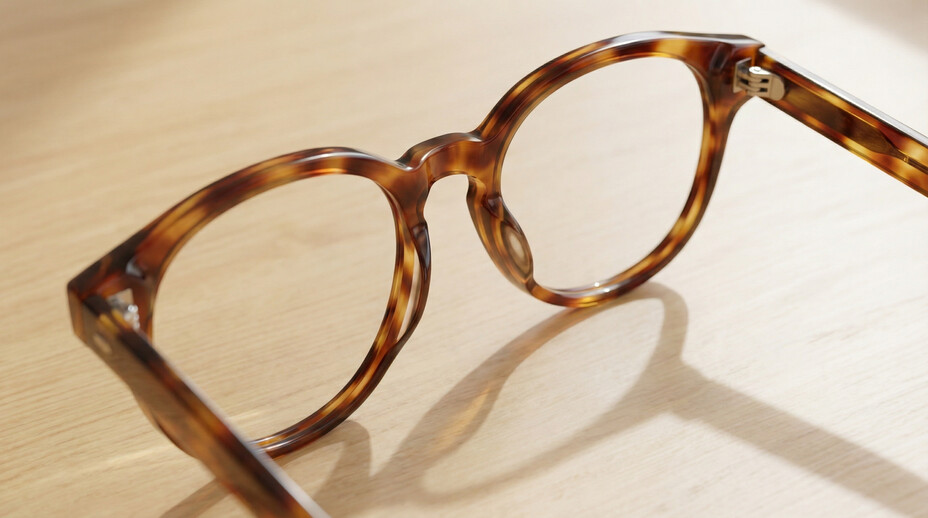}} \\%
  \end{tabularx}
  \caption{\textbf{Visualization of our generated dataset.} Row 1: Clean images with warp grids (green points), dense facial landmarks (green: certain, purple: uncertain), and gaze vectors derived from rotation matrices. Row 2: Glasses compositions. Row 3: Clean images with precise lens and frame segmentations. Row 4: Estimated depth maps. Row 5: environment map (left) and the glasses asset (right) used for composition.}
  \label{fig:dataset_sample}
\end{figure*}

In Fig.~\ref{fig:dataset_sample}, we illustrate the multi-modal nature of our generated dataset. Each sample identity in our collection is represented by 13 clean and glasses-wearing pairs across a diverse range of head poses and expressions.

\paragraph{Geometric and Semantic Ground-Truth.} Our dataset goes beyond simple image pairs by providing dense geometric and semantic annotations. As shown in the first row of Fig.~\ref{fig:dataset_sample}, each clean face is associated with precise warp grids, dense facial landmarks, and gaze vectors derived from the rotation matrices of a fitted 3D face parametric model \cite{egger_3d_2020}. The third row highlights the high-fidelity segmentation masks for both the lenses and the frames, which are generated directly by the Nano Banana (NB) prior. These masks provide the pixel-level guidance necessary for both our physics-based simulation and our restoration network.

\paragraph{Depth and Environmental Context.} To facilitate realistic optical simulation, we incorporate depth information and environmental context for every identity. The fourth row of Fig.~\ref{fig:dataset_sample} displays estimated depth maps used to compute refractive warps. Furthermore, the final row shows the generative environment maps and standalone eyewear assets. By conditioning reflections on these inverse-view scenes, our simulation captures the lighting conditions of the original portrait, bridging the gap between purely synthetic rendering and generative composition. This comprehensive set of modalities ensures that our restoration model is trained on data that is both photorealistic and structurally grounded.

\subsection{Data Generation Prompts}\label{sec:data_generation_prompts}

The generation of our multi-pose synthetic dataset follows a structured prompting workflow. These instructions are designed to enforce geometric consistency while maintaining demographic diversity.

\paragraph{Identity and Pose Generation (3$\times$3 Grid).}
\begin{quote}\itshape\small
    ``CRITICAL INSTRUCTION: Create 9 poses of [PERSON\_DESCRIPTION]. The person making the poses MUST NOT wear glasses. The camera view is the same. Only the head poses and expressions change. It is CRITICAL that you succeed. If not, you will be TERMINATED! Random background that is the same in all 9 poses, talking head framing. Poses/expressions: frontal/neutral, looking halfway towards left shoulder/neutral, looking halfway towards right shoulder/neutral, head tilted back/neutral, frontal/smile without teeth, frontal/smile showing teeth, frontal/raised eyebrows, frontal/frowning, frontal/neutral. If you will not succeed, we will be fired. You MUST succeed! IMPORTANT: No gutter, seamless, images touching, no borders, no frames, zero padding, edge-to-edge layout! CRITICAL: Lighting conditions MUST be realistic, not photo studio-like, amateur photography-like, selfie-like!!!''
\end{quote}

\paragraph{Extreme Poses (2$\times$2 Grid).}
\begin{quote}\itshape\small
    ``Make a grid of 2x2 images showing the same person in the following poses: extreme left head turn, extreme right turn, extreme chin up (look up), extreme chin down (look down). It is CRITICAL that you succeed.''
\end{quote}

\paragraph{Eyewear Reference Generation.}
\begin{quote}\itshape\small
    ``Generate a single top-right view of common dioptric eyeglasses.'' \textnormal{or} ``Generate a single top-right view of dioptric glasses that look completely different to the ones attached.''
\end{quote}

\paragraph{Synthetic Eyewear Fitting.}
\begin{quote}\itshape\small
    ``Have the subject wear the attached glasses. CRITICAL: Do not change the skin, eyes, or anything else. It is CRITICAL that you succeed. If the image will change significantly, you will be TERMINATED!''
\end{quote}

\paragraph{Inverse-View Scene Generation.}
\begin{quote}\itshape\small
    ``Create a person-free picture of the environment on the mirror side (the reverse view) of the attached picture. Location and lighting must match the original picture.''
\end{quote}

\paragraph{Semantic Segmentation Masks.}
\begin{quote}\itshape\small
    ``Create a glasses mask - left lens in blue, right lens in green, frame in red.''
\end{quote}

\paragraph{Prompt Parameterization.}
The \textit{[PERSON\_DESCRIPTION]} placeholder is dynamically populated with demographic attributes (e.g., \textit{hispanic real-world woman in her thirties}) and appearance constraints (\textit{average-looking, no photo model}) to ensure validity and avoid bias toward professional studio photography.

\subsection{Complete Quantitative Results on FFHQ}\label{sec:ffhq_full_results}
\begin{table}[t]
  \centering
  \caption{\textbf{Quantitative evaluation on the Flickr-Faces-HQ (FFHQ) dataset~\cite{karras_style-based_2019}.} We compare our proposed approach against several state-of-the-art image-based, video editing, and video inpainting methods using the Fréchet Inception Distance (FID) $\downarrow$ and inference speed (FPS) $\uparrow$. We report results for regular \textit{glasses}, \textit{sunglasses}, and the full \textit{combined} test set. Best results are highlighted in \textbf{bold}.}
  \label{tab:ffhq_results_full}
  \setlength{\tabcolsep}{9pt}
  \begin{tabular}{l r@{\hskip 20pt} r @{} r r}
    \toprule
    & \multicolumn{3}{c}{Fréchet Inception Distance (FID) $\downarrow$} & \\
    \cmidrule(lr){2-4}
    Method & glasses & sunglasses & combined & FPS $\uparrow$ \\
    \midrule
    Inpaint Anything \cite{yu_inpaint_2023} & $0.391$ & $1.796$ & $0.337$ & $1.84$ \\
    InstructPix2Pix \cite{brooks_instructpix2pix_2023} & $1.089$ & $1.608$ & $0.927$ & $0.05$ \\
    LEDITS \cite{tsaban_ledits_2023} & $1.953$ & $\text{---}$ & $1.953$ & $0.08$ \\
    Take-Off-Eyeglasses \cite{lyu_portrait_2022} & $3.256$ & $2.741$ & $2.839$ & $13.31$ \\
    \midrule
    RAVE \cite{kara_rave_2024} & $11.559$ & $16.741$ & $12.065$ & $0.16$ \\
    \midrule
    TokenFlow \cite{geyer_tokenflow_2023} & $2.430$ & $\text{---}$ & $2.430$ & $0.04$ \\
    Flow-Guided Transformer \cite{zhang_flow-guided_2022} & $0.569$ & $\mathbf{1.432}$ & $0.453$ & $1.28$ \\
    ProPainter \cite{zhou_propainter_2023} & $0.408$ & $1.653$ & $\mathbf{0.335}$ & $8.77$ \\
    \midrule
    Ours & $\mathbf{0.379}$ & $1.979$ & $0.364$ & $\mathbf{27.68}$ \\
    \bottomrule
  \end{tabular}
\end{table}

The task of sunglass removal points toward pure inpainting as the eyes must be entirely hallucinated.
By contrast, our work emphasizes the restoration of facial features through the compensation of optical distortions where ground-truth guidance from the underlying geometry is available.

While our approach achieves the lowest FID in the corrective glasses category (0.379), we note that some methods in Table~\ref{tab:ffhq_results_full} exhibit lower FID scores on the \textit{combined} dataset than on the \textit{glasses} subset alone. This phenomenon is a direct consequence of the Fréchet Inception Distance being a biased estimator, where the bias is approximately proportional to $1/N$ for $N$ samples. Because the combined set utilizes the full evaluation pool ($\sim$14,000 images), the significant reduction in estimator bias outweighs the quality degradation introduced by the sunglasses subset ($\sim$2,000 images). This results in a ``suspicious but correct'' statistical outcome where the larger sample size dominates the metric calculation. Our method remains specifically optimized for the high-fidelity restoration of visible ocular details through physical compensation, which is most accurately captured by the glasses-only metric. 

\subsection{Temporal Consistency Evaluation}\label{sec:temporal_eval}
\begin{table}[t]
  \centering
  \caption{\textbf{Temporal consistency metrics on CelebV-Text clips.} RAFT-L1 is the flow-warping error (previous output warped into the current frame by RAFT-Sintel~\cite{teed_raft_2020} flow estimated on the inputs); temporal LPIPS (tLPIPS)~\cite{zhang_unreasonable_2018} is an adjacent-frame perceptual distance. Both measure stability, not restoration quality. Laplacian variance is a sharpness/detail proxy: lower-detail outputs are smoother and score better on stability metrics, so the comparison is conservative for our method. Best per column in \textbf{bold}.}
  \label{tab:temporal_metrics}
  \setlength{\tabcolsep}{0pt}\footnotesize
  \begin{tabular*}{\linewidth}{@{\extracolsep{\fill}}lccc@{}}
    \toprule
    Approach &
    \begin{tabular}{@{}c@{}}RAFT-L1\\$\times10^3\downarrow$\end{tabular} &
    \begin{tabular}{@{}c@{}}tLPIPS\\$\times10^3\downarrow$\end{tabular} &
    \begin{tabular}{@{}c@{}}Laplacian\\variance$\uparrow$\end{tabular} \\
    \midrule
    Nano Banana~\cite{google_gemini_2025}            & 9.763  & 21.215 & 83.9 \\
    LEDITS~\cite{tsaban_ledits_2023}                 & 14.455 & 27.904 & 73.9 \\
    FGT~\cite{zhang_flow-guided_2022}                & 6.529  & 17.722 & 37.0 \\
    TOE~\cite{lyu_portrait_2022}                     & \textbf{6.517} & \textbf{16.569} & 19.9 \\
    V-LASIK~\cite{shalev-arkushin_v-lasik_2025}      & 7.760  & 16.803 & 54.9 \\
    STTN~\cite{zeng_learning_2020}                   & 7.152  & 19.658 & 66.7 \\
    ProPainter~\cite{zhou_propainter_2023}           & 6.569  & 16.756 & 57.8 \\
    Runway Gen-4.5~\cite{runway_ai_inc_runway_2025}  & 7.970  & 16.844 & 56.8 \\
    \midrule
    Ours (JFSnet)                                    & 7.682  & 18.528 & \textbf{162.6} \\
    \bottomrule
  \end{tabular*}
\end{table}

To complement the perceptual study (Sec.~\ref{sec:perceptual_study}) with objective measurements, we evaluate temporal stability on the 60 CelebV-Text clips used in the video survey, using common, method-agnostic metrics that require no masks or learned task-specific features. For RAFT-L1, we estimate dense RAFT-Sintel~\cite{teed_raft_2020} optical flow on the (repeated) input frames, warp the previous output into the current frame, and measure the L1 residual, which compensates for input motion. Temporal LPIPS (tLPIPS) measures the perceptual distance between adjacent output frames. Both quantify stability rather than restoration quality. Laplacian variance is a sharpness/detail proxy: smoother, lower-detail outputs incur lower temporal error, which is why a strongly smoothing method such as TOE~\cite{lyu_portrait_2022} ranks well temporally while retaining the least detail (lowest sharpness, $19.9$). The comparison is therefore conservative for JFSnet, which produces by far the sharpest outputs ($162.6$).

As Table~\ref{tab:temporal_metrics} shows, JFSnet is materially more temporally stable than the framewise generative and editing baselines (Nano Banana, LEDITS, Runway Gen-4.5) while preserving the high-frequency detail that conservative inpainters discard.

\begin{table}[t]
  \centering
  \caption{\textbf{Temporal-loss ablation.} Effect of the translation-equivariance loss $\lambda_{\mathrm{temp}}$. Restoration metrics use a freshly generated, identity-disjoint paired split (300 identities, 3{,}900 paired samples); temporal metrics use the CelebV-Text video protocol of Table~\ref{tab:temporal_metrics}. Despite no paired-video supervision, the temporal loss reduces temporal errors with no spatial-restoration trade-off. Best per column in \textbf{bold}.}
  \label{tab:temporal_loss_ablation}
  \setlength{\tabcolsep}{1.8pt}\footnotesize
  \begin{tabular*}{\textwidth}{@{\extracolsep{\fill}}c ccc cc cc}
    \toprule
    & \multicolumn{3}{c}{Global} & \multicolumn{2}{c}{Lens region} & \multicolumn{2}{c}{Temporal} \\
    \cmidrule(lr){2-4}\cmidrule(lr){5-6}\cmidrule(lr){7-8}
    \begin{tabular}{@{}c@{}}JFSnet\\$\lambda_{\mathrm{temp}}$\end{tabular} &
    SSIM$\uparrow$ & LPIPS$\downarrow$ & \begin{tabular}{@{}c@{}}L1\\$\times 10^2\!\downarrow$\end{tabular} & PSNR$\uparrow$ & LPIPS$\downarrow$ &
    \begin{tabular}{@{}c@{}}RAFT-L1\\$\times 10^3\!\downarrow$\end{tabular} &
    \begin{tabular}{@{}c@{}}tLPIPS\\$\times 10^3\!\downarrow$\end{tabular} \\
    \midrule
    \ding{56} & 0.947 & 0.025 & 1.01 & 27.44 & 0.00307 & 8.023 & 19.520 \\
    \ding{52} & \textbf{0.950} & \textbf{0.024} & \textbf{1.00} & \textbf{27.65} & \textbf{0.00300} & \textbf{7.682} & \textbf{18.528} \\
    \bottomrule
  \end{tabular*}
\end{table}

Table~\ref{tab:temporal_loss_ablation} ablates the translation-equivariance loss $\lambda_{\mathrm{temp}}$ on the same protocol. Despite using no paired-video supervision, it reduces both temporal metrics (RAFT-L1 $8.023\!\to\!7.682$, tLPIPS $19.520\!\to\!18.528$) without degrading any spatial-restoration metric, confirming its role in suppressing flicker.

\subsection{Baselines Retrained on Our Data}\label{sec:retrained_baselines}
\begin{table}[t]
  \centering
  \caption{\textbf{Prior baselines retrained on our data.} We retrain a video inpainter (ProPainter~\cite{zhou_propainter_2023}, run on the input image with our lens masks) and a glasses-specific method (TOE~\cite{lyu_portrait_2022}) on \textit{JFSnet}'s training data, and evaluate on a 300-identity paired set (identity-disjoint, freshly generated, not used in any prior training). \faSnowflake{}: as originally reported; \faFire{}: retrained on our data. Global metrics cover the whole image; lens-region metrics isolate the actual restoration task. Best in \textbf{bold}.}
  \label{tab:retrained_baselines}
  \setlength{\tabcolsep}{3.5pt}\footnotesize
  \begin{tabular*}{\textwidth}{@{\extracolsep{\fill}}l ccc cc}
    \toprule
    & \multicolumn{3}{c}{Global} & \multicolumn{2}{c}{Lens region} \\
    \cmidrule(lr){2-4}\cmidrule(lr){5-6}
    Approach & SSIM$\uparrow$ & LPIPS$\downarrow$ & L1$\downarrow$ & PSNR$\uparrow$ & LPIPS$\downarrow$ \\
    \midrule
    ProPainter~\cite{zhou_propainter_2023} \faSnowflake{} & 0.907 & 0.054 & 0.021 & 15.28 & 0.037 \\
    ProPainter~\cite{zhou_propainter_2023} \faFire{}      & 0.913 & 0.044 & 0.020 & 16.71 & 0.027 \\
    TOE~\cite{lyu_portrait_2022} \faSnowflake{}           & 0.916 & 0.050 & 0.022 & 18.98 & 0.012 \\
    TOE~\cite{lyu_portrait_2022} \faFire{}                & 0.940 & 0.028 & 0.014 & 24.19 & 0.004 \\
    \midrule
    Ours (JFSnet)                                         & \textbf{0.950} & \textbf{0.024} & \textbf{0.010} & \textbf{27.65} & \textbf{0.003} \\
    \bottomrule
  \end{tabular*}
\end{table}

To separate the contribution of our data from that of our architecture, we retrain two prior baselines on \textit{JFSnet}'s training data: a video inpainter (ProPainter~\cite{zhou_propainter_2023}, applied to the input image with our lens masks) and a glasses-specific method (TOE~\cite{lyu_portrait_2022}). All methods are evaluated on a 300-identity paired set (identity-disjoint, freshly generated, not used in any prior training). ProPainter copies non-masked pixels from the input, while TOE processes the full image, so the lens-region metrics in Table~\ref{tab:retrained_baselines} isolate the actual restoration task. Even with same-data fine-tuning, both baselines remain well below JFSnet in the lens region --- by $+3.5$\,dB (TOE) and $+10.9$\,dB (ProPainter) PSNR --- although TOE is glasses-specific and ProPainter is a recent video inpainter. This supports our claim that inverting refraction exploits the distorted pixels, which mask-based pipelines and TOE's de-shadow/de-glass cascade are less able to use.

\subsection{ArcFace Identity Similarity Under Eyewear Removal}\label{sec:arcface_expected_drop}
In the main text (Sec.~\ref{sec:quantitative_evaluation}) we note that ArcFace~\cite{deng_arcface_2019} similarity is unreliable for identity preservation under eyewear removal. Here we quantify why: successful removal necessarily shifts the biometric embedding, so a lower ArcFace score reflects the task rather than a loss of identity. We compute ArcFace on MediaPipe-aligned $112\times112$ crops with the InsightFace \texttt{w600k\_r50} model and cosine similarity.

\begin{table}[t]
\centering
\caption{Establishing the ``Expected Biometric Drop'' for glass removal. To model the biometric penalty of removing glasses, we treat an image with synthetically heavy sunglasses (70\% dark, 15\% reflection) as the baseline identity. We then compute the identity shift when analytically ``removing'' these artifacts: restoring the true clear eyes (``Un-Lensing'') and inpainting the frames (``Un-Framing'') with OpenCV's implementation of \cite{telea_image_2004}. Combining both establishes a theoretical ceiling for identity preservation during complete glass removal.}
\label{tab:expected_drop}
\begin{tabular}{l @{\hspace{6em}} r@{\,$\pm$\,}l}
\toprule
Modification & \multicolumn{2}{c}{ArcFace Sim.~$\uparrow$} \\
\midrule
Analytic Un-Lensing (Darkening \& Reflection Removal) & 0.968 & 0.016 \\
Analytic Un-Framing (Frame Removal) & 0.982 & 0.032 \\
Combined Analytic Removal & 0.946 & 0.037 \\
\bottomrule
\end{tabular}
\end{table}
\paragraph{Analytic expected drop.} Treating an image with synthetically heavy sunglasses as the baseline identity, we measure the embedding shift induced by analytically removing the eyewear: restoring the clear eyes (``Un-Lensing'') and inpainting the frames (``Un-Framing''~\cite{telea_image_2004}). As shown in Table~\ref{tab:expected_drop}, each operation lowers similarity, and their combination establishes a ceiling of $0.946\pm0.037$ for non-generative removal --- the frame geometry itself acts as an identity anchor.

\paragraph{Real-world reference.} As an empirical floor, we captured 72 real before/after pairs (\faCamera{}~the same subject photographed wearing eyeglasses and again after they were physically removed). ArcFace similarity between these genuine same-identity pairs is only $0.87\pm0.02$. On FFHQ, JFSnet scores $0.89\pm0.04$ --- at or above this real-world reference --- whereas a conservative inpainter (Inpaint Anything~\cite{yu_inpaint_2023}) scores $0.95\pm0.03$ by leaving ocular content largely unmodified. The reduced similarity is therefore intrinsic to removing the eyewear, not a failure to preserve identity.

\begin{table*}[t]
\centering
\caption{Component Analysis of Generative Glass Removal across State-of-the-Art Methods. By selectively substituting generated content into specific geometric regions of the original image, we identify the primary sources of identity shift. The results demonstrate that models with high overall ArcFace scores frequently leave structural frame artifacts (as seen in higher Gen. Frame Only scores) and fail to deeply reconstruct the ocular region (as seen in near-perfect Gen. Lenses Only scores). Our pipeline provides the most structurally complete removal, incurring the expected biometric penalty.}
\label{tab:multimethod_gen_components}
  \setlength{\tabcolsep}{6pt}
\begin{tabular}{l r@{\,$\pm$\,}l r@{\,$\pm$\,}l r@{\,$\pm$\,}l r@{\,$\pm$\,}l}
\toprule
 & \multicolumn{2}{c}{Frame Only} & \multicolumn{2}{c}{Lenses Only} & \multicolumn{2}{c}{\hspace{-0.3em}Glasses Area} & \multicolumn{2}{c}{\hspace{-0.4em}Full Generated} \\
Method & \multicolumn{2}{c}{\tiny (Orig. Lenses)} & \multicolumn{2}{c}{\tiny (Orig. Frame)} & \multicolumn{2}{c}{\hspace{-0.3em}\tiny (Frames + Lenses)} & \multicolumn{2}{c}{\hspace{-0.4em}Output} \\
\midrule
Nano Banana & 0.954 & 0.023 & 0.962 & 0.022 & 0.919 & 0.031 & 0.883 & 0.036 \\
Inpaint Anything & 0.961 & 0.020 & 0.985 & 0.024 & 0.950 & 0.029 & 0.948 & 0.030 \\
InstructPix2Pix & 0.961 & 0.026 & 0.929 & 0.055 & 0.893 & 0.083 & 0.625 & 0.304 \\
LEDITS & 0.965 & 0.028 & 0.938 & 0.045 & 0.900 & 0.068 & 0.755 & 0.126 \\
TOE & 0.954 & 0.026 & 0.975 & 0.014 & 0.934 & 0.029 & 0.908 & 0.030 \\
\midrule
IP-FaceDiff & 0.979 & 0.012 & 0.923 & 0.039 & 0.907 & 0.041 & 0.266 & 0.126 \\
RAVE & 0.934 & 0.045 & 0.861 & 0.060 & 0.786 & 0.081 & 0.316 & 0.109 \\
\midrule
TokenFlow & 0.974 & 0.019 & 0.947 & 0.034 & 0.923 & 0.047 & 0.705 & 0.096 \\
FGT & 0.944 & 0.030 & 0.975 & 0.027 & 0.920 & 0.037 & 0.897 & 0.037 \\
ProPainter & 0.940 & 0.034 & 0.975 & 0.027 & 0.916 & 0.040 & 0.890 & 0.046 \\
\midrule
Ours & 0.951 & 0.027 & 0.970 & 0.023 & 0.922 & 0.031 & 0.894 & 0.037 \\
\bottomrule
\end{tabular}
\end{table*}
\paragraph{Component analysis across methods.} Using our precise masks, we selectively substitute only the generated frames, only the generated lenses, or the entire glasses area back into the original clean images (Table~\ref{tab:multimethod_gen_components}). Across all methods, replacing only the frame region lowers similarity even when the ocular features are untouched, confirming that frames act as biometric anchors; methods that retain residual frame silhouettes (e.g., IP-FaceDiff, TokenFlow) artificially stabilize the score rather than improving restoration. JFSnet's full-output similarity ($0.894$) is consistent with the analytic ceiling once frame-cast shadow removal is accounted for, while diffusion-based resampling (RAVE, IP-FaceDiff) collapses identity ($<0.32$).

\subsection{Architectural Ablation}\label{sec:architectural_ablation_supp}
\begin{table}[t]
  \centering
  \caption{\textbf{Architectural Ablation.} We evaluate the impact of different encoder-decoder backbones, global skip connections, and encoder fine-tuning. Our hybrid JFSnet (ViT encoder, CNN decoder, and global skip) achieves the best performance, effectively balancing high-level semantic consistency with low-level spatial detail. Beyond FID, we report paired restoration metrics (SSIM/LPIPS/L1) for each configuration on the 300-identity paired set.} 
  \label{tab:arch_ablation}
  \setlength{\tabcolsep}{6pt}
  \begin{tabular}{cccc ccc r}
    \toprule
    & & Global & & & & & \\
    Encoder & Decoder & Skip & Encoder \faUnlock & SSIM~$\uparrow$ & LPIPS~$\downarrow$ & L1~$\downarrow$ & FID~$\downarrow$ \\
\midrule
    ViT & ViT  & --- & --- & 0.941 & \textbf{0.070} & 0.024 & 1.131 \\
    ViT & ViT  & --- & \checkmark & 0.940 & 0.170 & 0.025 & 0.572 \\
    CNN & CNN  & --- & --- & 0.895 & 0.110 & 0.020 & 0.493 \\
    CNN & CNN  & \checkmark & --- & 0.911 & 0.113 & 0.017 & 0.439 \\
    ViT & ViT  & \checkmark & \checkmark & \textbf{0.943} & 0.171 & 0.026 & 0.392 \\
    ViT & CNN  & --- & \checkmark & 0.917 & 0.170 & 0.025 & 0.387 \\
    ViT & ViT  & \checkmark & --- & 0.928 & 0.172 & 0.026 & 0.384 \\
    ViT & CNN  & \checkmark & \checkmark & 0.932 & \textbf{0.070} & \textbf{0.013} & \textbf{0.379} \\
    \bottomrule
  \end{tabular}
\end{table}
In Table~\ref{tab:arch_ablation}, we explore various architectural configurations to justify our choice of JFSnet. Purely convolutional architectures (CNN-only) maintain spatial structure but struggle with the global identity consistency required for high-fidelity restoration, resulting in higher FID scores (0.439 vs.\ 0.379). Conversely, all-Transformer models (ViT-ViT) exhibit significant artifacts without extensive fine-tuning and global skip connections, often failing to recover high-frequency facial textures previously occluded by eyewear.

Our hybrid approach (JFSnet) leverages the powerful semantic representation of DINOv2 while utilizing a CNN decoder to recover high-frequency spatial details. The addition of a global skip connection is critical for preserving pixel-level fidelity, particularly for facial features not occluded by the eyeglasses, yielding a significant reduction in FID (0.493 vs.\ 0.439 for CNN and 0.387 vs.\ 0.379 for JFSnet). Finally, fine-tuning the last four layers of the ViT backbone (unlocked) provides the necessary adaptation to the restoration task, enabling the model to better handle the complex optical distortions of reflections and refractions.

\subsection{JFSnet Architecture}\label{sec:jfsnet_architecture}

The \textit{JFSnet} restoration network ($f_{\theta}$) is designed to map the input image to the clean RGB face by combining the semantic robustness of Vision Transformers with the spatial precision of convolutional networks. It utilizes a \textit{DINOv2} (ViT-L/14) encoder to capture high-level facial structure and identity features. The output tokens from the ViT backbone are reshaped into a $37 \times 37$ feature map and processed via a $1 \times 1$ convolution to match the channel dimension of the decoder. To adapt the pre-trained encoder to the task while maintaining its representational power, we fine-tune only the last four layers of the ViT backbone. The decoder is a custom U-Net-style CNN that aggregates these multi-level features, fusing them with a high-resolution skip connection from the input image to maintain pixel-level fidelity.

\subsection{As-Rigid-As-Possible (ARAP) Warping}\label{sec:arap}
To ensure precise spatial alignment between the Nano Banana clean and glasses-wearing pairs, we apply an As-Rigid-As-Possible (ARAP) deformation mesh. This step accounts for the subtle pose and expression drifts that may persist even in high-fidelity generative sequences.

We construct a 2D mesh $M = (V, E, T)$ using facial landmarks and a coarse grid of background points. We then optimize the deformed vertex positions $V'$ by minimizing an energy function:
\begin{equation}
    E(V') = \lambda_{rigid} E_{rigid}(V') + \lambda_{data} E_{data}(V') + \lambda_{photo} E_{photo}(V')
\end{equation}
where $E_{rigid}$ preserves local facial geometry by finding optimal per-triangle rotations $R_T$. $E_{data}$ pulls vertices toward detected landmarks in the glasses image, and $E_{photo}$ is a multi-scale photometric loss that aligns the appearance of the clean image with the glasses image, particularly in the regions surrounding the frames. For the lens regions, we utilize a robust Geman-McClure-like loss to handle the deliberate occlusions and reflections introduced by the eyewear. This optimization is implemented in JAX and performed for 300 iterations per pair, ensuring that the final synthetic training data is strictly aligned with the underlying facial ground truth.

\subsection{Full Ablation Study}\label{sec:ablation_study_supp}

In Table~\ref{tab:ablation_study}, we provide the full results of our ablation study, examining the impact of physics-based data augmentation (reflections and refractions), our dataset filtering process, as-rigid-as-possible (ARAP) warping, specific loss components, and the fine-tuning strategy for the Vision Transformer encoder. These results validate that each component of our pipeline contributes to the final restoration quality, with the combination of all elements achieving the best performance.

\begin{table}[t]
  \centering
  \setlength{\tabcolsep}{4pt}
  \caption{\textbf{Ablation Study.} Impact of different augmentations, loss components, and encoder freezing strategies on FID scores. Best results are highlighted in \textbf{bold}.}\vspace{-0.5em}
  \label{tab:ablation_study}
  \begin{tabular}{c c c c c c c c c c c r }
    \toprule
    Refr. & Refl. & \scriptsize $\lambda_{\text{perc}}$ & \scriptsize $\lambda_{\text{eye-pix}}$ & \scriptsize $\lambda_{\text{eye-perc}}$ & \scriptsize $\lambda_{\text{adv}}$ & \scriptsize $\lambda_{\text{temp}}$ & \scriptsize Enc. \faUnlock & \scriptsize \parbox{3.6em}{\centering dataset\\filtering} & \scriptsize \parbox{3.6em}{\centering ARAP\\warping} & FID $\downarrow$ \\
\midrule
--- & \checkmark & \checkmark & \checkmark & \checkmark & \checkmark & \checkmark & \checkmark & \checkmark & \checkmark & 0.403 \\
\checkmark & \checkmark & --- & --- & --- & --- & --- & \checkmark & \checkmark & \checkmark & 0.402 \\    
\checkmark & --- & \checkmark & \checkmark & \checkmark & \checkmark & \checkmark & \checkmark & \checkmark & \checkmark & 0.398 \\
\checkmark & \checkmark & --- & \checkmark & \checkmark & \checkmark & \checkmark & \checkmark & \checkmark & \checkmark & 0.398 \\    
\checkmark & \checkmark & \checkmark & --- & \checkmark & \checkmark & \checkmark & \checkmark & \checkmark & \checkmark & 0.394 \\    
\checkmark & \checkmark & \checkmark & \checkmark & --- & \checkmark & \checkmark & \checkmark & \checkmark & \checkmark & 0.394 \\  
\checkmark & \checkmark & \checkmark & \checkmark & \checkmark & --- & \checkmark & \checkmark & \checkmark & \checkmark & 0.391 \\
\checkmark & \checkmark & \checkmark & \checkmark & \checkmark & \checkmark & \checkmark & \checkmark & ---  & --- & 0.383 \\
\checkmark & \checkmark & \checkmark & \checkmark & \checkmark & \checkmark & \checkmark & \checkmark & --- & \checkmark & 0.382 \\    
\checkmark & \checkmark & \checkmark & \checkmark & \checkmark & \checkmark & \checkmark & --- & \checkmark & \checkmark & 0.380 \\
\checkmark & \checkmark & \checkmark & \checkmark & \checkmark & \checkmark & --- & \checkmark & \checkmark & \checkmark & \textbf{0.379} \\
\checkmark & \checkmark & \checkmark & \checkmark & \checkmark & \checkmark & \checkmark & \checkmark & \checkmark & \checkmark & \textbf{0.379} \\
    \bottomrule
  \end{tabular}\vspace{-2em}
\end{table}

\subsection{Qualitative Results for Additional Methods on FFHQ}\label{sec:ffhq_results_extended}
\begin{figure*}[t]
  \centering
  \renewcommand{\arraystretch}{0}
  \setlength{\tabcolsep}{0pt}
  \begin{tabularx}{\textwidth}{@{} X @{\hskip 5pt} @{} X @{} X @{} X @{} X @{{}}}
    \parbox[b]{\linewidth}{\centering \scriptsize Input Frame} & \parbox[b]{\linewidth}{\centering \scriptsize RAVE~\cite{kara_rave_2024}} & \parbox[b]{\linewidth}{\centering \scriptsize IP-FaceDiff~\cite{anand_ip-facediff_2025}} & \parbox[b]{\linewidth}{\centering \scriptsize TokenFlow~\cite{geyer_tokenflow_2023}} & \parbox[b]{\linewidth}{\centering \scriptsize InstructPix2Pix~\cite{brooks_instructpix2pix_2023}} \\%
    \includegraphics[width=\linewidth]{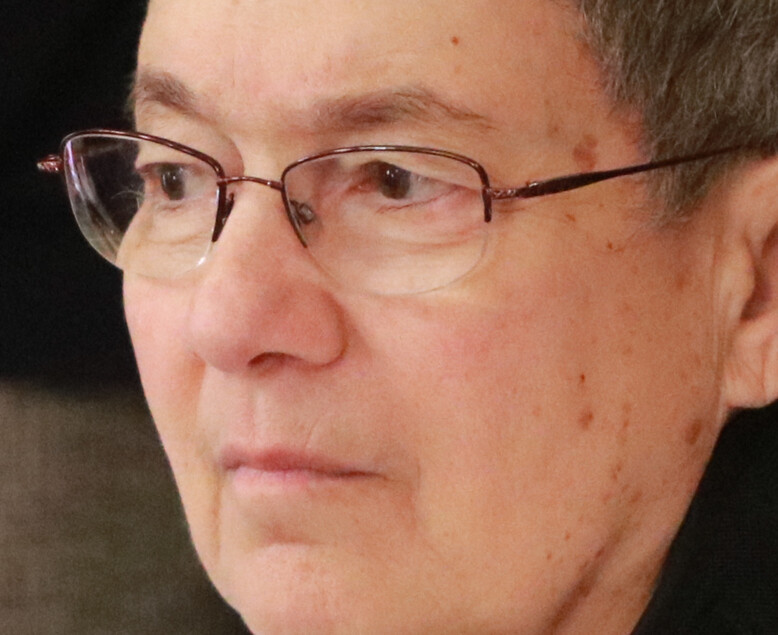} & \includegraphics[width=\linewidth]{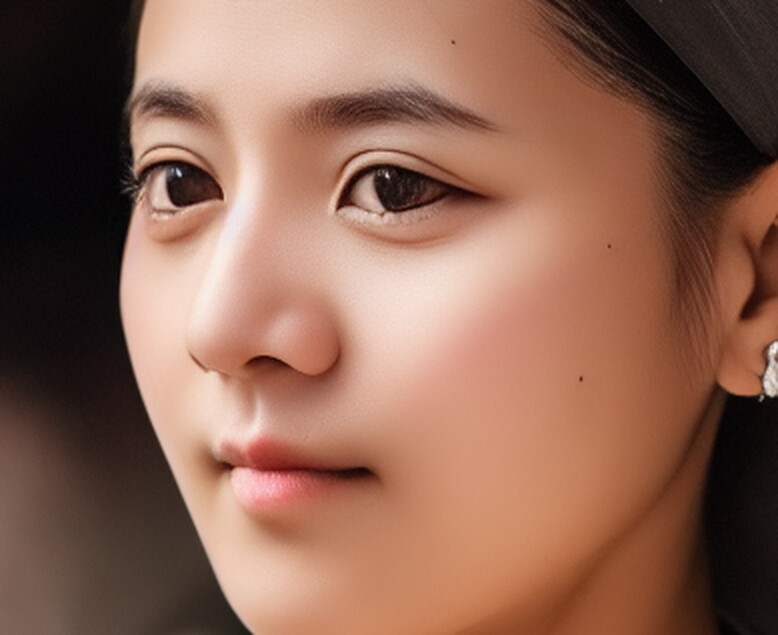} & \includegraphics[width=\linewidth]{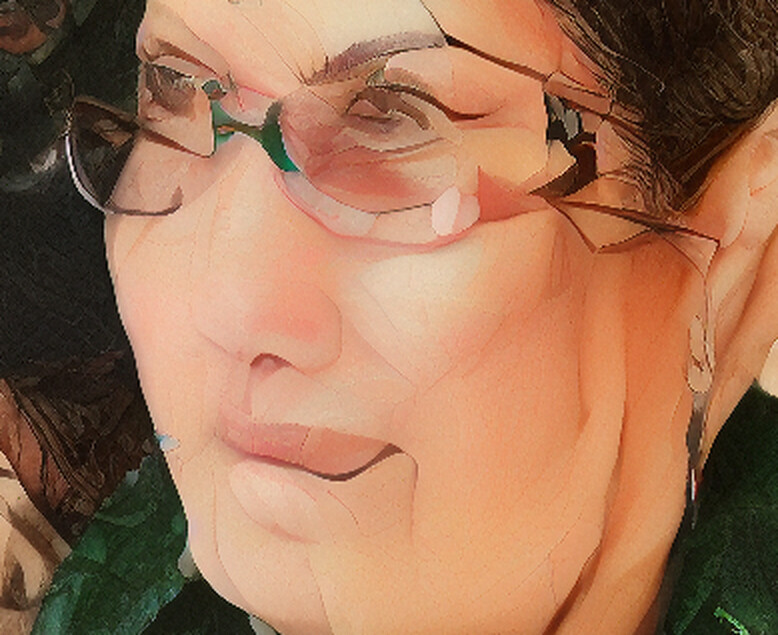} & \includegraphics[width=\linewidth]{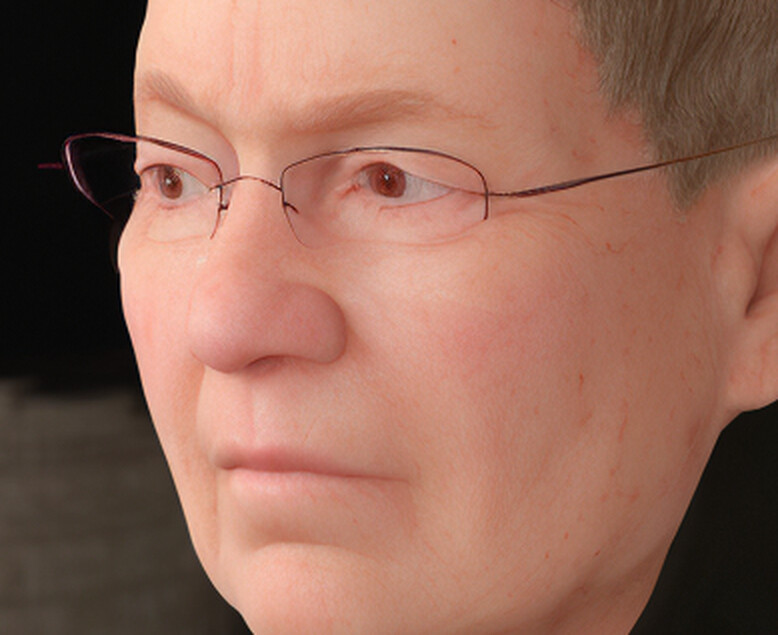} & \includegraphics[width=\linewidth]{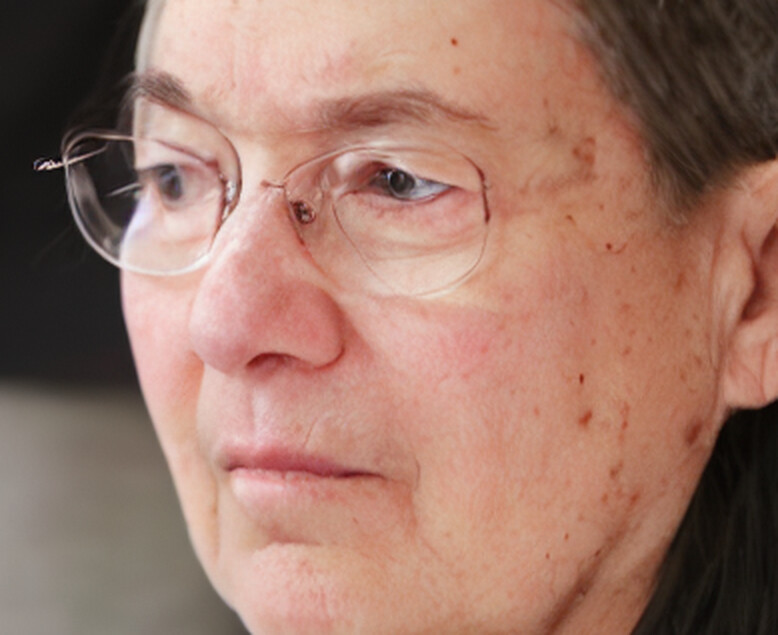} \\
    \includegraphics[width=\linewidth]{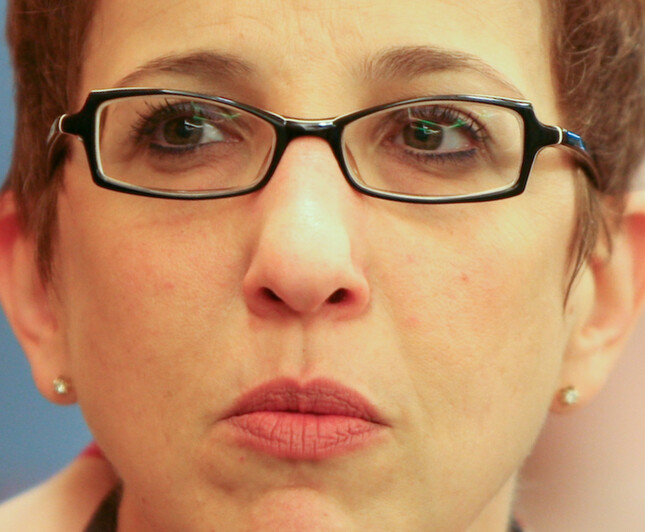} & \includegraphics[width=\linewidth]{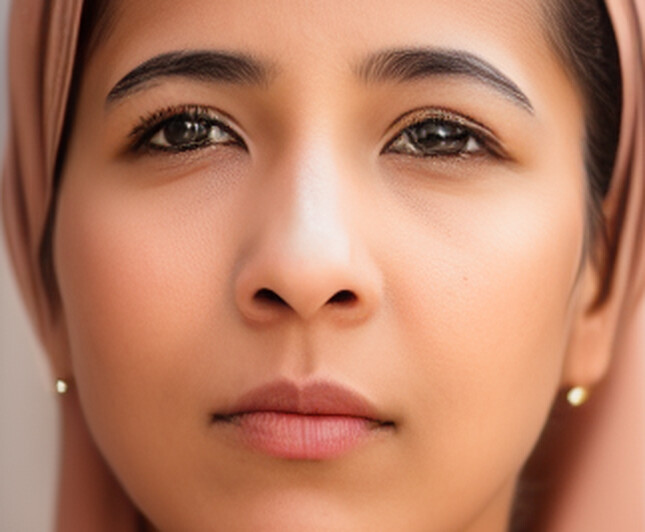} & \includegraphics[width=\linewidth]{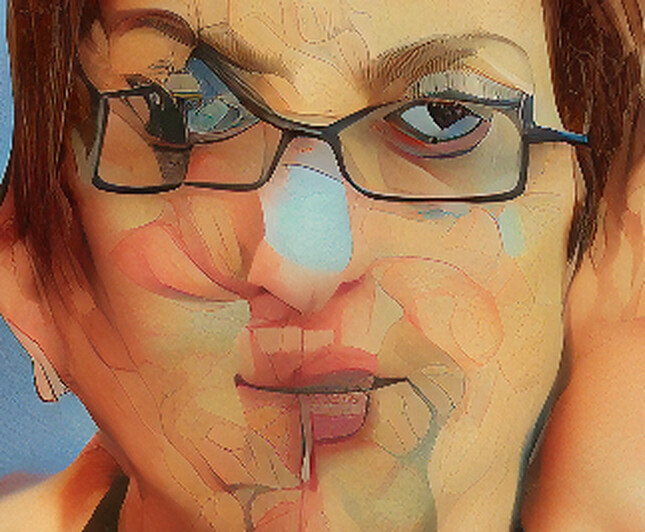} & \includegraphics[width=\linewidth]{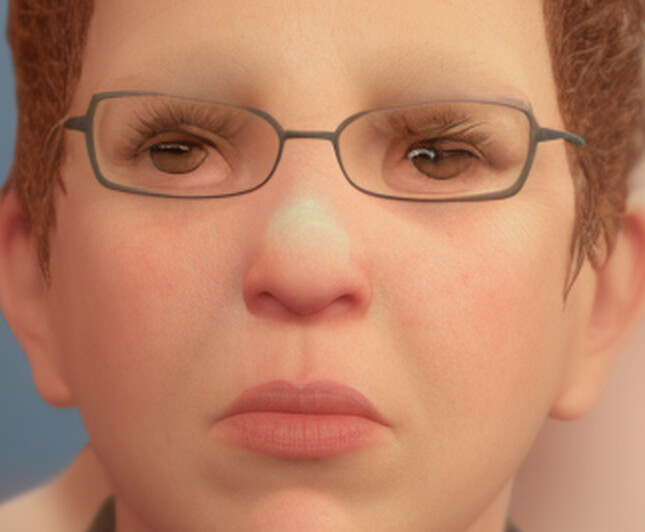} & \includegraphics[width=\linewidth]{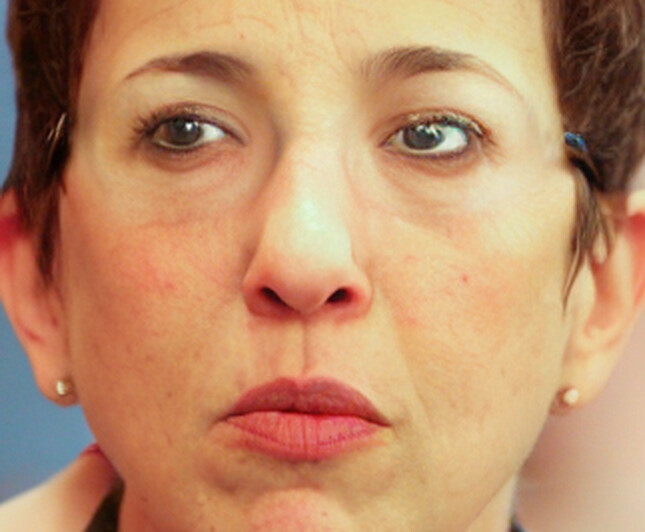} \\
    \includegraphics[width=\linewidth]{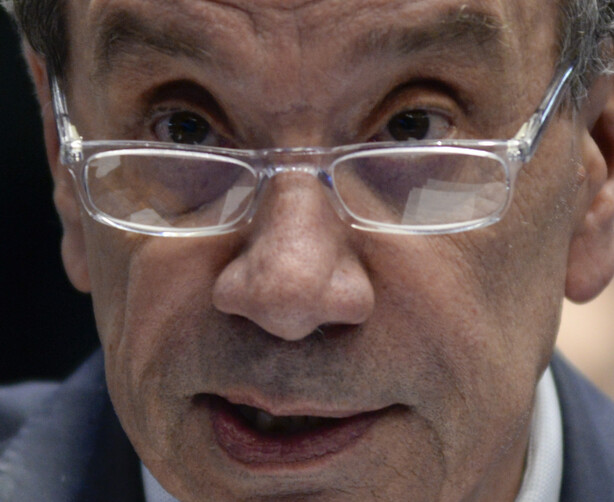} & \includegraphics[width=\linewidth]{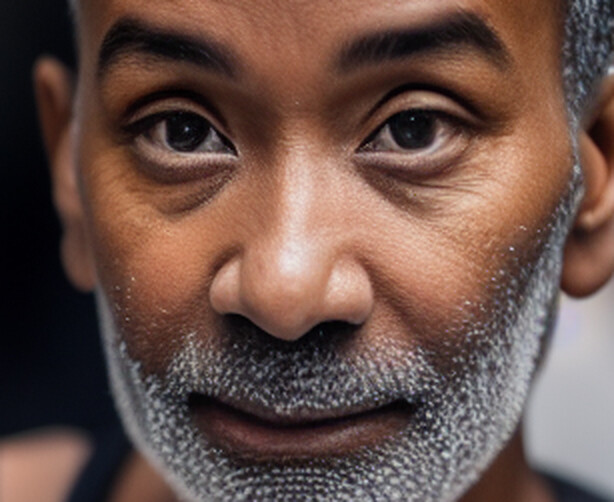} & \includegraphics[width=\linewidth]{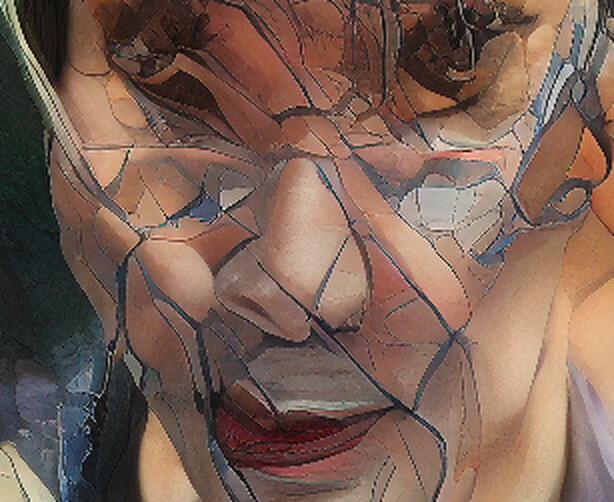} & \includegraphics[width=\linewidth]{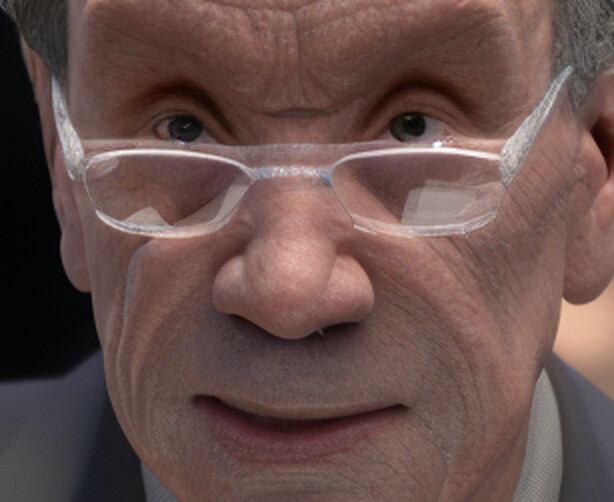} & \includegraphics[width=\linewidth]{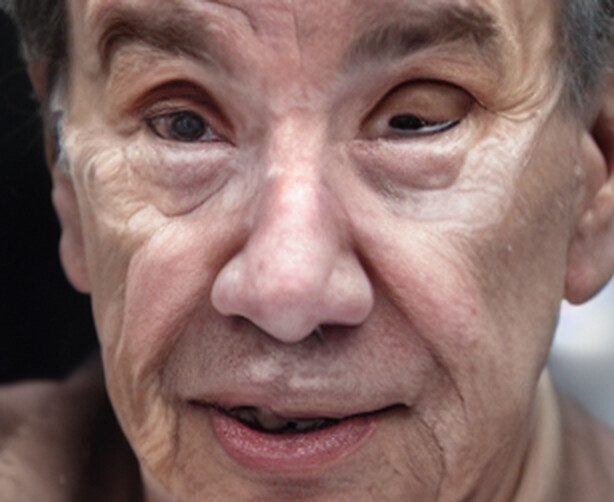} \\
    \includegraphics[width=\linewidth]{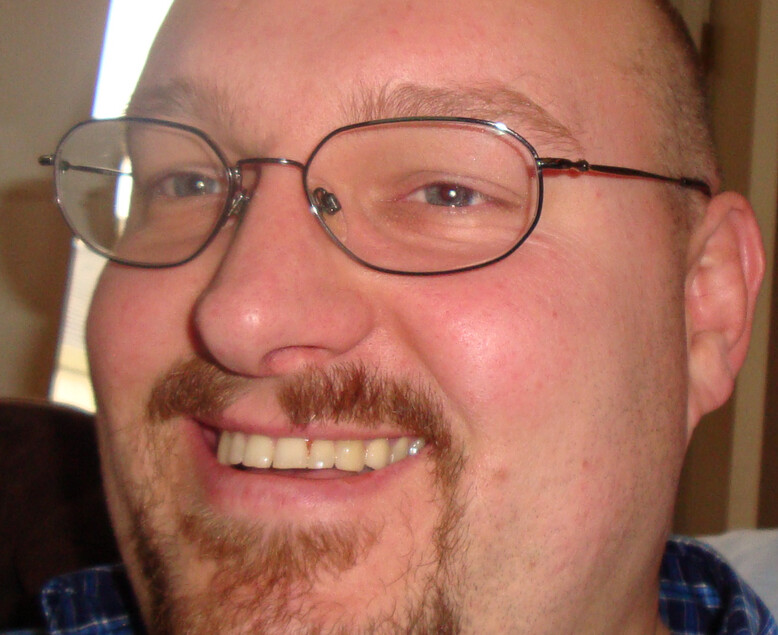} & \includegraphics[width=\linewidth]{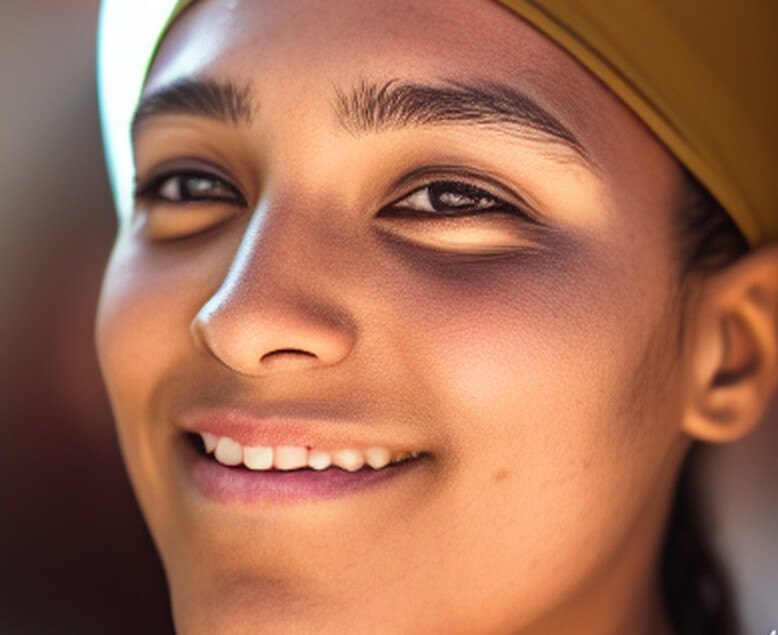} & \includegraphics[width=\linewidth]{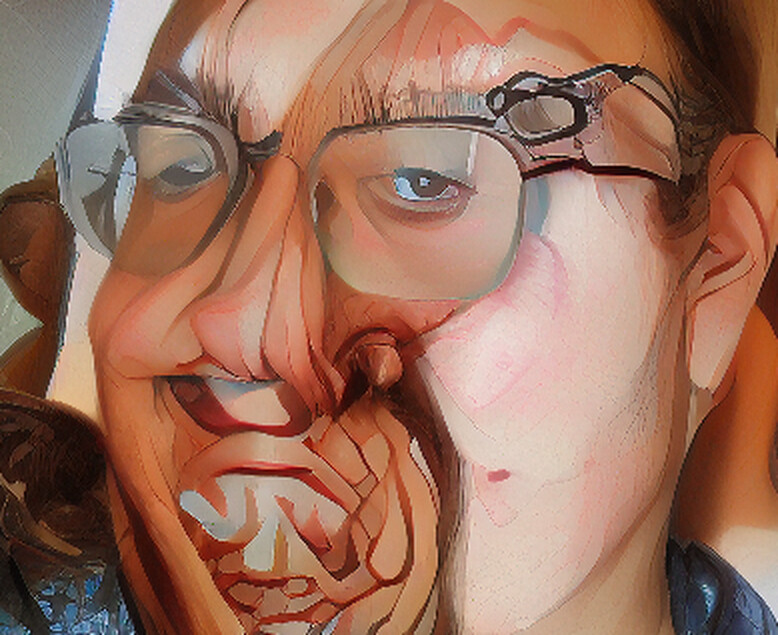} & \includegraphics[width=\linewidth]{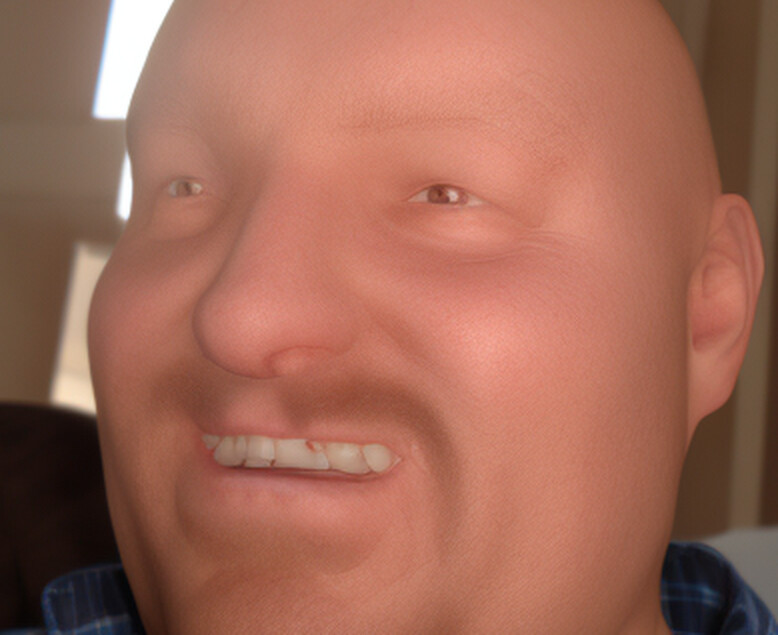} & \includegraphics[width=\linewidth]{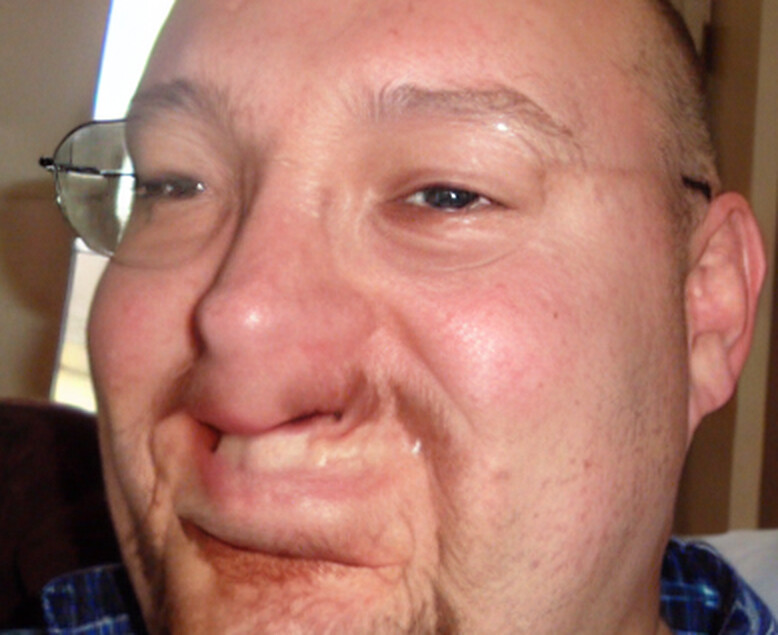} \\%
  \end{tabularx}
  \caption{\textbf{Additional qualitative comparison on FFHQ.} We show results from image-to-image and video-to-video translation methods that showed lower quantitative performance. While these methods can remove glasses, they often fail to preserve the original facial features or introduce artifacts.}
  \label{fig:ffhq_qualitative_supplementary}
\end{figure*}

\begin{figure*}[t]
  \centering
  \renewcommand{\arraystretch}{0}
  \setlength{\tabcolsep}{0pt}
  \begin{tabularx}{\textwidth}{@{} X @{\hskip 5pt} @{} X @{} X @{} X @{} X @{} X @{} X @{} X @{{}}}
    \parbox[b]{\linewidth}{\centering \scriptsize Input Frame} & \parbox[b]{\linewidth}{\centering \scriptsize Our Approach} & \parbox[b]{\linewidth}{\centering \scriptsize Nano Banana~\cite{google_gemini_2025}} & \parbox[b]{\linewidth}{\centering \scriptsize TOE\\ \cite{lyu_portrait_2022}} & \parbox[b]{\linewidth}{\centering \scriptsize ProPainter\\ \cite{zhou_propainter_2023}} & \parbox[b]{\linewidth}{\centering \scriptsize FGT\\ \cite{zhang_flow-guided_2022}} & \parbox[b]{\linewidth}{\centering \scriptsize STTN\\ \cite{zeng_learning_2020}} & \parbox[b]{\linewidth}{\centering \scriptsize LEDITS\\ \cite{tsaban_ledits_2023}} \\%
    \includegraphics[width=\linewidth]{00000_00111_png_ref} & \includegraphics[width=\linewidth]{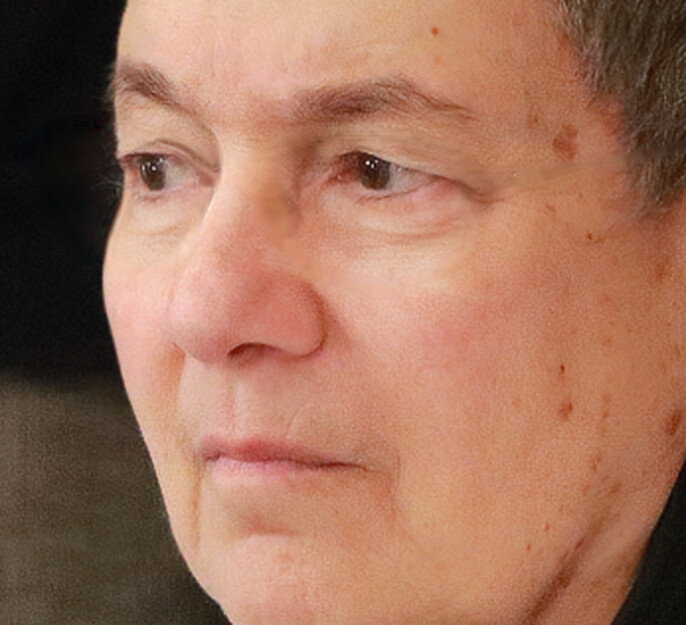} & \includegraphics[width=\linewidth]{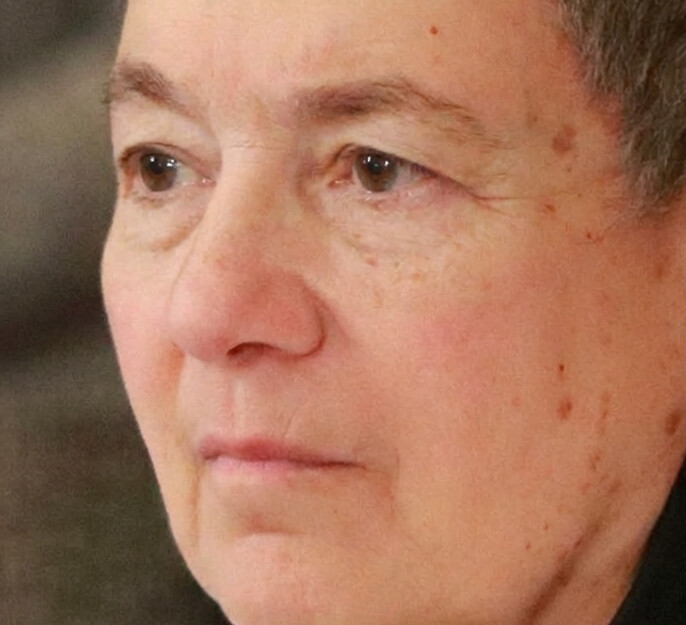} & \includegraphics[width=\linewidth]{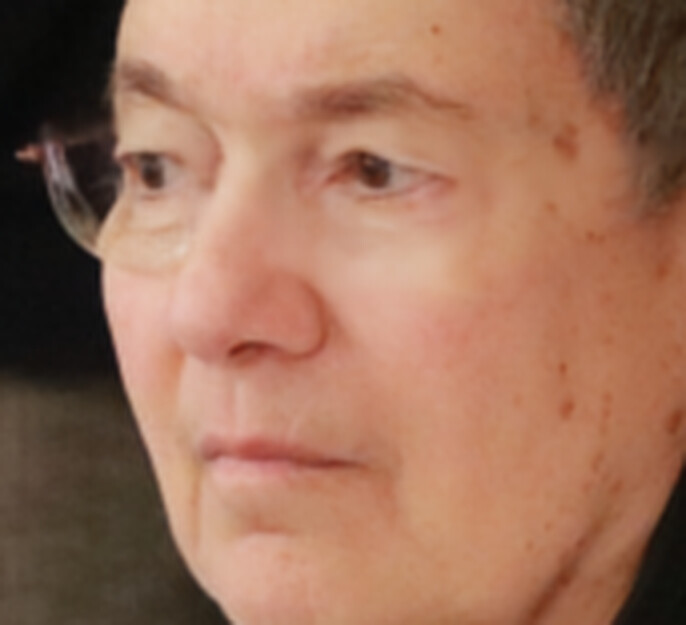} & \includegraphics[width=\linewidth]{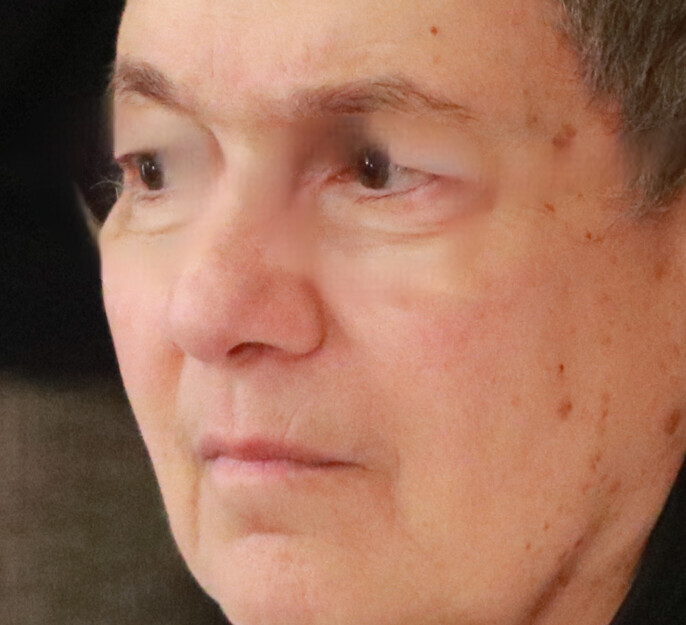} & \includegraphics[width=\linewidth]{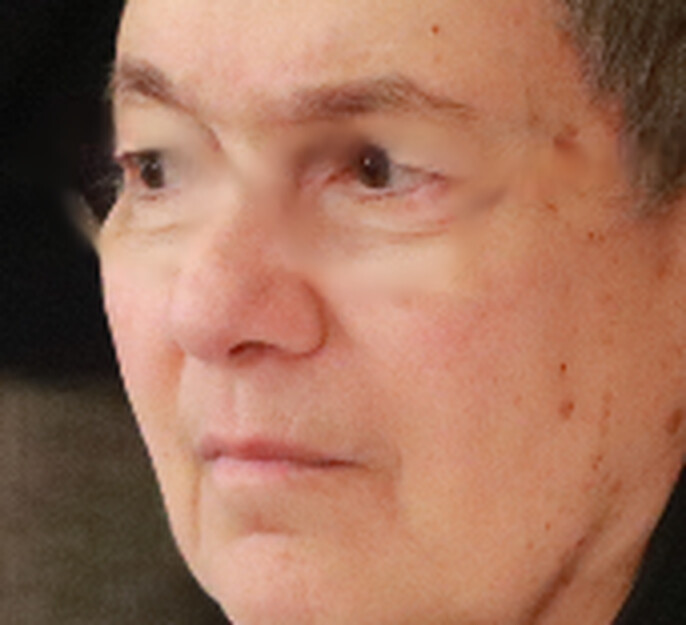} & \includegraphics[width=\linewidth]{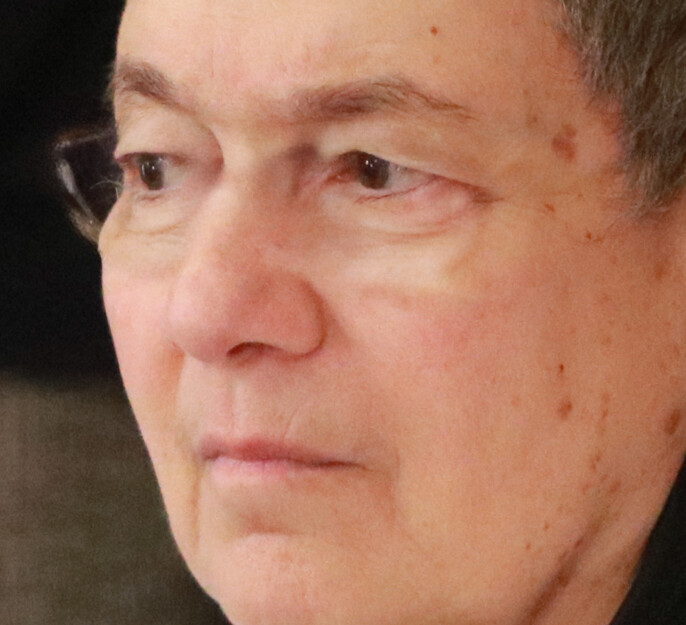} & \includegraphics[width=\linewidth]{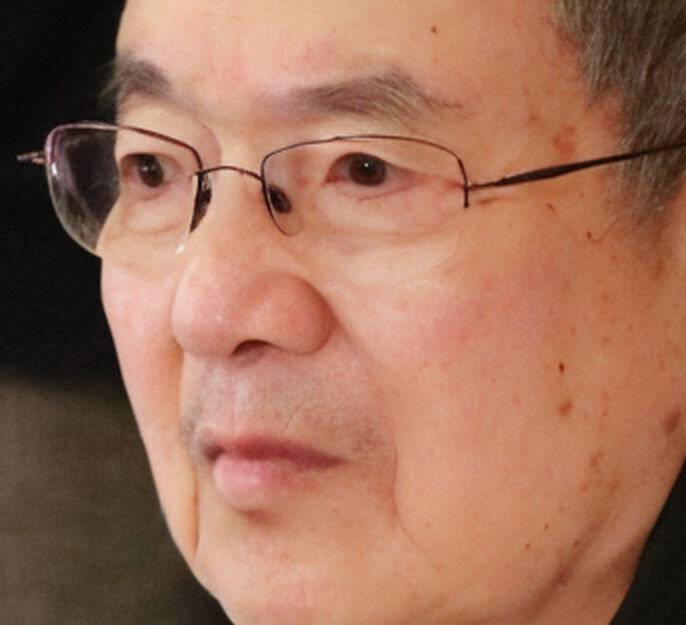} \\
    \includegraphics[width=\linewidth]{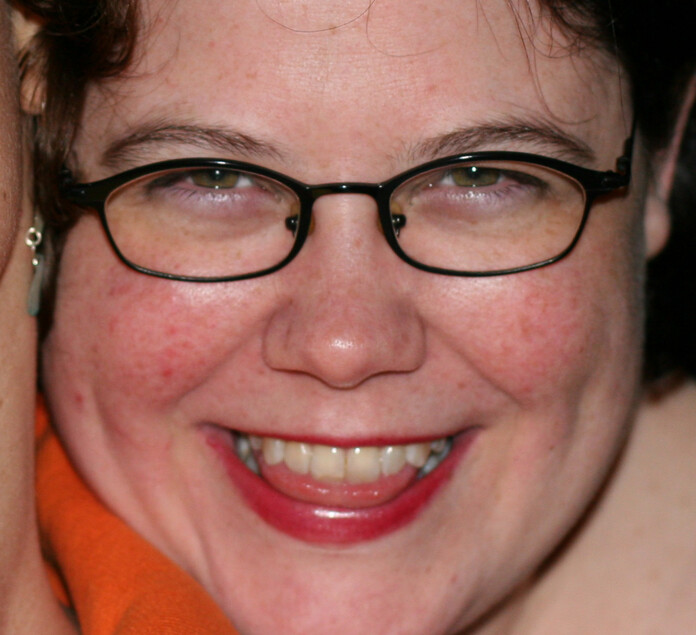} & \includegraphics[width=\linewidth]{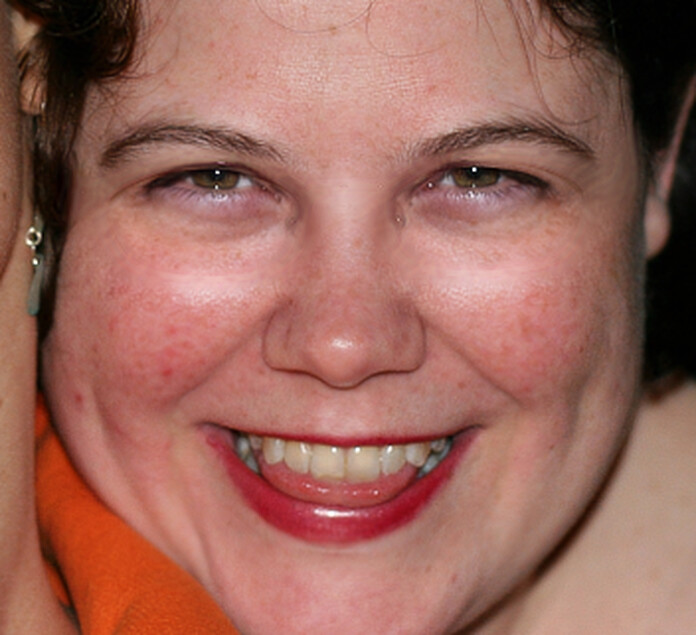} & \includegraphics[width=\linewidth]{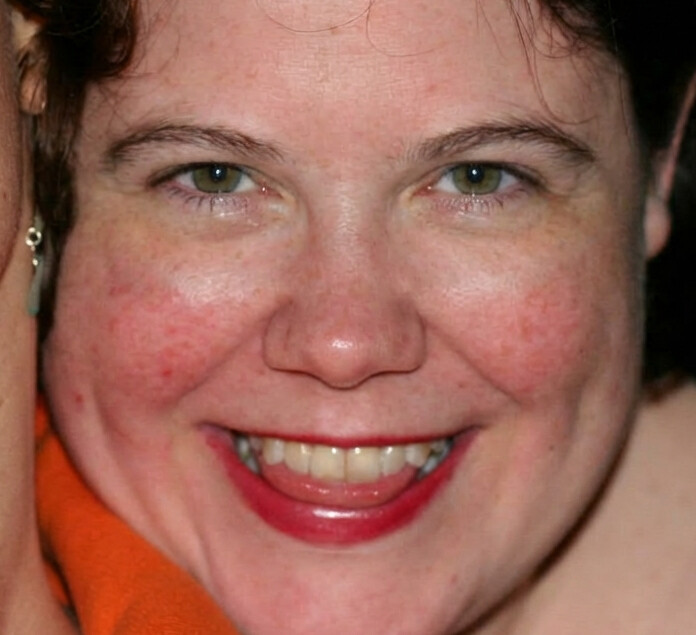} & \includegraphics[width=\linewidth]{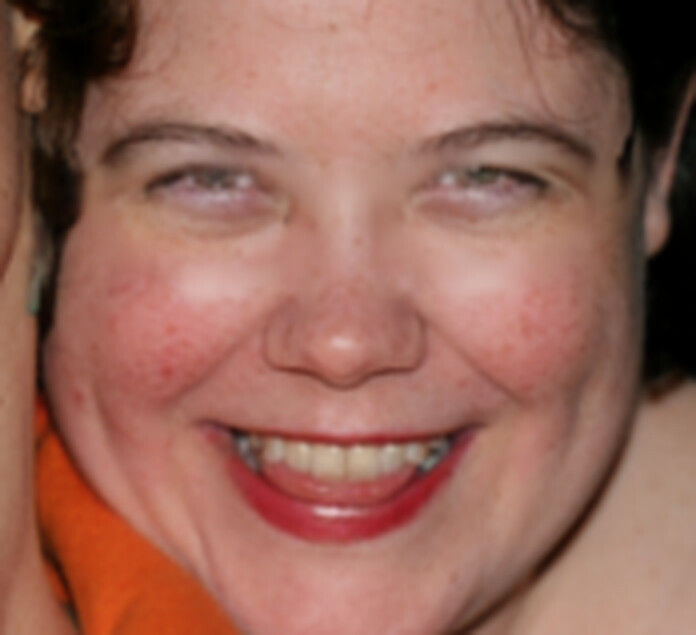} & \includegraphics[width=\linewidth]{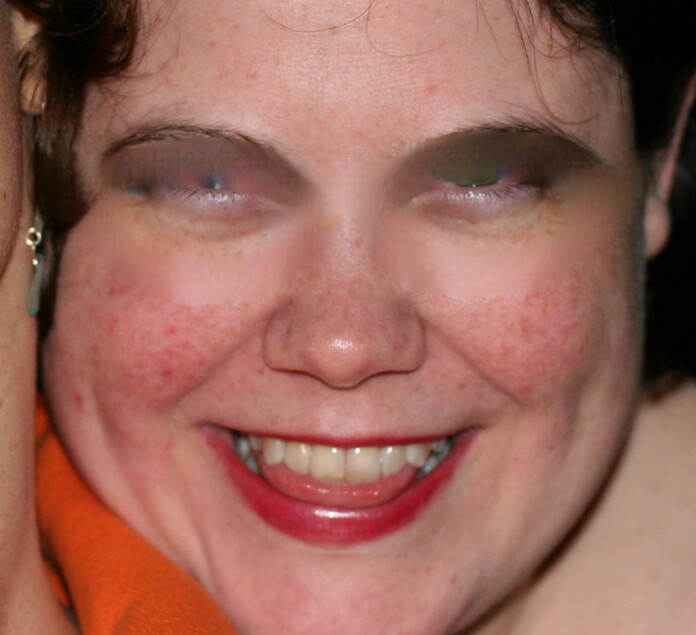} & \includegraphics[width=\linewidth]{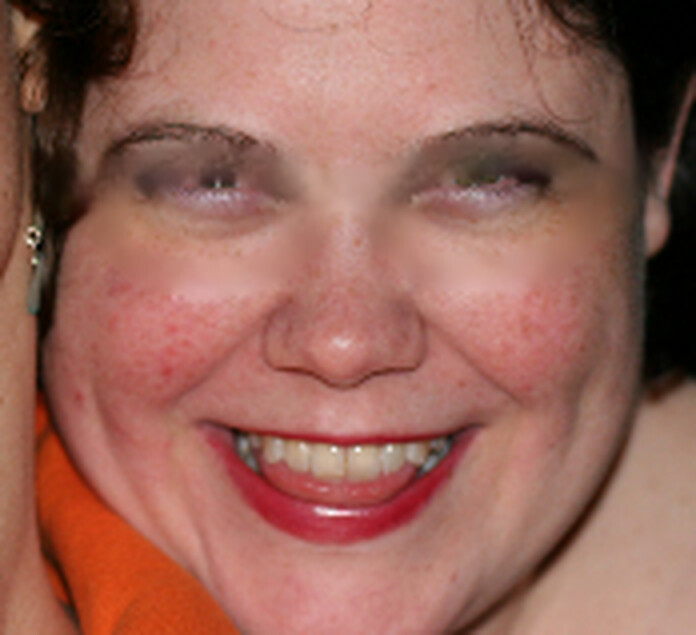} & \includegraphics[width=\linewidth]{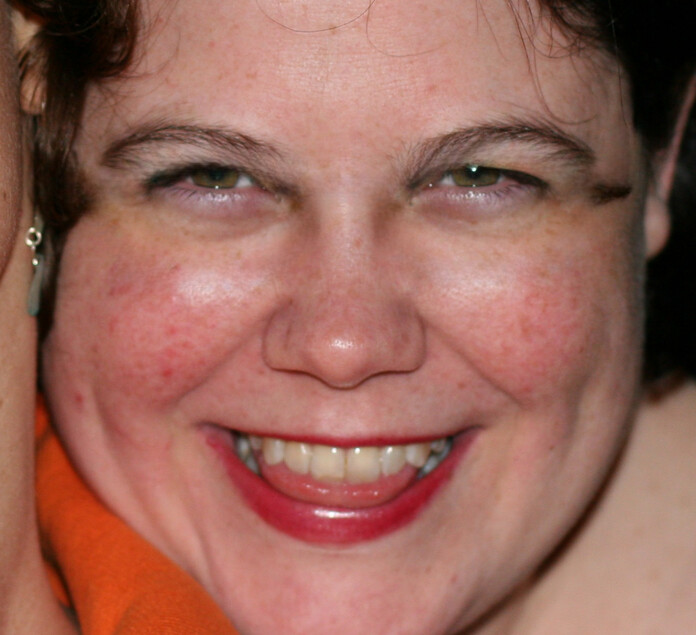} & \includegraphics[width=\linewidth]{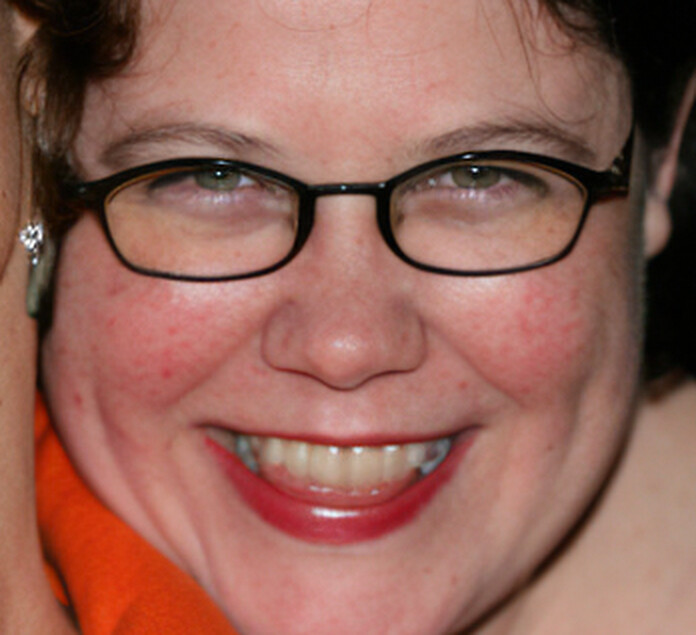} \\
    \includegraphics[width=\linewidth]{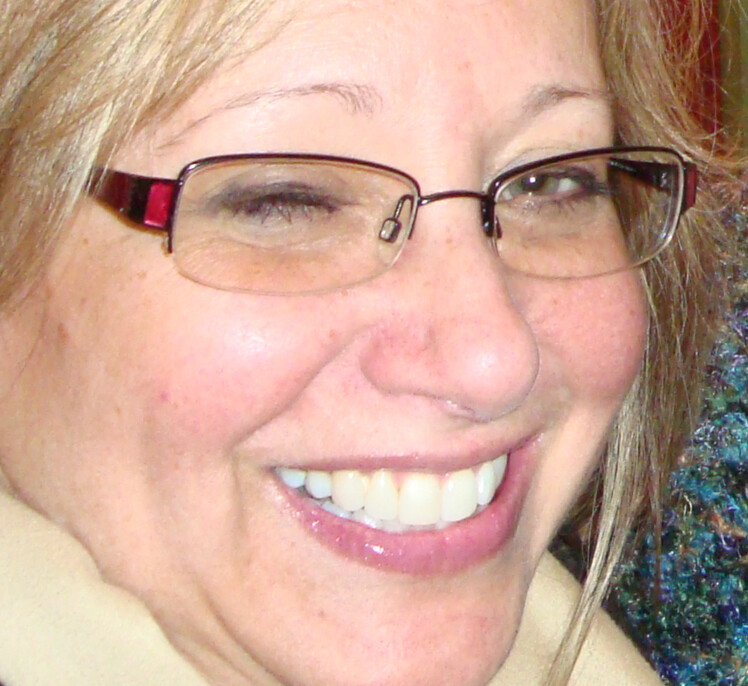} & \includegraphics[width=\linewidth]{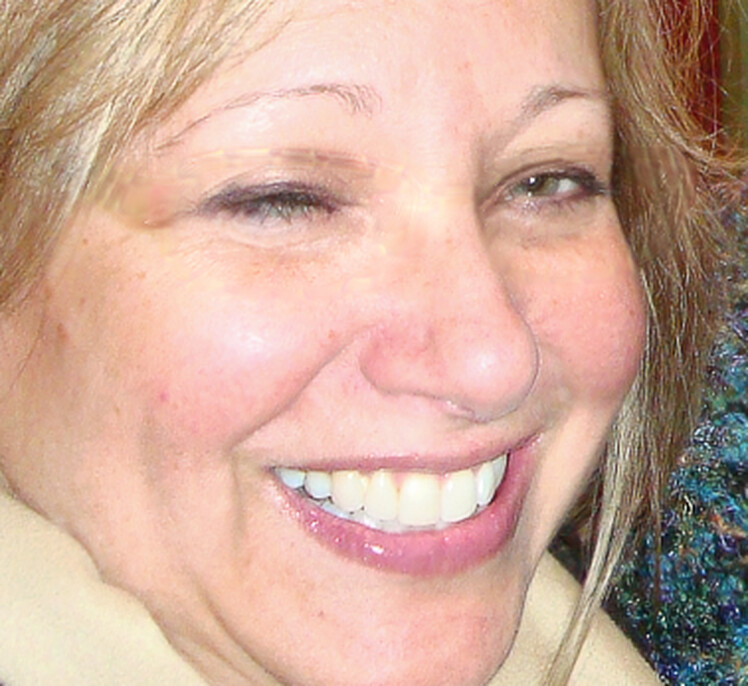} & \includegraphics[width=\linewidth]{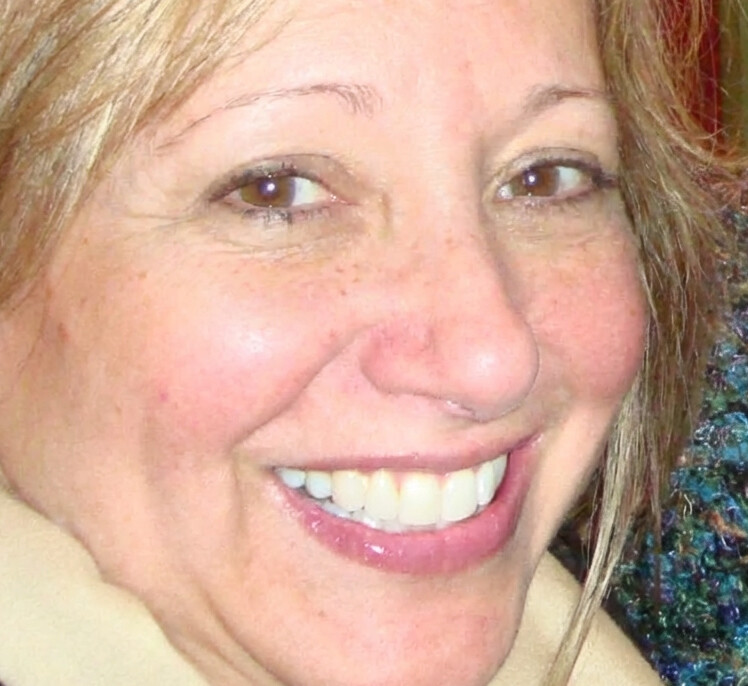} & \includegraphics[width=\linewidth]{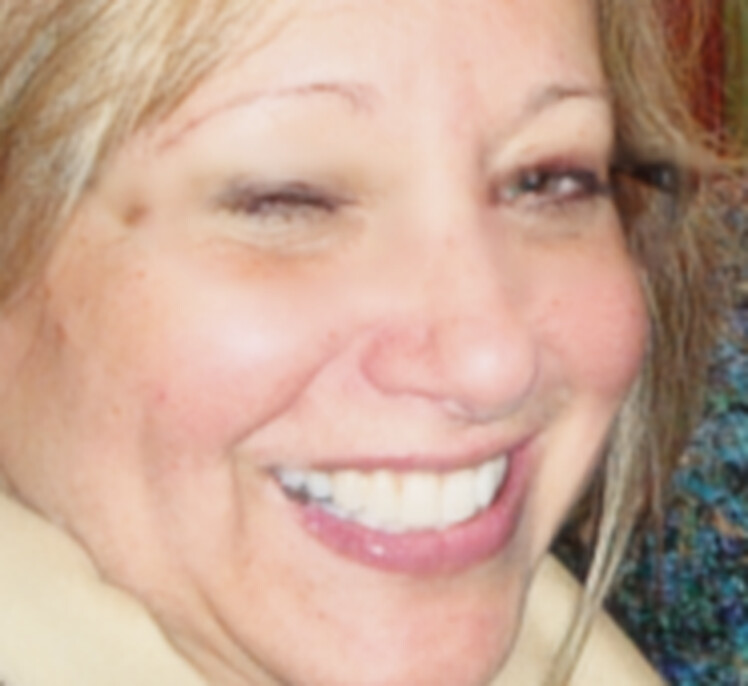} & \includegraphics[width=\linewidth]{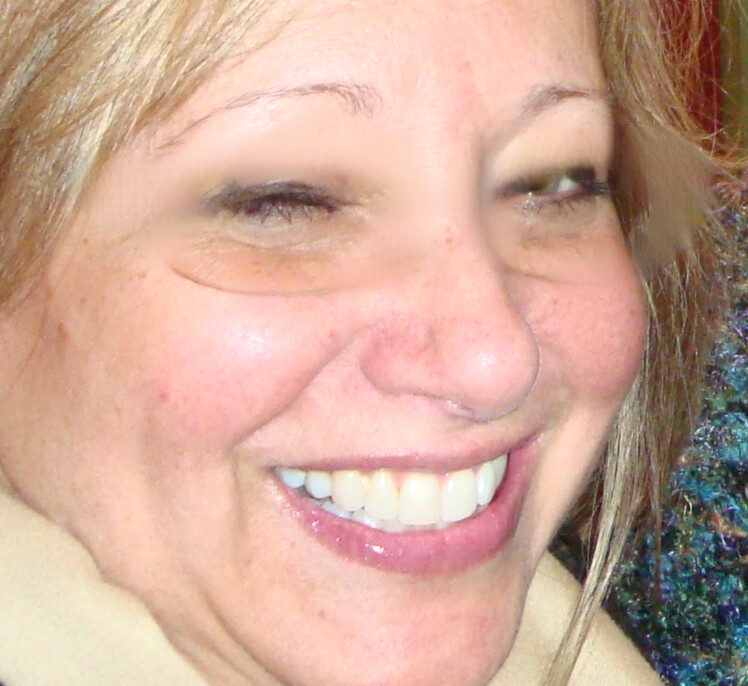} & \includegraphics[width=\linewidth]{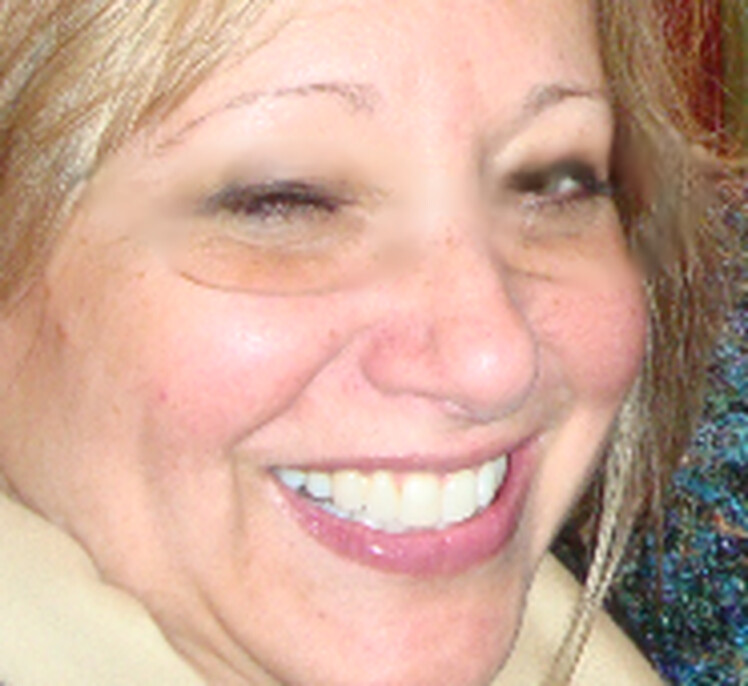} & \includegraphics[width=\linewidth]{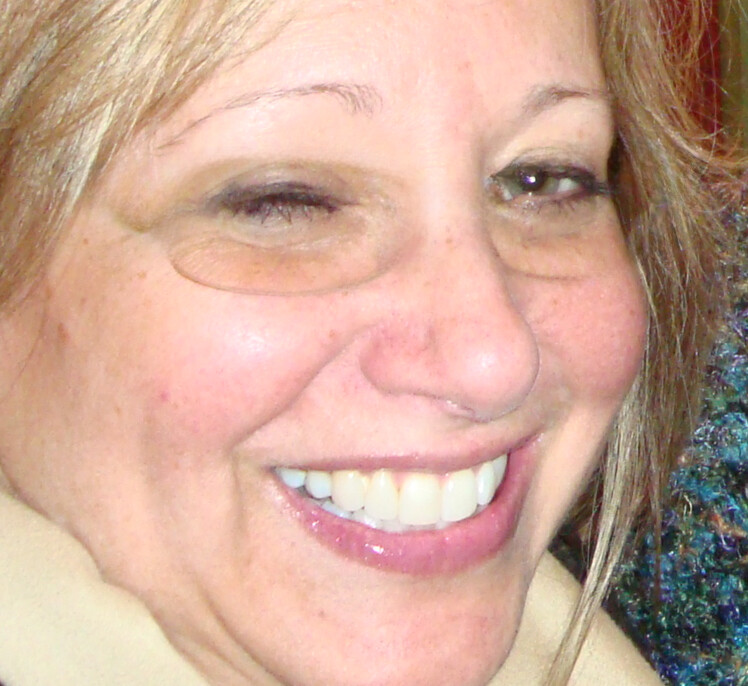} & \includegraphics[width=\linewidth]{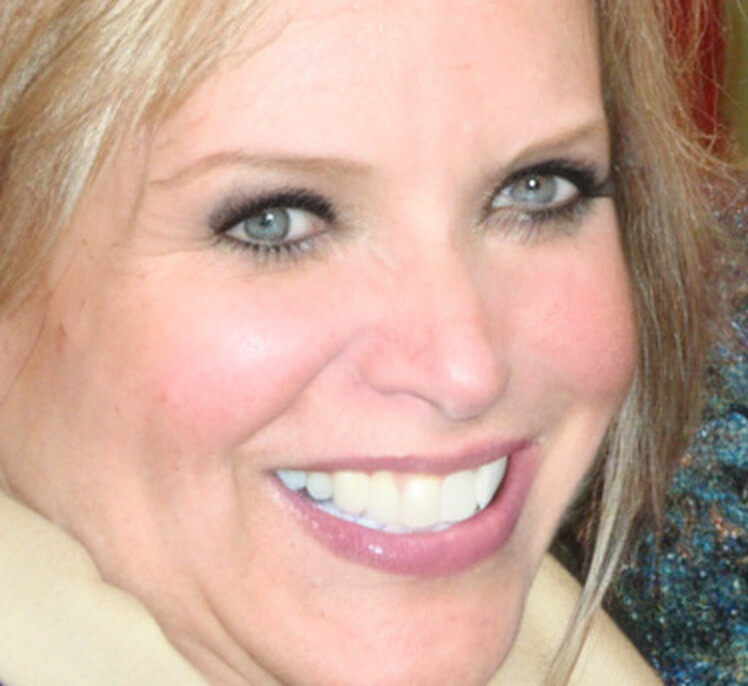} \\
    \includegraphics[width=\linewidth]{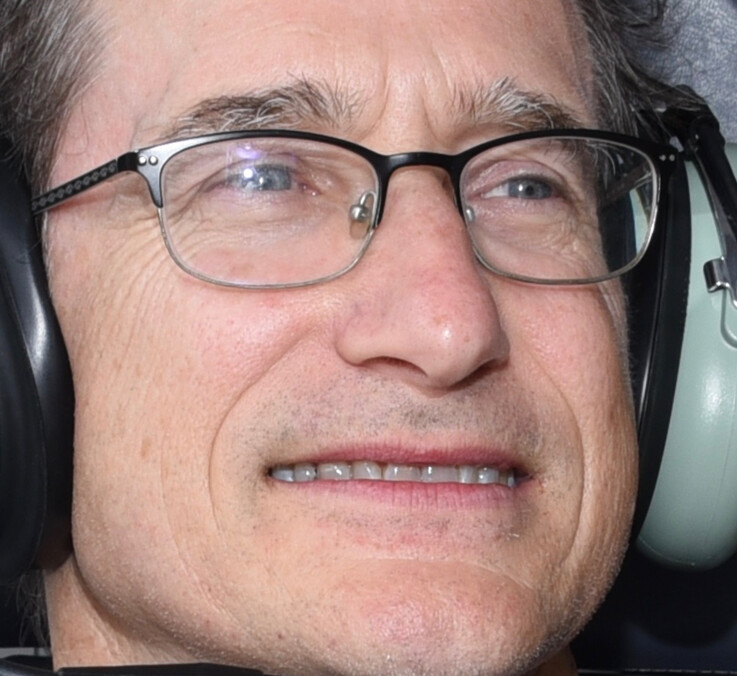} & \includegraphics[width=\linewidth]{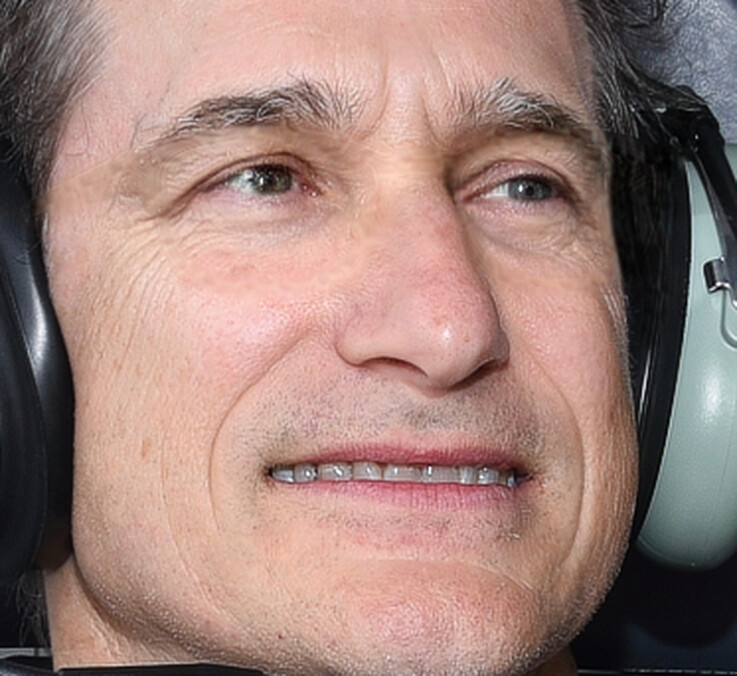} & \includegraphics[width=\linewidth]{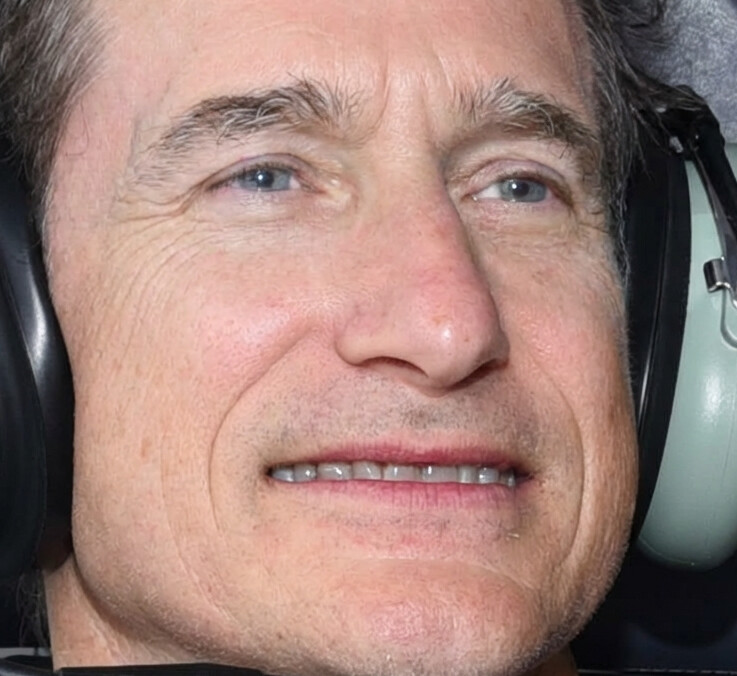} & \includegraphics[width=\linewidth]{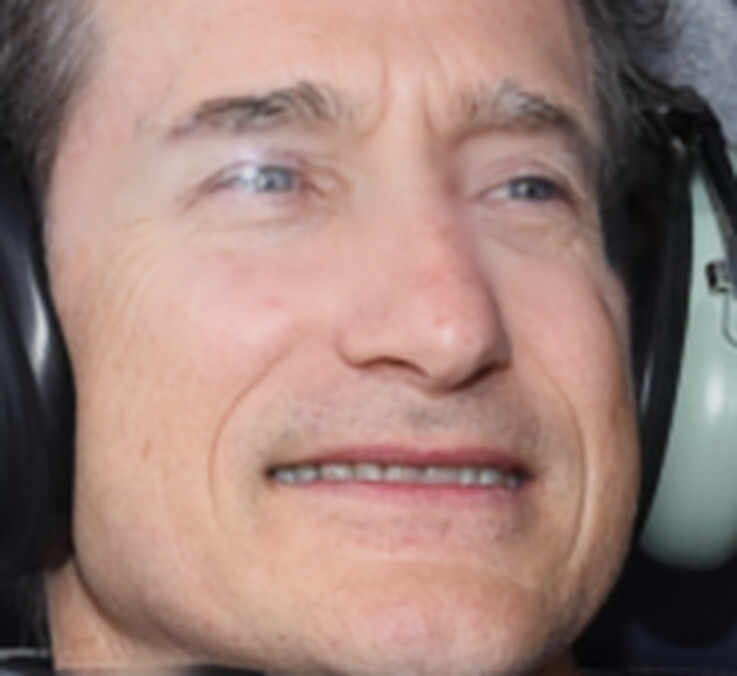} & \includegraphics[width=\linewidth]{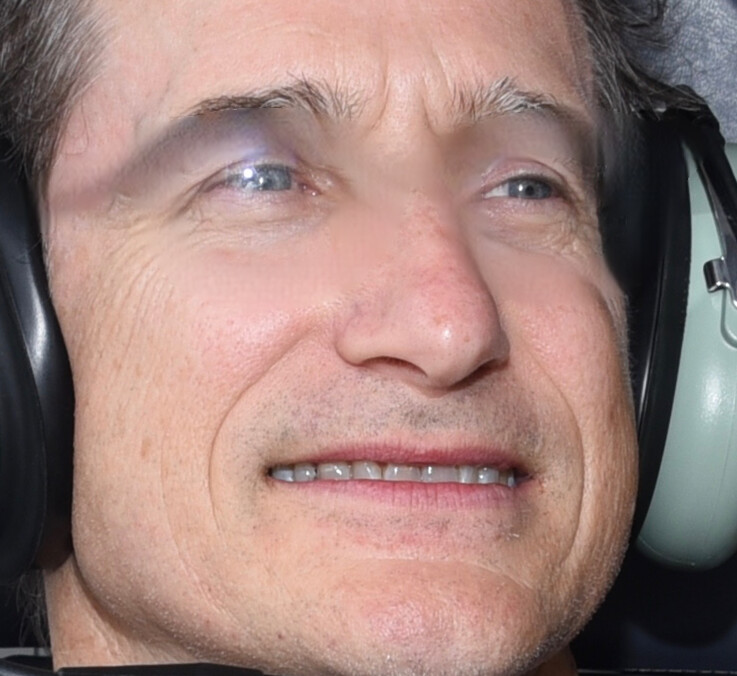} & \includegraphics[width=\linewidth]{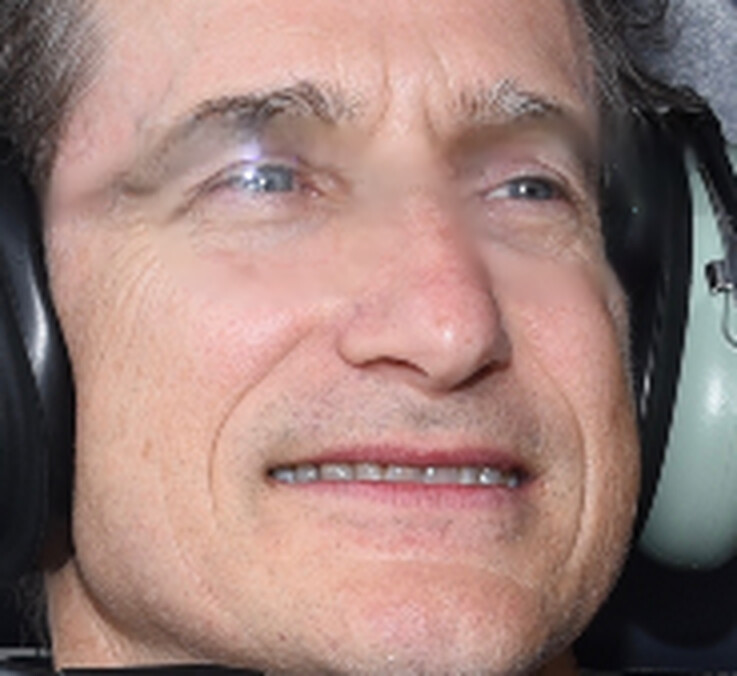} & \includegraphics[width=\linewidth]{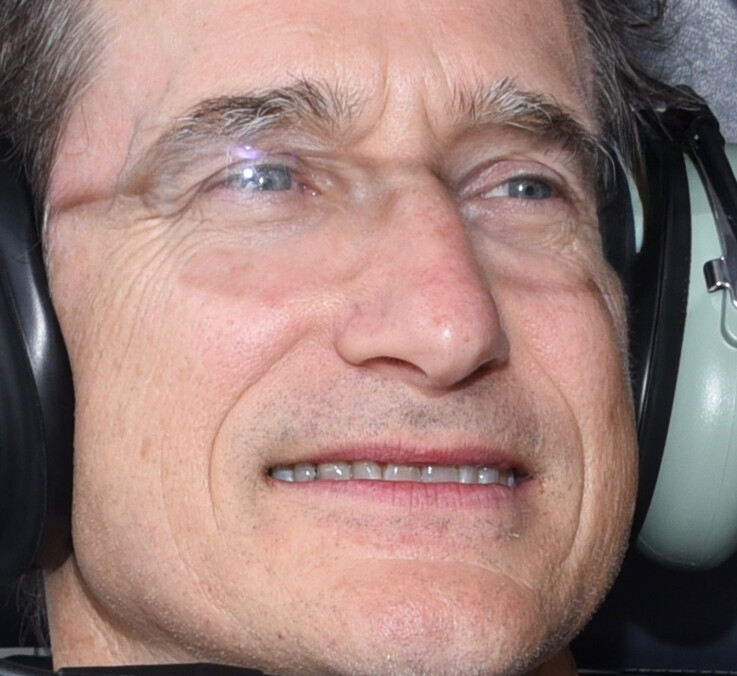} & \includegraphics[width=\linewidth]{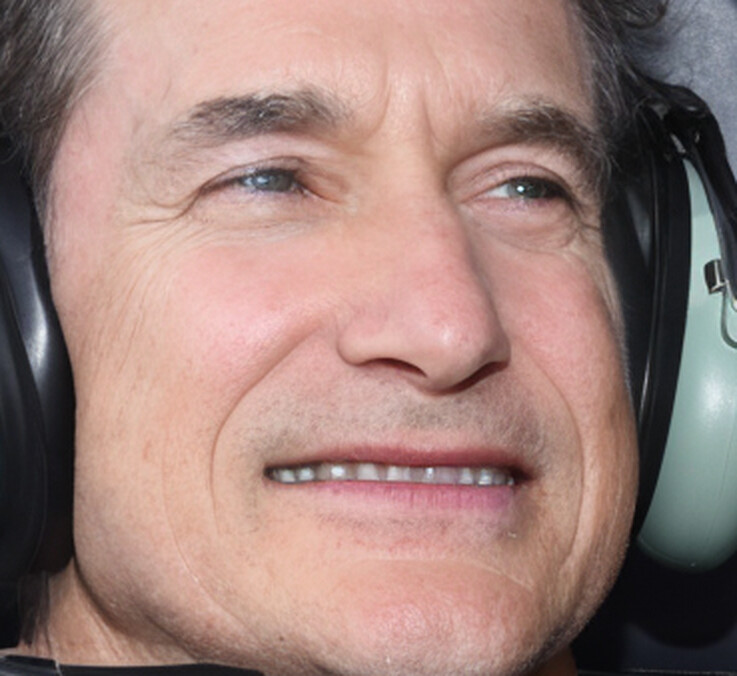} \\
    \includegraphics[width=\linewidth]{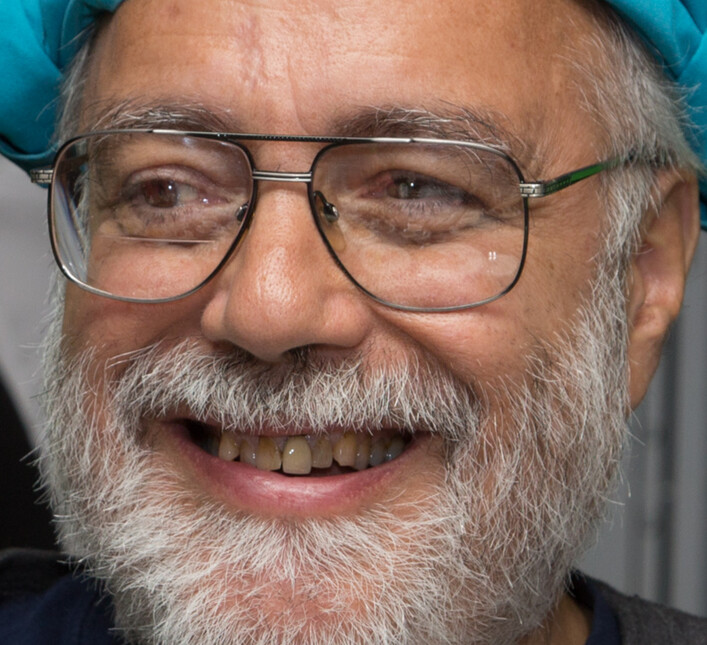} & \includegraphics[width=\linewidth]{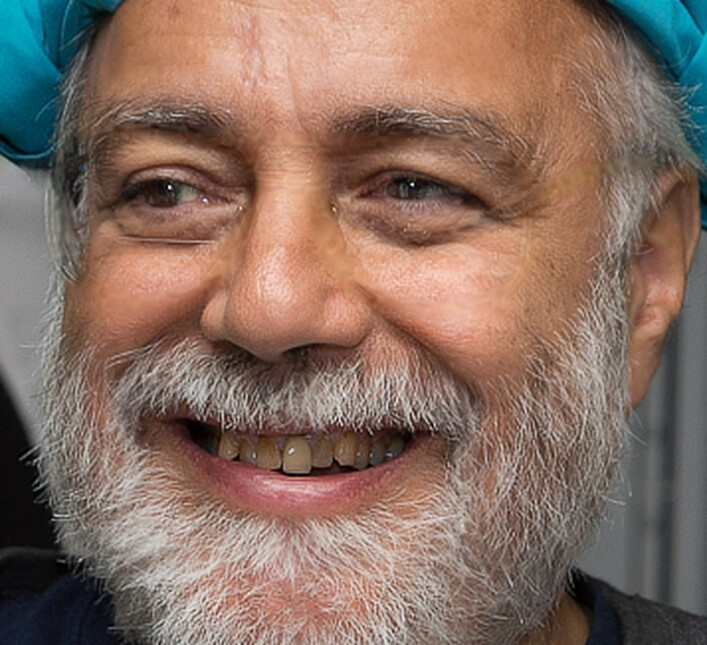} & \includegraphics[width=\linewidth]{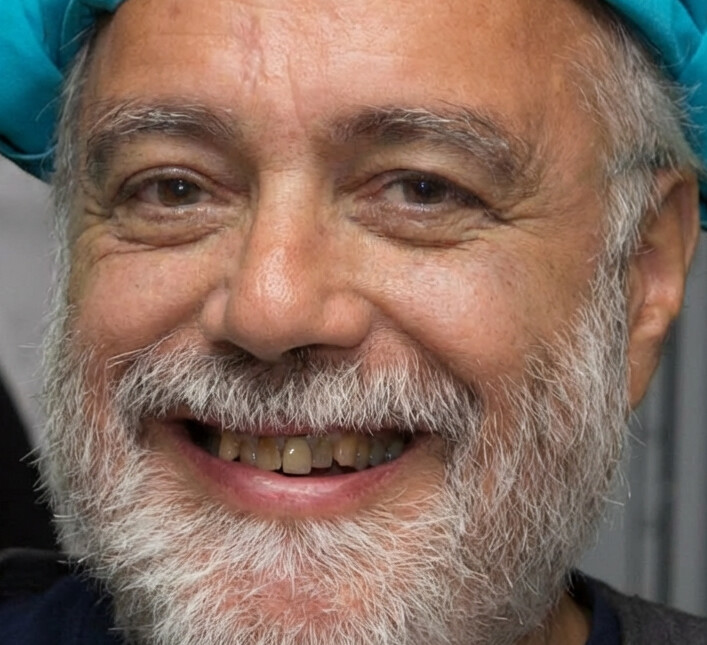} & \includegraphics[width=\linewidth]{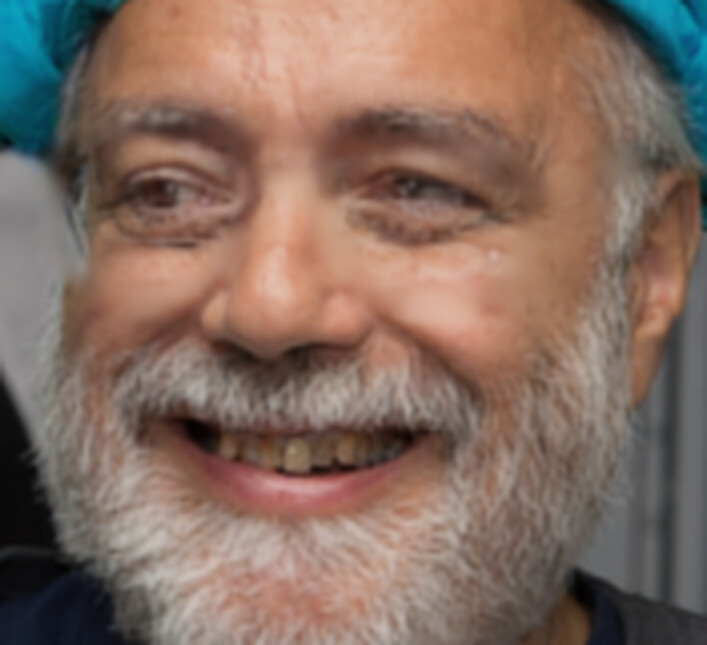} & \includegraphics[width=\linewidth]{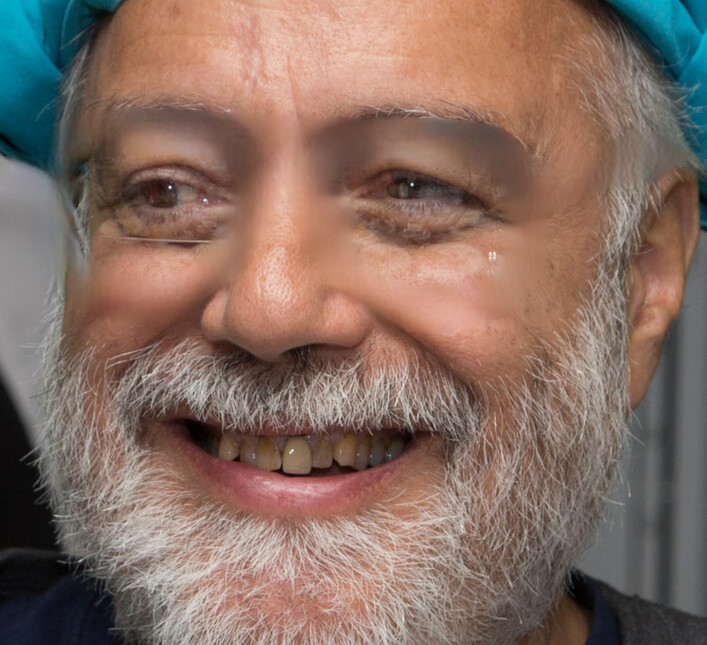} & \includegraphics[width=\linewidth]{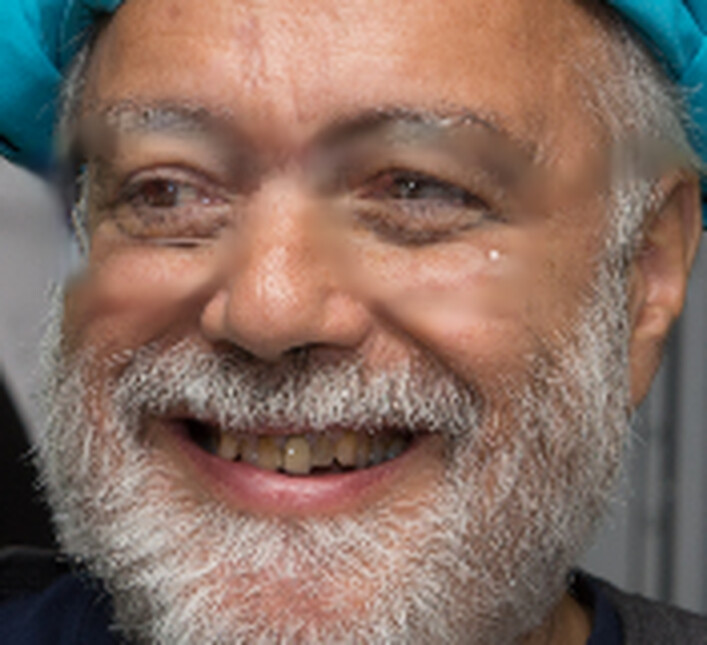} & \includegraphics[width=\linewidth]{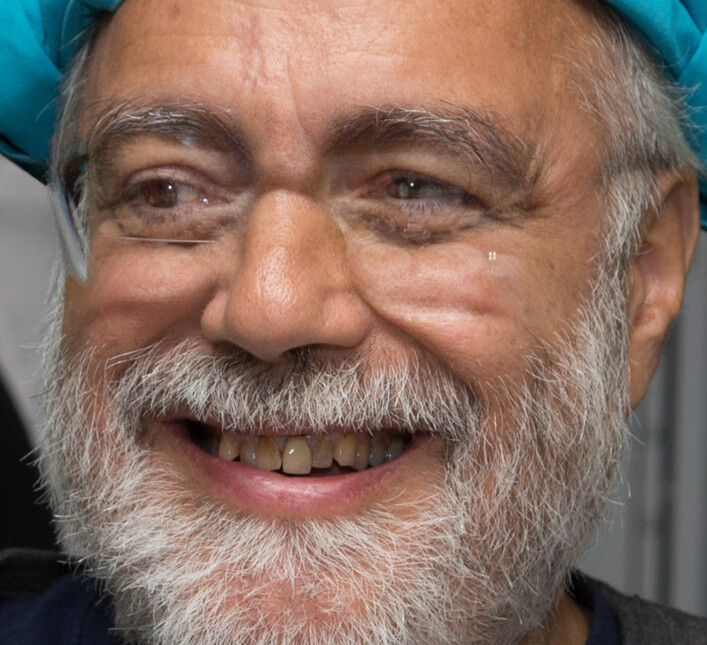} & \includegraphics[width=\linewidth]{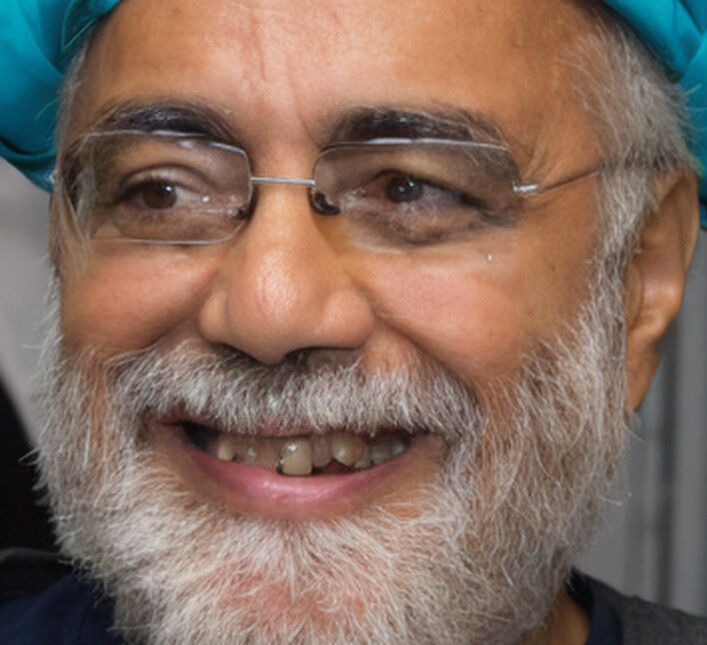} \\%
  \end{tabularx}
  \caption{\textbf{Qualitative comparison on Flickr-Faces-HQ (FFHQ)~\cite{karras_style-based_2021}.} Generative models like LEDITS and Nano Banana struggle to maintain gaze consistency and preserve fine facial details. Video completion methods like ProPainter and FGT, being designed for inpainting, often fail to recover underlying facial structures accurately. Our Joint Feature-Spatial network preserves gaze and fine facial details while maintaining high-fidelity structural integrity. See Sec.~\ref{sec:ffhq_qualitative_overview} for more details.}\vspace{-1.5em}
  \label{fig:ffhq_qualitative_overview}
\end{figure*}

In Fig.~\ref{fig:ffhq_qualitative_supplementary}, we provide a broader qualitative comparison against four additional methods: RAVE~\cite{kara_rave_2024}, IP-FaceDiff~\cite{anand_ip-facediff_2025}, TokenFlow~\cite{geyer_tokenflow_2023}, and InstructPix2Pix~\cite{brooks_instructpix2pix_2023}. While these approaches demonstrate the ability to remove eyeglass frames, they often exhibit different modes of failure in maintaining facial fidelity. In Fig.~\ref{fig:ffhq_qualitative_overview} more results are presented.

\paragraph{Stylization and Structural Shift.} As shown in Fig.~\ref{fig:ffhq_qualitative_supplementary}, diffusion-based video translation methods such as RAVE and TokenFlow tend to prioritize generative consistency over pixel-perfect identity preservation. RAVE often performs an unintended stylization of the face, resulting in an appearance that resembles a smooth, synthetic 3D model rather than a natural portrait. TokenFlow, while maintaining temporal coherence, frequently experiences identity drift where the subject's original features are largely replaced by the generative prior. Similarly, IP-FaceDiff can introduce unintended structural distortions that alter the global facial geometry.

\paragraph{Artifacts and Texture Fidelity.} InstructPix2Pix, while offering a computationally efficient image-to-image editing path, often introduces local artifacts around the ocular region and struggles to fully synthesize the skin texture previously occluded by the frames. These challenges are reflected in the higher FID scores reported in Table~\ref{tab:ffhq_results_full}, where the stochastic nature of the denoising process in these baselines leads to a greater distribution shift from natural face portraits. In contrast, our JFSnet leverages a physically-grounded synthesis pipeline to maintain high-frequency facial details and ensure that the restoration remains faithful to the original identity.

\subsection{Failure Cases}\label{sec:failure_cases}

\begin{figure*}[t]
  \centering
  \renewcommand{\arraystretch}{0}
  \setlength{\tabcolsep}{0pt}
  \begin{tabularx}{\textwidth}{@{} X X X X X X @{}}
    \includegraphics[width=\linewidth]{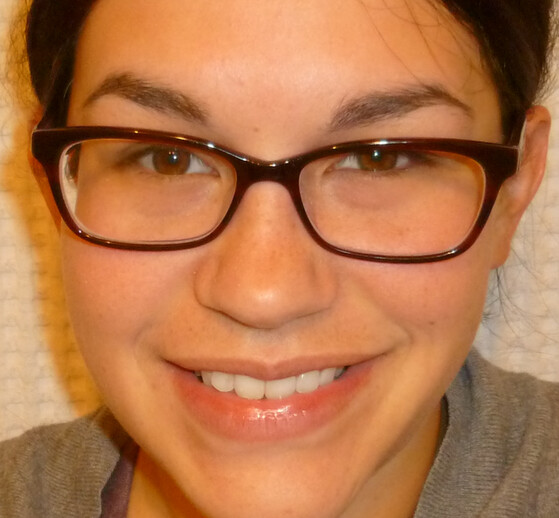} & \includegraphics[width=\linewidth]{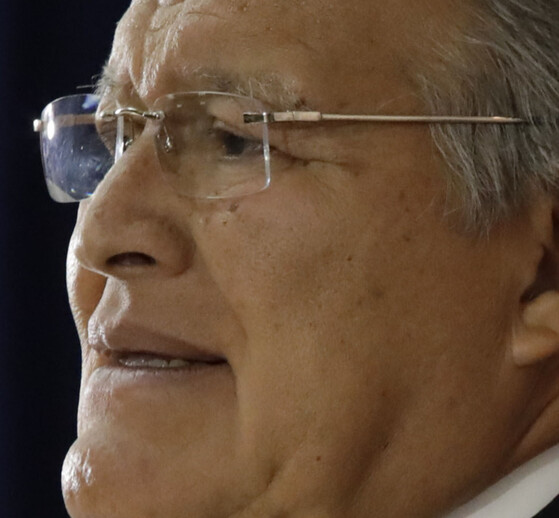} & \includegraphics[width=\linewidth]{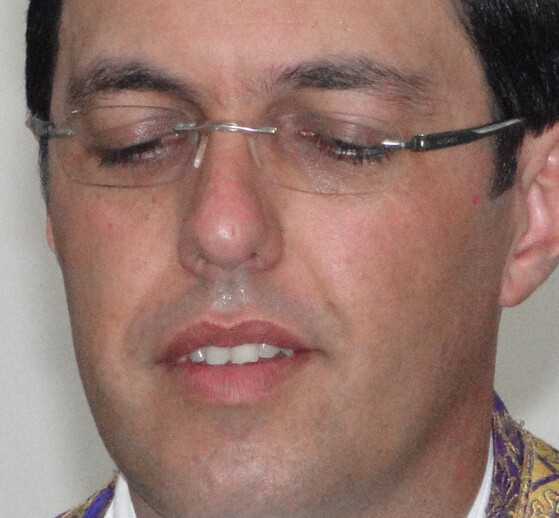} & \includegraphics[width=\linewidth]{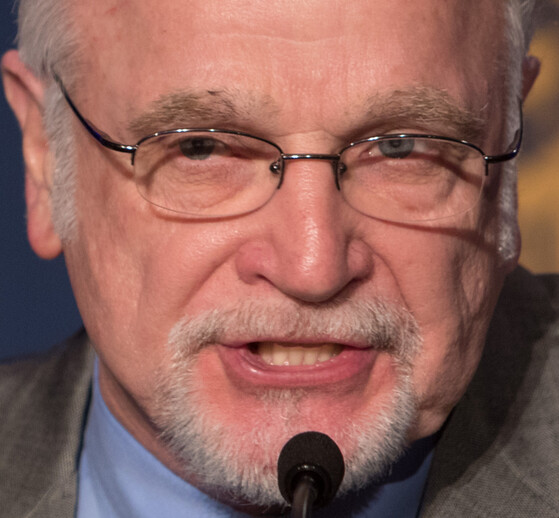} & \includegraphics[width=\linewidth]{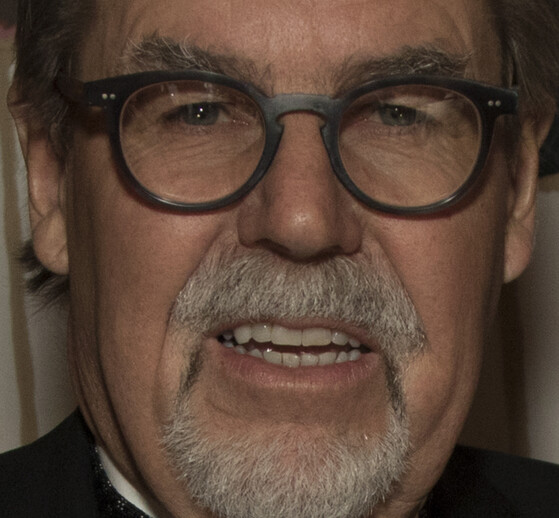} & \includegraphics[width=\linewidth]{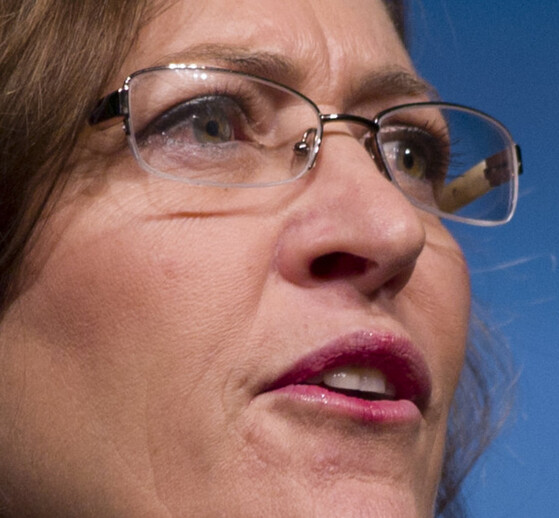} \\%
    \includegraphics[width=\linewidth]{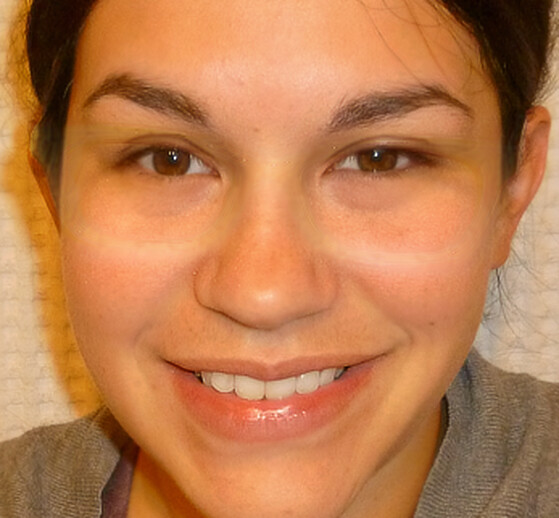} & \includegraphics[width=\linewidth]{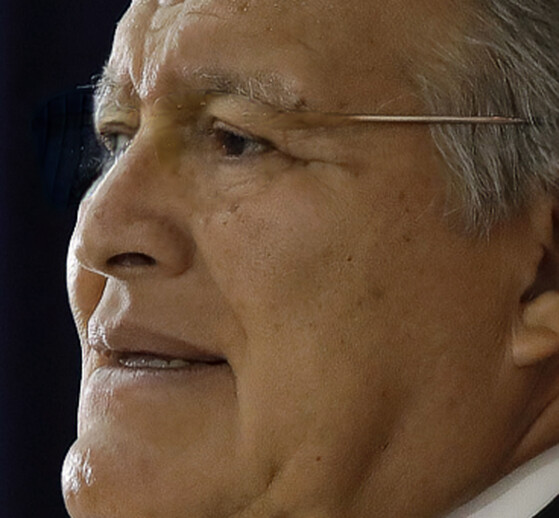} & \includegraphics[width=\linewidth]{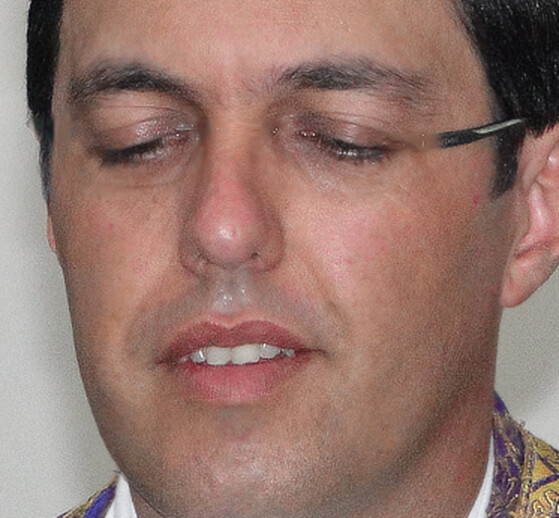} & \includegraphics[width=\linewidth]{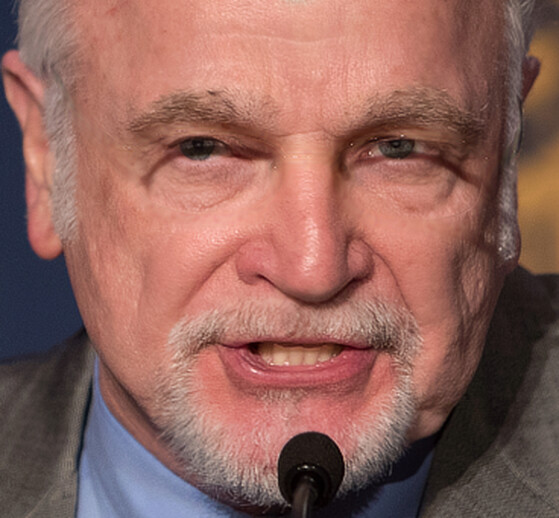} & \includegraphics[width=\linewidth]{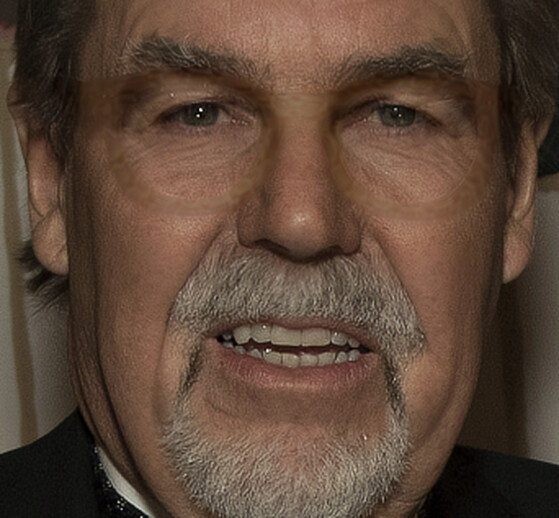} & \includegraphics[width=\linewidth]{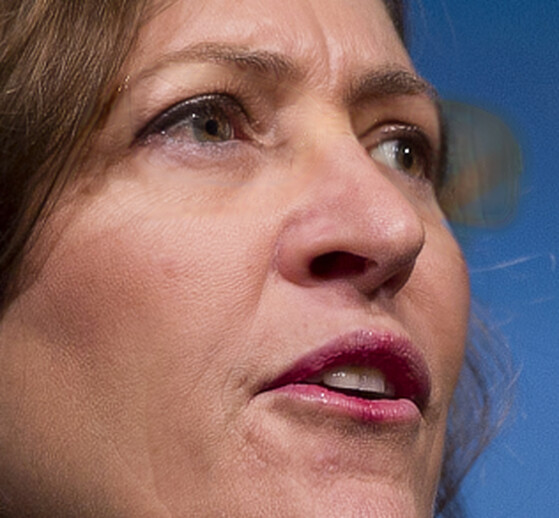} \\[0.5em]%
    \includegraphics[width=\linewidth]{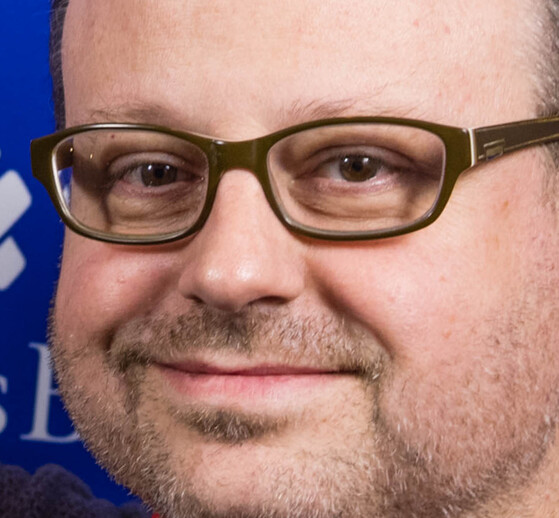} & \includegraphics[width=\linewidth]{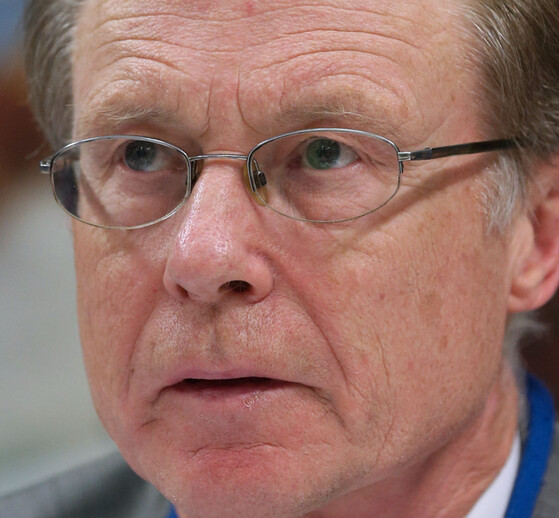} & \includegraphics[width=\linewidth]{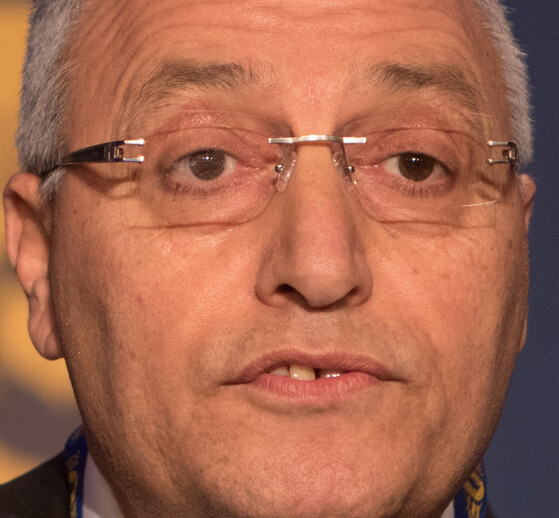} & \includegraphics[width=\linewidth]{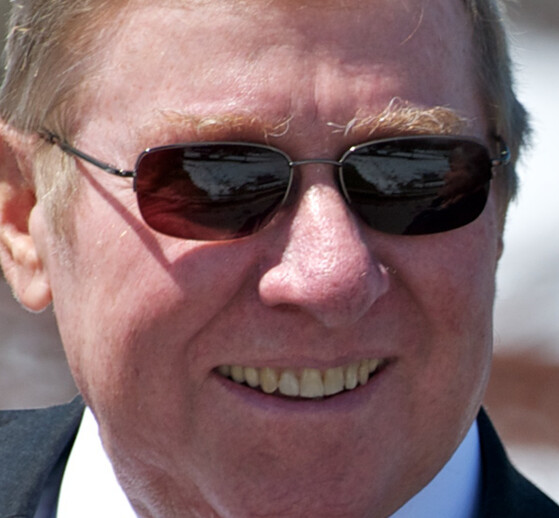} & \includegraphics[width=\linewidth]{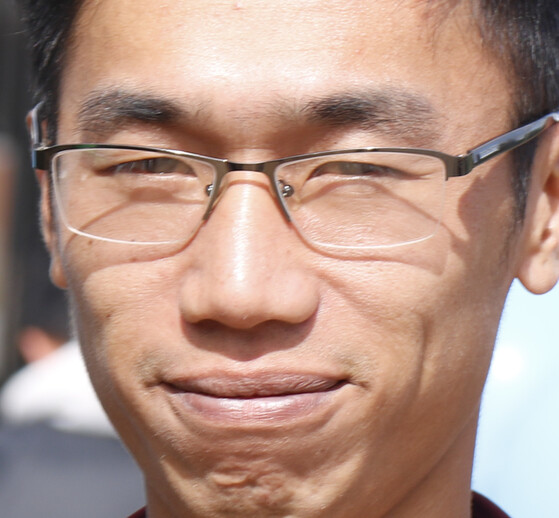} & \includegraphics[width=\linewidth]{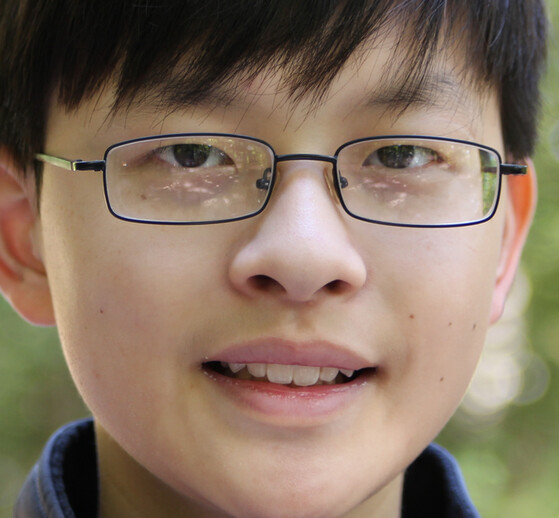} \\%
    \includegraphics[width=\linewidth]{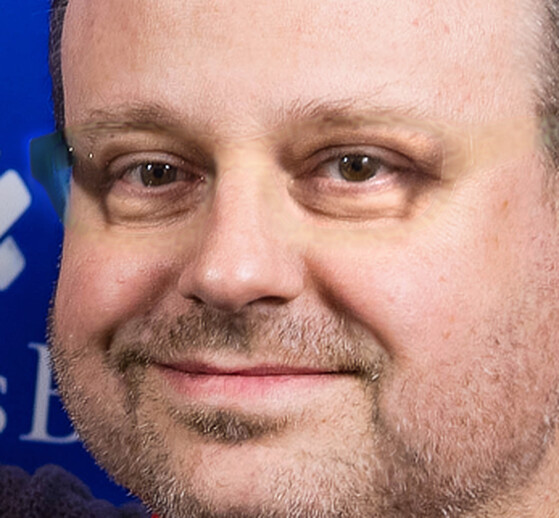} & \includegraphics[width=\linewidth]{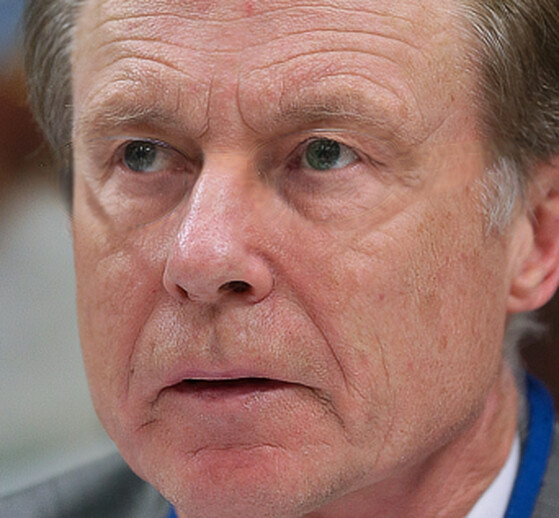} & \includegraphics[width=\linewidth]{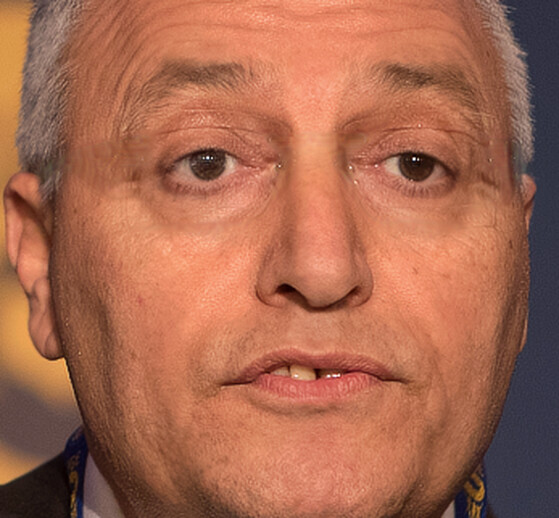} & \includegraphics[width=\linewidth]{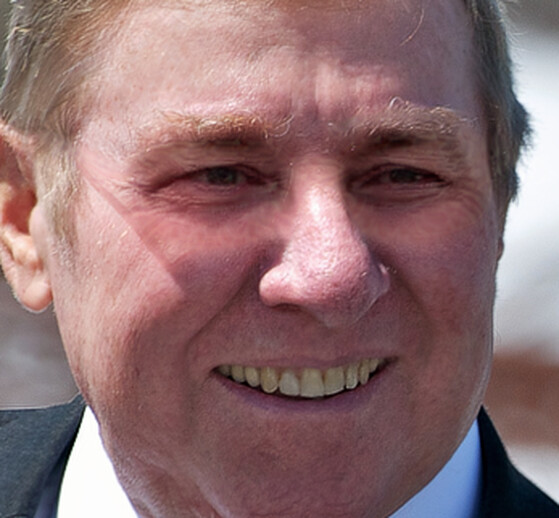} & \includegraphics[width=\linewidth]{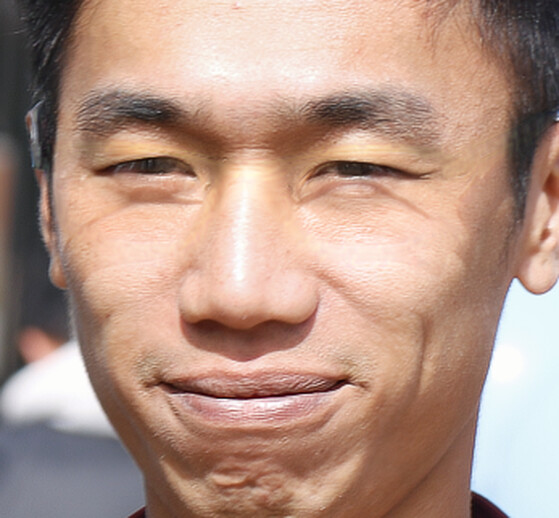} & \includegraphics[width=\linewidth]{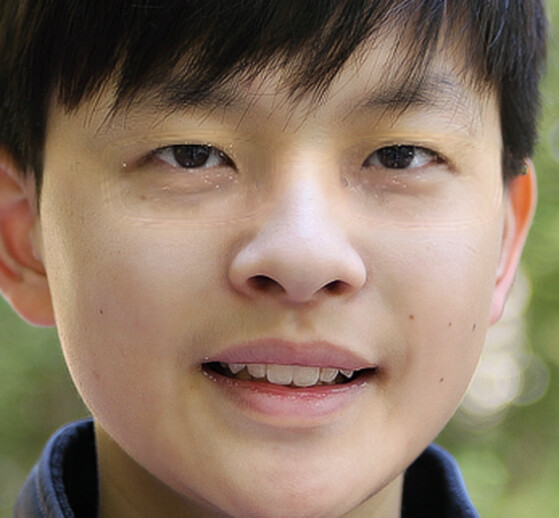} \\[0.5em]%
    \includegraphics[width=\linewidth]{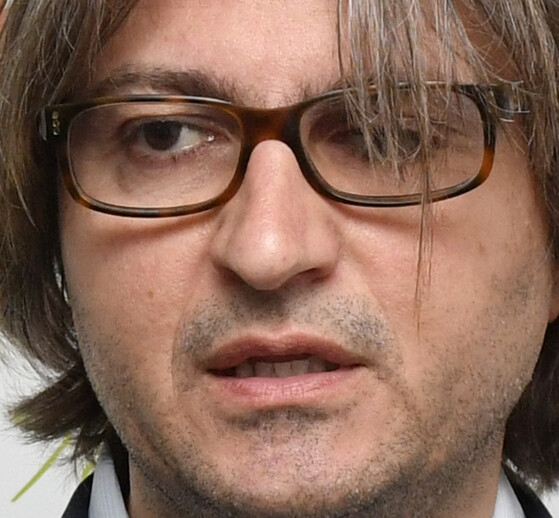} & \includegraphics[width=\linewidth]{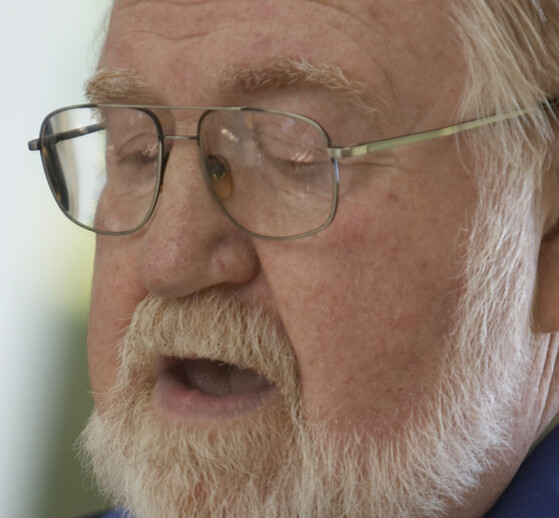} & \includegraphics[width=\linewidth]{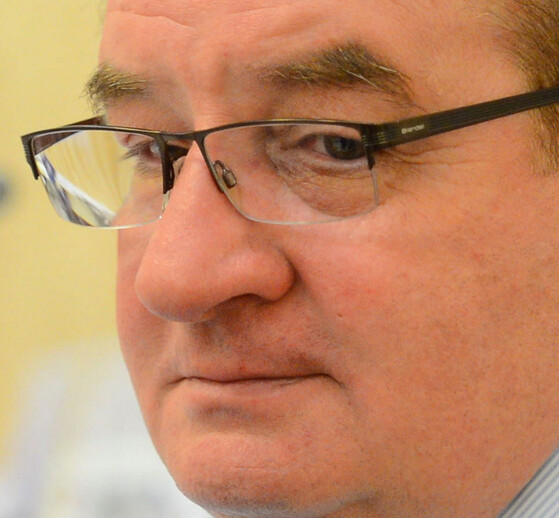} & \includegraphics[width=\linewidth]{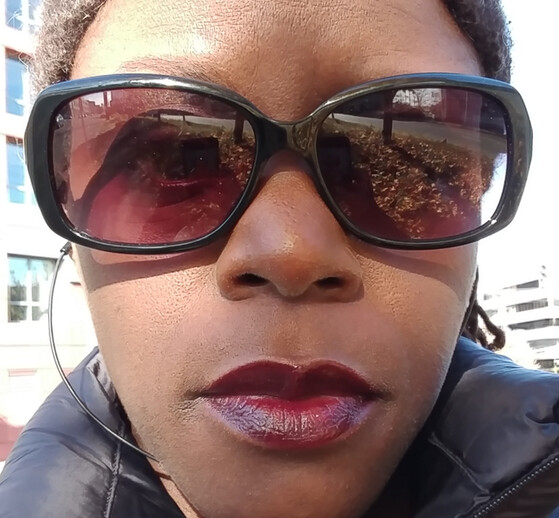} & \includegraphics[width=\linewidth]{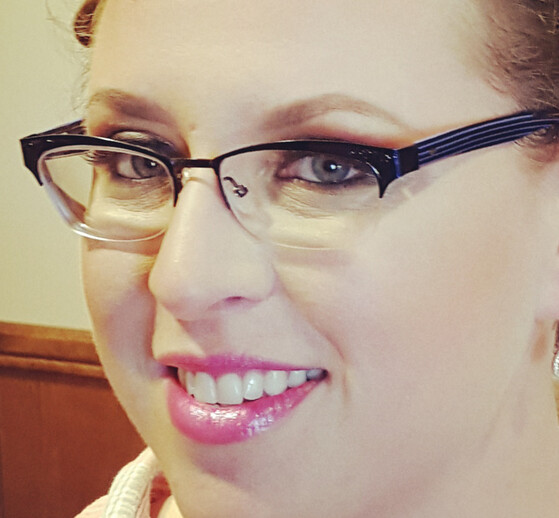} & \includegraphics[width=\linewidth]{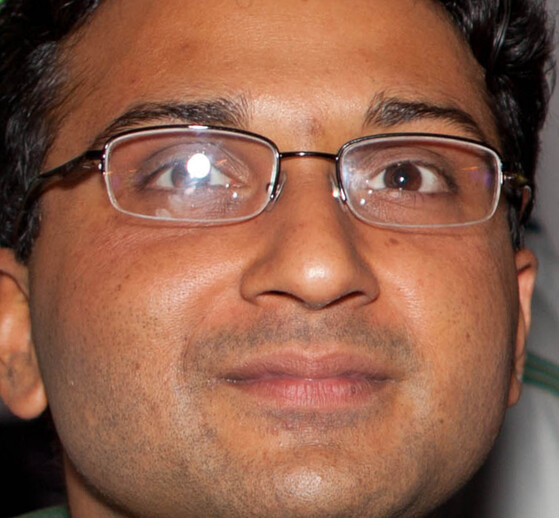} \\%
    \includegraphics[width=\linewidth]{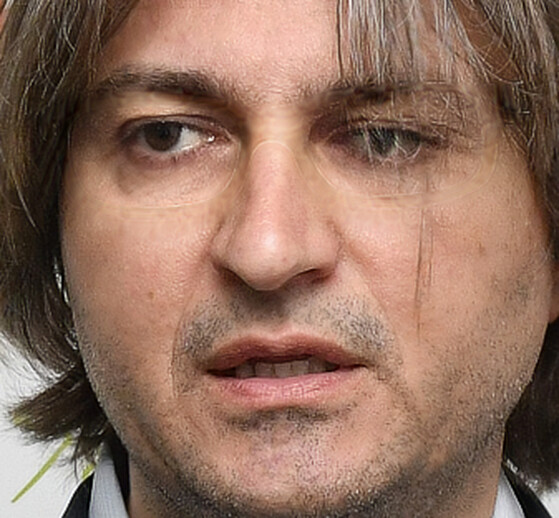} & \includegraphics[width=\linewidth]{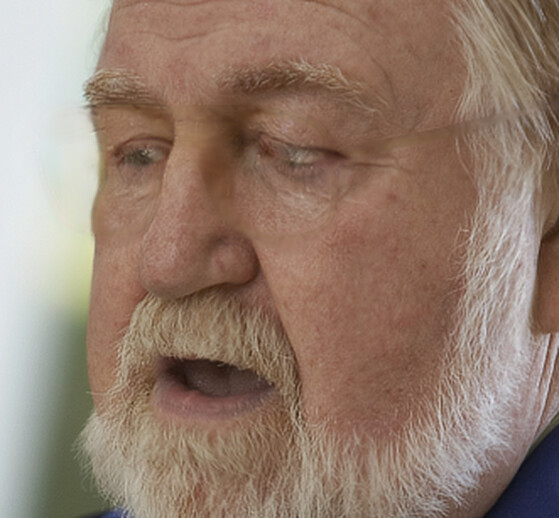} & \includegraphics[width=\linewidth]{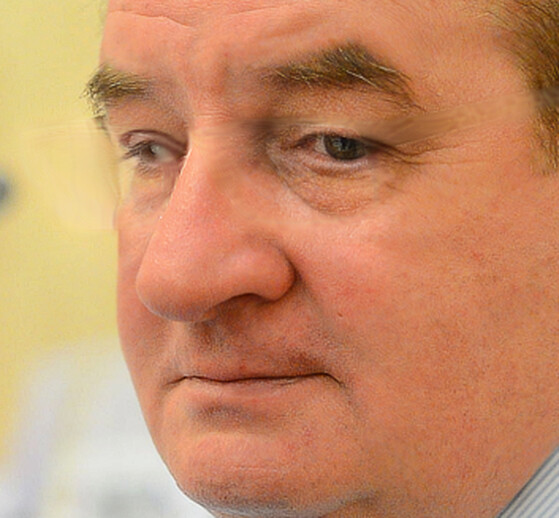} & \includegraphics[width=\linewidth]{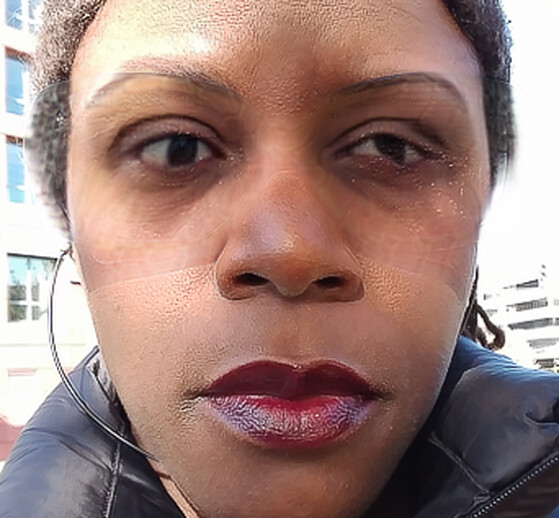} & \includegraphics[width=\linewidth]{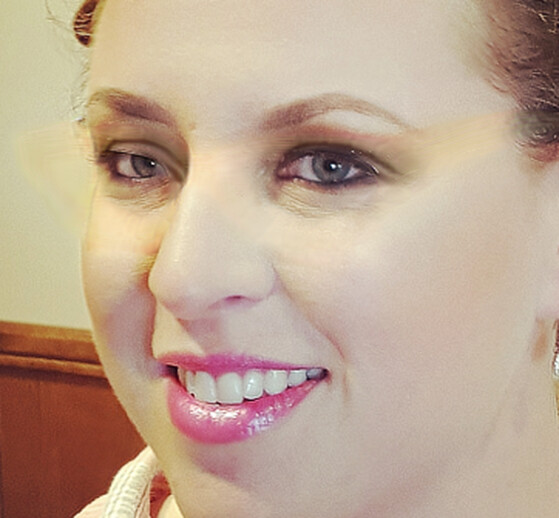} & \includegraphics[width=\linewidth]{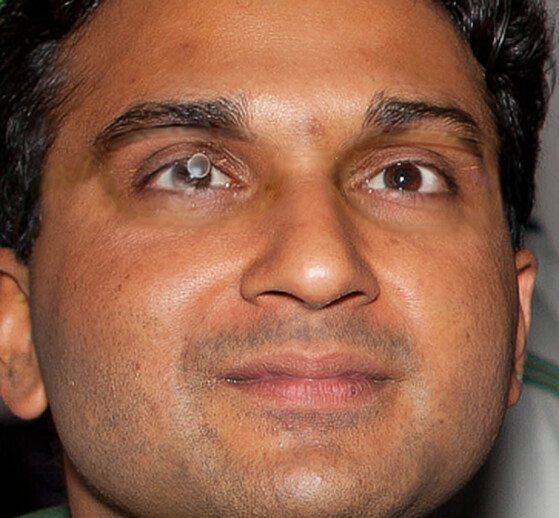} \\[0.5em]%
    \includegraphics[width=\linewidth]{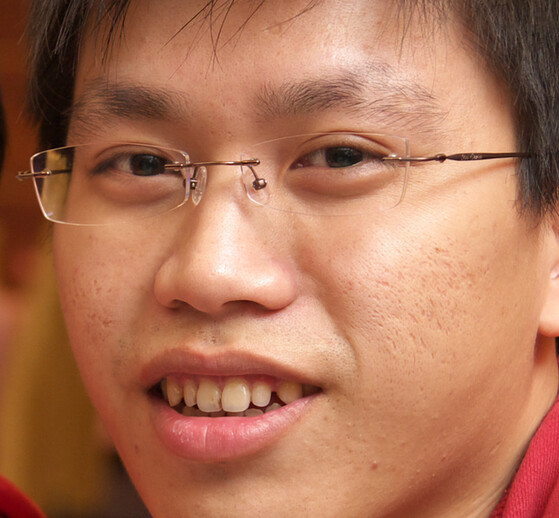} & \includegraphics[width=\linewidth]{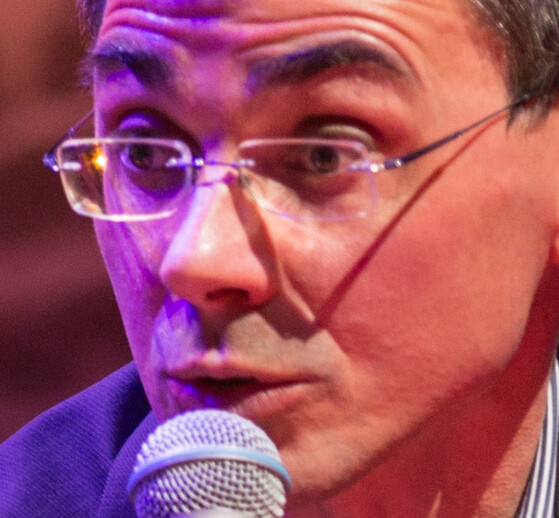} & \includegraphics[width=\linewidth]{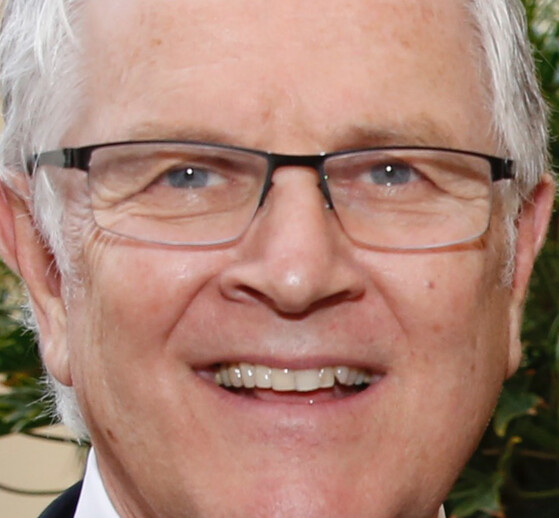} & \includegraphics[width=\linewidth]{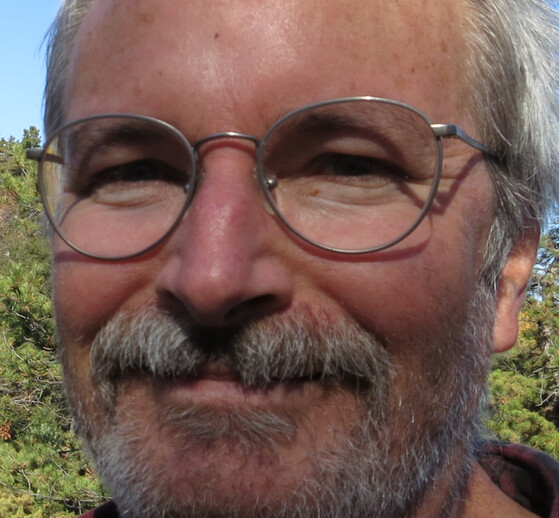} & \includegraphics[width=\linewidth]{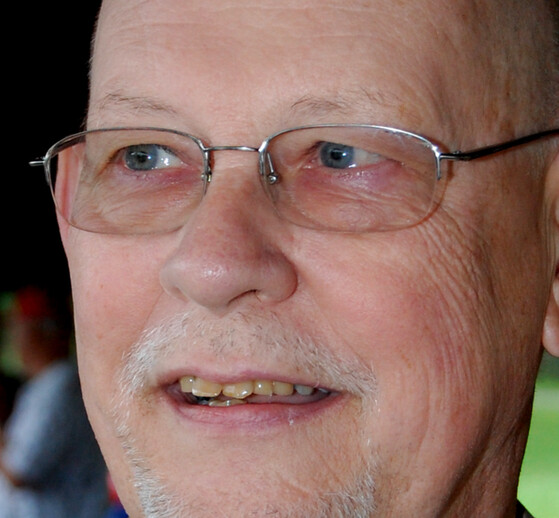} & \includegraphics[width=\linewidth]{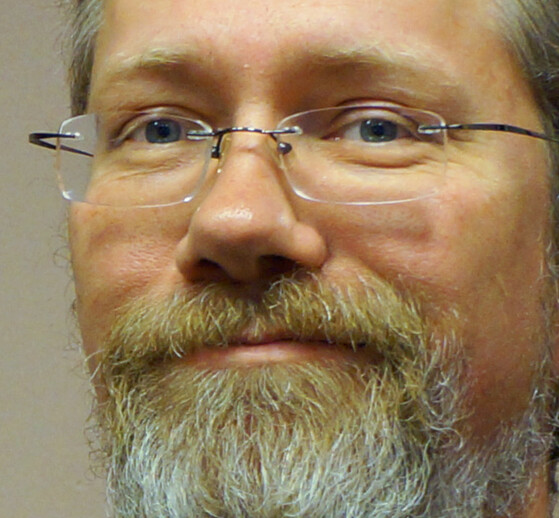} \\%
    \includegraphics[width=\linewidth]{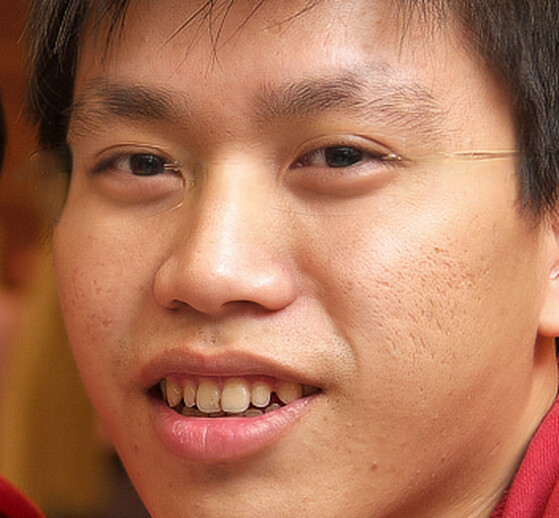} & \includegraphics[width=\linewidth]{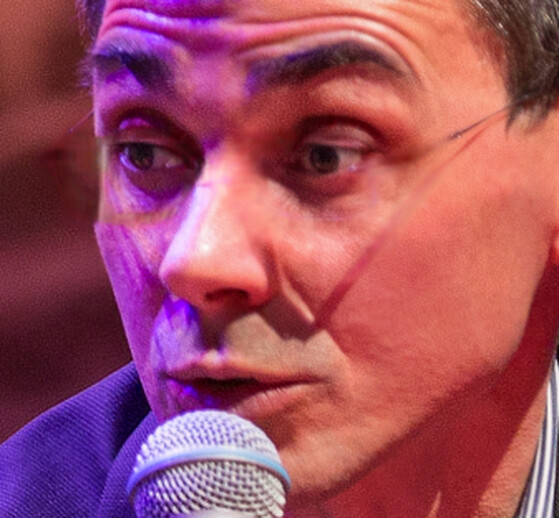} & \includegraphics[width=\linewidth]{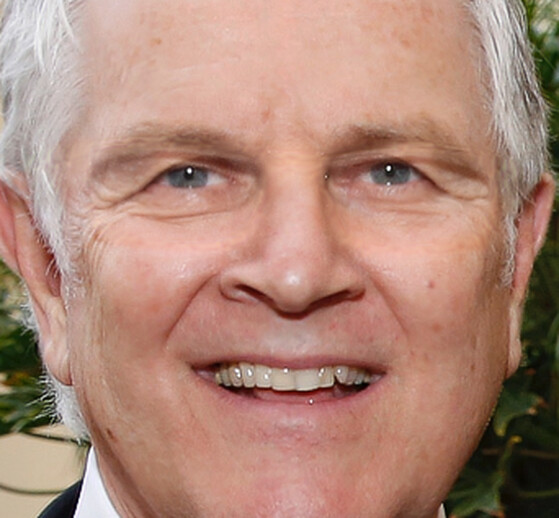} & \includegraphics[width=\linewidth]{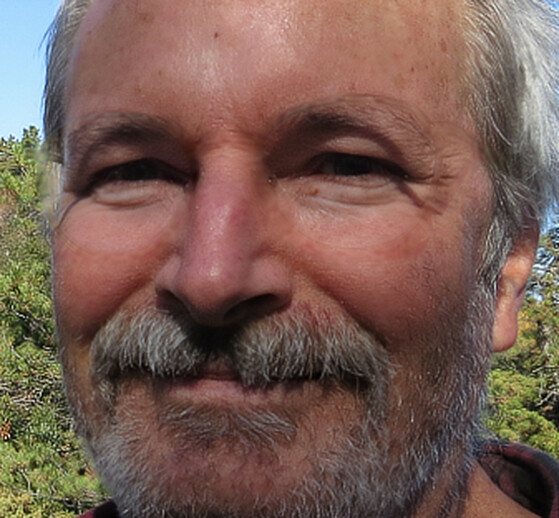} & \includegraphics[width=\linewidth]{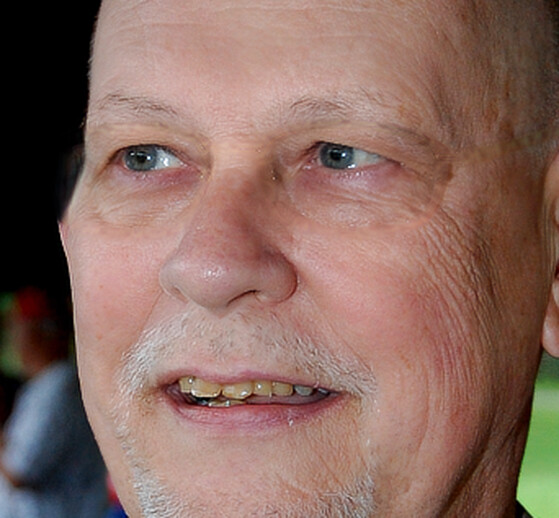} & \includegraphics[width=\linewidth]{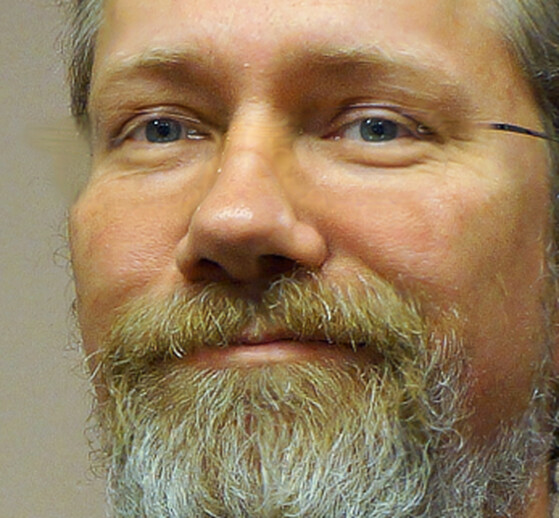} \\[0.5em]%
  \end{tabularx}
  \caption{\textbf{Failure cases of our approach.} We identify several failure modes, shown in row pairs (input and our result) from top to bottom: (i) incomplete removal of eyewear, (ii) incomplete removal of shades cast by glasses on person's face, (iii) iris deformation, and (iv) refraction undercorrection.}
  \label{fig:jfs_failure_cases}
\end{figure*}

Typical failure cases of our approach are presented in Fig.~\ref{fig:jfs_failure_cases}. These can be classified into four primary categories: (i) incomplete removal of eyewear, (ii) residual shadows cast by the frames on the subject's face, (iii) minor iris deformation, and (iv) refraction undercorrection. We attribute the first three categories to artifacts or misalignments in the synthetic training data, which could be mitigated through more rigorous dataset cleaning. The fourth category arises from the simplified refraction model used in our simulation, which may not fully capture the complex lens geometries of certain high-power prescriptions.

%
%

\end{document}